\documentclass[conference]{IEEEtran}
\IEEEoverridecommandlockouts

\usepackage{cite}
\usepackage{amsmath,amssymb,amsfonts}
\usepackage{algorithm}
\usepackage{algorithmic}
\usepackage{bm}
\usepackage{bbding}
\usepackage{pifont}
\usepackage{diagbox}
\usepackage{enumitem}
\usepackage{graphicx}
\usepackage{booktabs}
\usepackage{multirow}
\usepackage{makecell}
\usepackage[table]{xcolor}
\usepackage{textcomp}
\usepackage{url}
\usepackage{verbatim}
\usepackage{xspace}
\usepackage{eso-pic}

\graphicspath{{last/}}

\definecolor{mypink}{rgb}{.99,.91,.95}
\definecolor{lightgreen}{rgb}{0.98, 0.98, 0.82}
\definecolor{graycolor}{rgb}{0.86,0.86,0.86}
\definecolor{lightgraycolor}{rgb}{0.92,0.92,0.92}

\newcommand{\red}[1]{\textcolor{black}{\textbf{#1}}}
\newcommand{\blue}[1]{\textcolor{black}{\underline{#1}}}
\newcommand{\gray}[1]{\textcolor{gray}{#1}}
\AtBeginDocument{%
  \setlength{\abovedisplayskip}{4pt plus 1pt minus 1pt}%
  \setlength{\belowdisplayskip}{4pt plus 1pt minus 1pt}%
  \setlength{\abovedisplayshortskip}{2pt plus 1pt}%
  \setlength{\belowdisplayshortskip}{2pt plus 1pt minus 1pt}}

\renewcommand{\footnoterule}{%
  \ifnum\value{page}=1\relax
    \kern-3pt
    \hrule width\columnwidth height0.4pt
    \kern2.6pt
  \fi}

\begin{document}
\pagestyle{empty}
\AddToShipoutPictureFG*{%
  \AtPageLowerLeft{%
    \put(\LenToUnit{\dimexpr 1in+\oddsidemargin\relax},\LenToUnit{12.7mm}){%
      \parbox[b]{\columnwidth}{%
        \normalfont\footnotesize\raggedright
        \textsuperscript{*}Corresponding author: Ying Fu.\\
        979-8-3195-3836-9/26/\$31.00 \textcopyright{}2026 IEEE}}}}

\title{Unified RAW Image Exposure Correction and Restoration via Linear Illumination and Frequency-Adaptive Enhancement
\thanks{This work was supported by the National Natural Science Foundation of China (62331006, 62671049), and the Open Project Program of State Key Laboratory of CNS/ATM.}}

\author{
\IEEEauthorblockN{1\textsuperscript{st} Yingkai Zhang}
\IEEEauthorblockA{\textit{School of Computer Science}\\
\textit{and Technology}\\
\textit{Beijing Institute of Technology}\\
Beijing, China\\
zhangyingkai@bit.edu.cn}
\and
\IEEEauthorblockN{2\textsuperscript{nd} Yunhao Zou}
\IEEEauthorblockA{\textit{School of Computer Science}\\
\textit{and Technology}\\
\textit{Beijing Institute of Technology}\\
Beijing, China\\
yunhaozou@outlook.com}
\and
\IEEEauthorblockN{3\textsuperscript{rd} Ying Fu\textsuperscript{*}}
\IEEEauthorblockA{\textit{School of Computer Science}\\
\textit{and Technology}\\
\textit{Beijing Institute of Technology}\\
Beijing, China\\
fuying@bit.edu.cn}
}

\maketitle
\thispagestyle{empty}

\begin{abstract}
Incorrect exposure degrades both image brightness and recoverable scene detail. Existing sRGB-based correction methods are limited by the non-linear image signal processing pipeline, whereas RAW-based methods are usually developed for a single degradation type, such as low-light denoising or high-light recovery. 
In this paper, we focus on joint exposure correction and restoration directly in the RAW domain and propose a unified RAW Image Exposure Correction and Restoration (RIECR) framework for both under- and over-exposed images.
RIECR decomposes the problem into two coupled sub-problems that follow the RAW imaging model: linear illumination correction and residual degradation restoration.
First, a Linear Illumination Correction Network (LICN) estimates exposure-aware linear gains by using channel-specific RAW cues. It exploits the high signal-to-noise ratio of Green channels for robust brightening and the relatively unsaturated Red/Blue responses for high-light darkening.
Second, a dual-branch Frequency-based Enhancement Network (FEN) restores the remaining degradation with a Frequency Dynamic Convolution (FDC) module. FDC decomposes convolutional responses into low- and high-frequency components and adaptively suppresses amplified noise or recovers clipped textures according to the exposure context.
We further build a real paired RAW benchmark containing 223 scenes and 1,784 pairs covering complex multi-exposure errors. Experiments against sixteen state-of-the-art methods show that RIECR achieves state-of-the-art performance on real RAW exposure correction.
\end{abstract}

\begin{IEEEkeywords}
\itshape
RAW image restoration, exposure correction, linear illumination adjustment, frequency enhancement, RAW image dataset
\end{IEEEkeywords}

\begin{figure*}[tbp]
\begin{center}
\setlength{\tabcolsep}{0.1cm}
    \begin{minipage}[h]{0.49\textwidth}
          \centering
          \small
          \includegraphics[width=1\linewidth]{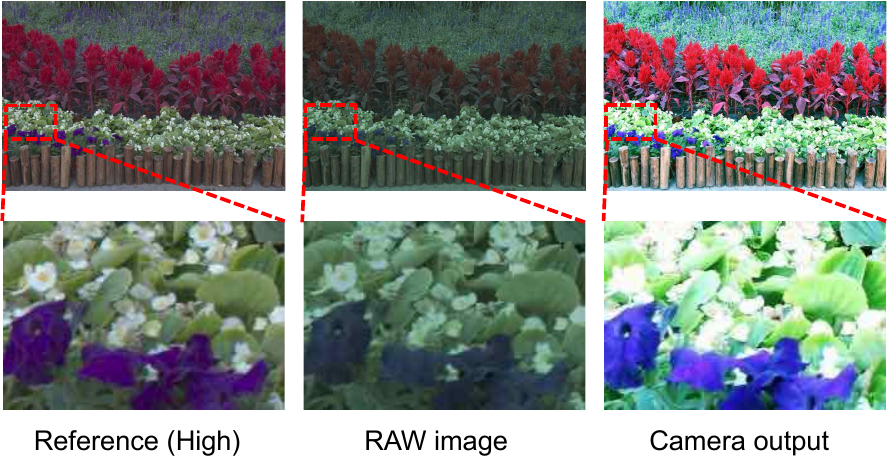}
      \end{minipage}
    \hfill
    \begin{minipage}[h]{0.49\textwidth}
          \centering
          \small
          \includegraphics[width=1\linewidth]{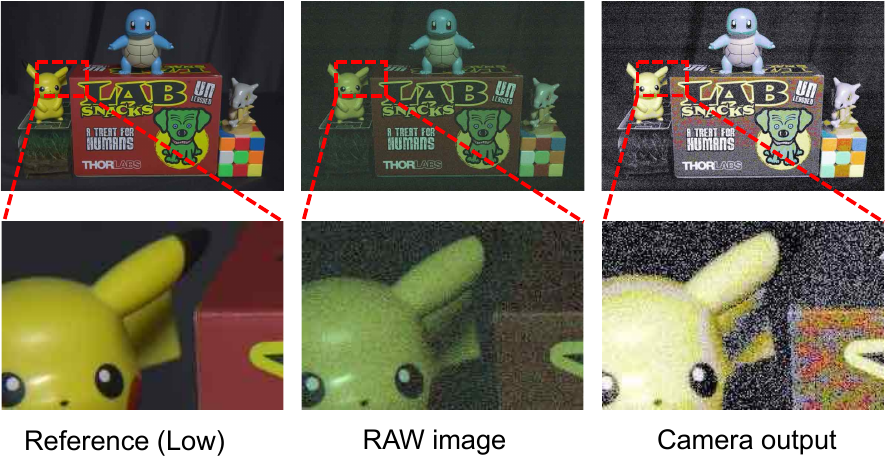}
      \end{minipage}
\end{center}
   \caption{The RAW/sRGB images of indoor/outdoor scenes under different lighting conditions. Due to the limited dynamic range of the camera, the 8-bit sRGB output loses significant information from the scene under high-light conditions. The stamens of the white flowers are almost invisible in the sRGB image but are visible in the 14-bit RAW image. Due to the low signal-to-noise (SNR) and low signal intensity under low-light conditions, the sRGB image loses considerable detail. The tail of Pikachu and the edge of the box are essentially unrecognizable in the sRGB image, whereas they are discernible in the RAW image (Zoom in for better details).}
    \label{fig: raw/rgb information}
\end{figure*}

\section{Introduction}

Exposure correction improves visibility and preserves scene detail for both photography and downstream tasks such as segmentation and detection~\cite{chen2023instance,oza2023unsupervised,chen2026survey,liu2024infrared,liu2026rdanet}. Yet exposure errors are not a single degradation~\cite{fu2024technique}. Limited photons in under-exposed scenes produce photon shot noise, sensor noise~\cite{wei2020physics}, and color distortion~\cite{guo2023low}; limited sensor range in over-exposed scenes saturates and irreversibly clips bright structures~\cite{fu2023raw}. A practical model must therefore restore exposure-dependent content as well as brightness.


Learning-based sRGB correction has progressed rapidly~\cite{afifi2021learning,huang2022deep,huang2022exposure-consistency,huang2023learning,baek2023luminance,jung2025exposure,li2024color,yang2023learning}. However, sRGB is produced by a nonlinear ISP involving demosaicking, white balance, tone mapping, and quantization~\cite{chen2018learning}; amplified noise or clipped highlights are then difficult to recover. RAW data, as shown in Fig.~\ref{fig: raw/rgb information}, retains higher bit depth and a near-linear radiance response~\cite{fu2023raw,huang2022towards,liu2021benchmarking}, making it a better domain for physical exposure correction. Nevertheless, RAW methods remain task-specific, separating low-light denoising~\cite{chen2018learning,wei2021physics} from high-light recovery~\cite{fu2023raw}. A unified model must estimate a valid linear gain and restore the direction-dependent residual within one network.

To this end, we propose RAW Image Exposure Correction and Restoration \textbf{(RIECR)}, which explicitly decomposes the task into two coupled sub-problems. \textbf{First}, the Linear Illumination Correction Network \textbf{(LICN)} predicts the exposure direction and estimates linear gains from complementary RAW channels: Green offers robust high-SNR cues for low-light brightening, while Red/Blue often remain less saturated and guide high-light darkening. \textbf{Second}, the Frequency-based Enhancement Network \textbf{(FEN)} restores the remaining degradation through local convolution and global attention. Its Frequency Dynamic Convolution \textbf{(FDC)} decomposes kernel responses and adaptively suppresses amplified high-frequency noise or emphasizes missing texture according to the exposure context.

Because existing RAW datasets usually cover only one exposure subtask~\cite{chen2018learning,fu2023raw,hanji2022sihdr}, we collect Complex Multi-Exposure RAW (CMER), with 223 real scenes and 1,784 pairs spanning both directions, and synthesize additional pre-training pairs from MIT-Adobe FiveK~\cite{bychkovsky2011learning}. CMER supplies pixel-aligned supervision for unified evaluation, while the synthetic data tests whether broader exposure coverage benefits real-data fine-tuning. Extensive experiments show consistent improvements under both training protocols and exposure directions.
In summary, the main contributions are summarized as follows:
\begin{itemize}[noitemsep,topsep=2pt]
    \interlinepenalty=10000 
    \item We establish a unified RAW-domain formulation for simultaneous exposure correction and restoration. By addressing divergent degradations, our framework provides a holistic solution to the RAW-domain multi-exposure errors within a single model.
    \item We propose two physically motivated components: a linear illumination correction network for exposure-aware linear gain estimation from channel-specific RAW cues, and a frequency-based enhancement network with dynamic convolution for adaptive restoration of divergent frequency degradations.
    \item We construct CMER, a paired real RAW benchmark with 223 scenes and 1,784 multi-exposure pairs for unified RAW-based exposure correction and restoration tasks under diverse lighting conditions.
\end{itemize}

\section{Related Work}
\label{sec:related-work}

\textit{Single Image High Dynamic Range Reconstruction:}
SIHDR expands local dynamic range through inverse tone mapping~\cite{banterle2009high}, direct CNN/Transform reconstruction~\cite{eilertsen2017hdr,marnerides2018expandnet,chen2023improving}, exposure-stack synthesis~\cite{lee2018deep}, or inverse-camera modeling~\cite{liu2020single,chen2021hdrunet}. RawHDR further exploits high-bit-depth RAW measurements~\cite{zou2023rawhdr}. These methods recover localized dynamic-range loss from a nominally exposed input and usually assume a usable global tone, whereas exposure correction must infer the global rescaling.

\textit{Low-Light Image Enhancement:}
Classical enhancement adjusts histograms, suppresses noise, or decomposes illumination and reflectance~\cite{tian2017global,hajiaboli2011anisotropic,zhang2018high}. Learning-based methods employ illumination decomposition, color spaces, frequency embedding, transformers, and Mamba architectures~\cite{li2023pixel,cai2023retinexformer,yan2025hvi,huang2025decouplenet,chen2025retinex,li2025noise,li2026statistical,zheng2024quad,liu2025pyramid}. RAW restoration additionally models sensor noise for synthesis and removal~\cite{Zou_2023_ICCV,Zou_2022_CVPR,wei2020physics,wei2021physics,zhang2026real}. These methods are effective for photon-limited inputs but typically assume a known exposure ratio and do not address clipping-dominated over-exposure.

\textit{Image Exposure Correction:}
Classical exposure correction uses histogram, tone, or Retinex adjustment~\cite{tian2017global,ke2010bita,fu2016weighted}, while high-light-specific networks recover brightness and clipped content~\cite{liu2022attention,cao2020over,fu2023raw}. 
Deep learning has seen rapid development recently~\cite{tian2023transformer,fu2022reinforcement,fang2023dual,zhang2024deep2}, unified sRGB methods learn coarse-to-fine correction, exposure consistency, luminance, color-space, or task-shared mappings~\cite{afifi2021learning,huang2022deep,huang2022exposure-consistency,huang2023learning,baek2023luminance,jung2025exposure,li2024color,yang2023learning,li2023difficulty}. Guided reconstruction can exploit frequency cues or cross-modal structure~\cite{hu2022bi,chen2024frequency,zhang2025unaligned,li2025fcdfusion,zhang2026enhancing,li2023joint,zhang2026supervise}, but does not connect them to RAW exposure formation. Existing RAW methods still separate low-light restoration~\cite{huang2022towards,wei2021physics,chen2018learning} and high-light recovery~\cite{fu2023raw}; RIECR instead couples RAW-linear gain estimation with exposure-adaptive residual restoration.

\begin{figure}[t]
    \scriptsize
    \setlength\tabcolsep{2.5 pt}
    \resizebox{\linewidth}{!}{%
    \begin{tabular}{ccc}
    \centering
    \includegraphics[width=0.31\linewidth]{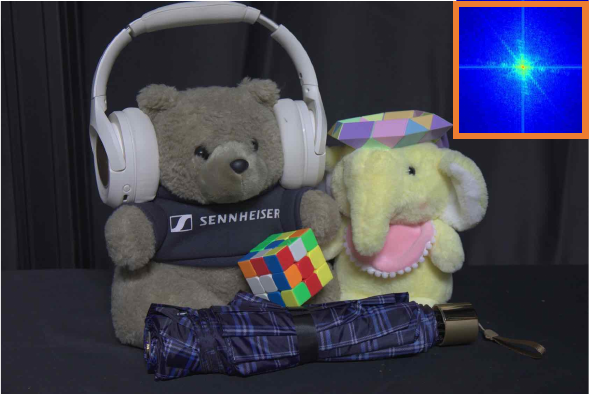} &
    \includegraphics[width=0.31\linewidth]{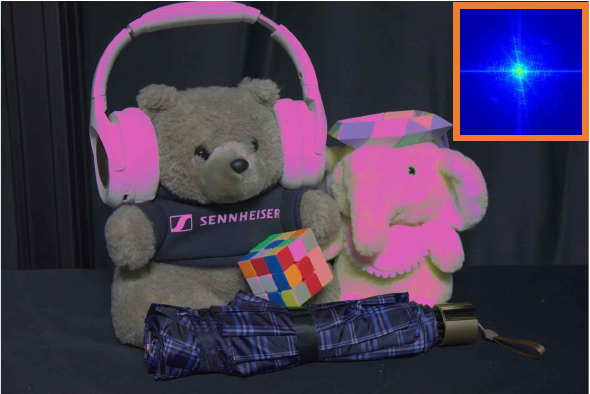} &
    \includegraphics[width=0.31\linewidth]{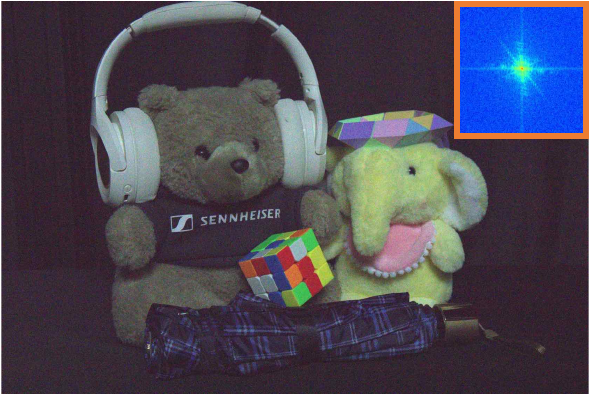} \\
    (a) Reference & (b) Linearly Scaled (High.) & (c) Linearly Scaled (Low.) \\
    \includegraphics[width=0.31\linewidth]{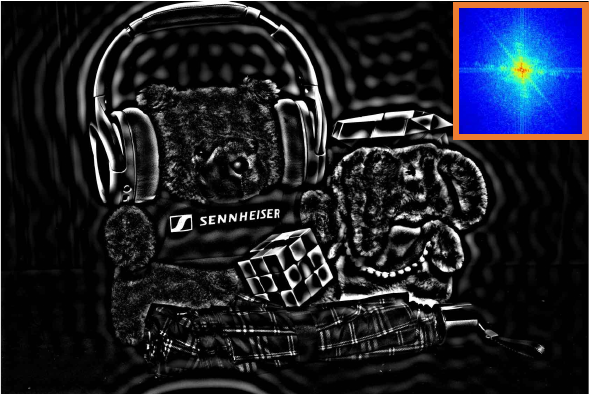} &
    \includegraphics[width=0.31\linewidth]{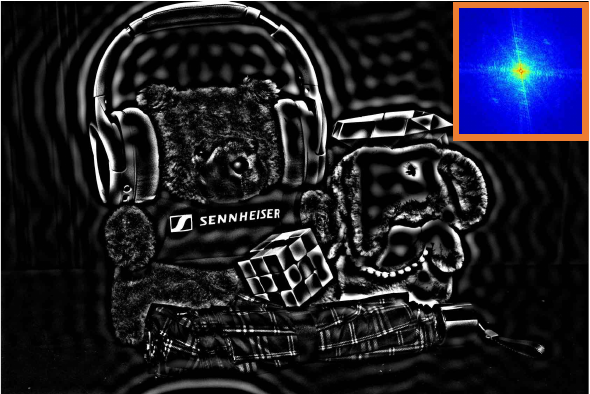} &
    \includegraphics[width=0.31\linewidth]{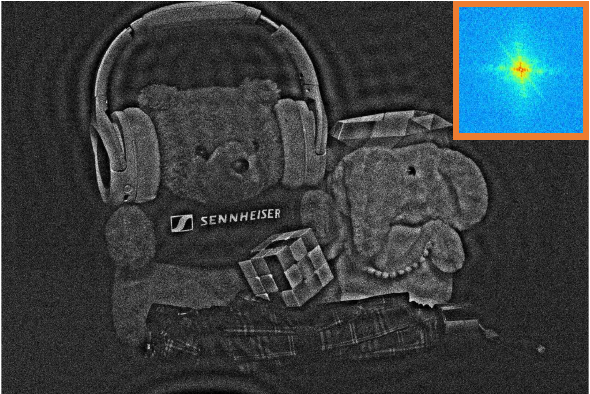} \\
    (d) Reference \textit{HF} & (e) High Light \textit{HF} & (f) Low Light \textit{HF} \\
    \includegraphics[width=0.31\linewidth]{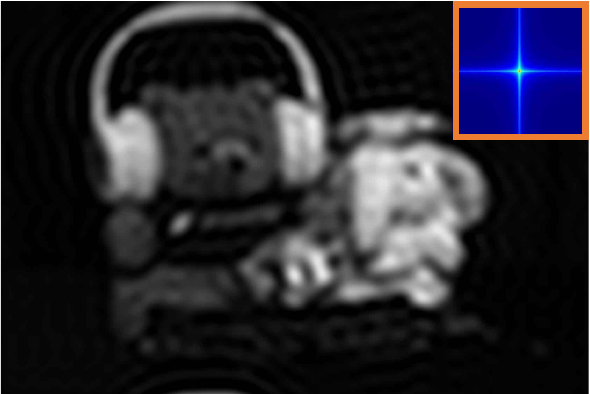} &
    \includegraphics[width=0.31\linewidth]{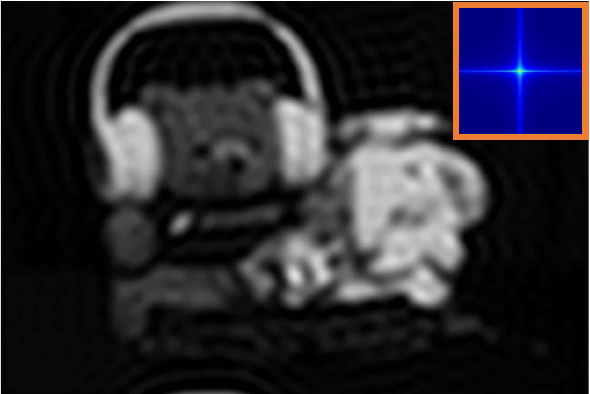} &
    \includegraphics[width=0.31\linewidth]{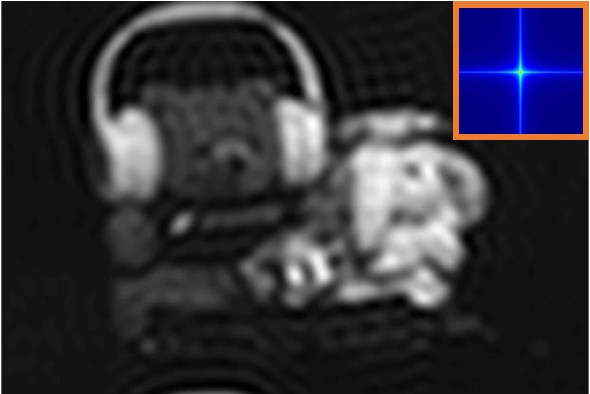} \\
    (g) Reference \textit{LF} & (h) High Light \textit{LF} & (i) Low Light \textit{LF}
    \end{tabular}}
    \caption{Frequency analysis after linear scaling. (a--c) Scaled images, (d--f) high-frequency (\textit{HF}) components, and (g--i) low-frequency (\textit{LF}) components. Over-exposure removes edges, whereas under-exposure amplifies noise.
Under different lighting conditions, the low-frequency information is similar to that in the reference image, indicating that information such as illumination remains essentially unchanged after linear adjustment.}
    \label{fig:fft comparison}
\end{figure}

\begin{figure*}[htbp]
  \centering
  \includegraphics[width=1\linewidth]{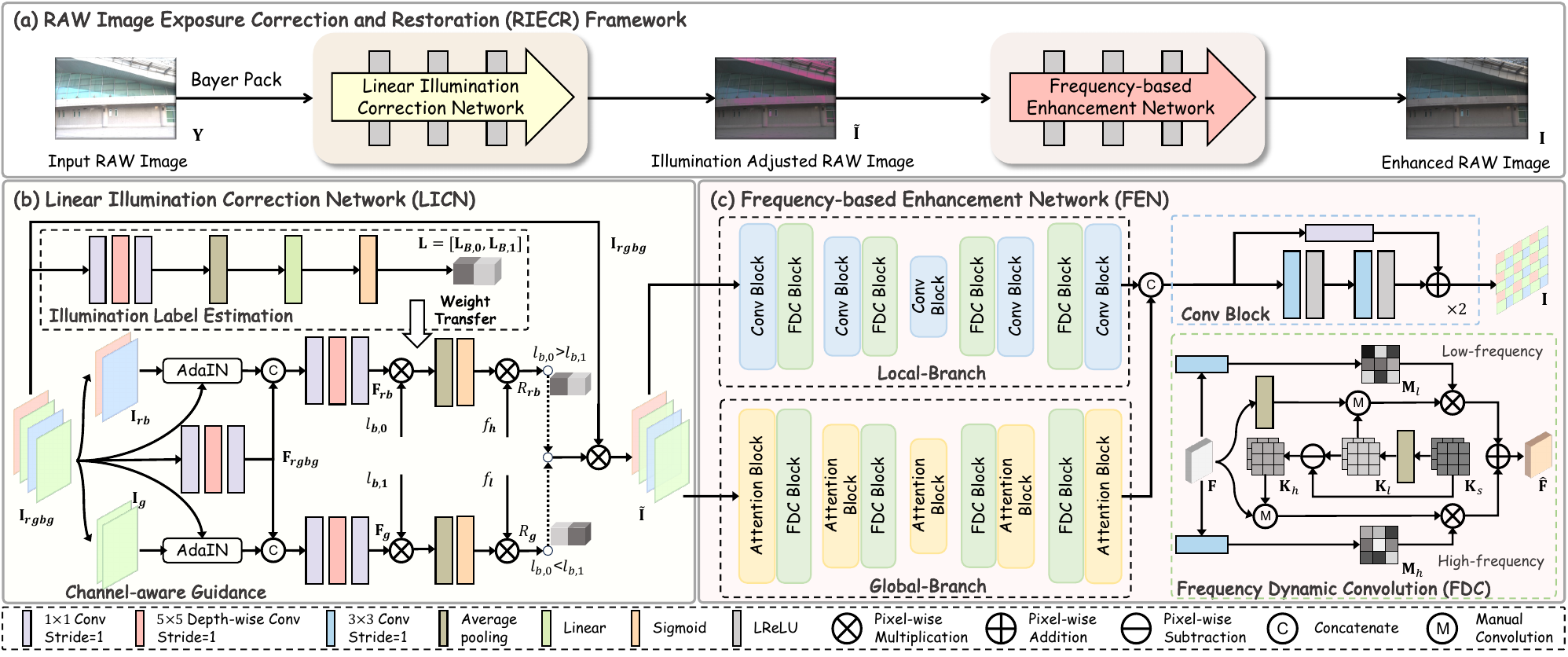}

   \caption{The overview framework of the proposed method (RIECR), including Linear Illumination Correction Network (LICN) to adjust the illumination and Frequency-based Enhancement Network (FEN) to address the divergent degradation.}
   \label{fig:framework}
\end{figure*}

\section{The Proposed Method}
\label{sec:method}

\subsection{Problem Formulation}
\label{sec: formulation}

Let $\mathbf{X}$ be the clean normal-exposure RAW image and $\mathbf{Y}$ an observed RAW image. The linear sensor response:
\begin{equation}
	\mathbf{Y} = \begin{cases}
				r\mathbf{X} + \mathbf{N}=r(\mathbf{X}+\tilde{\mathbf{N}}), & \text{if }0 < r < 1 \\
				\text{min}(r\mathbf{X}, 1)=r \cdot \text{min}(\mathbf{X}, \frac{1}{r}), & \text{if }r \ge 1
			\end{cases}
			\label{eq: formula}
\end{equation}
where $r$ is the exposure ratio, $\mathbf{N}$ contains photon and sensor noise, and $\min(\cdot,1)$ denotes saturation. Noise and clipping make inverse scaling insufficient.

For low-light inputs ($0<r<1$), inverse scaling amplifies the normalized noise $\tilde{\mathbf{N}}$:
\begin{equation}
	\mathbf{X}=\frac{\mathbf{Y}}{r}-\tilde{\mathbf{N}}.
	\label{eq: low-light formula}
\end{equation}
For high-light inputs ($r\geq1$), let $\mathbf{S}$ denote the structural and textural signal lost by clipping. Restoration becomes:
\begin{equation}
	\begin{aligned}
		\mathbf{Y}&=r(\mathbf{X}-\mathbf{S}),\\
		\mathbf{X}&=\frac{\mathbf{Y}}{r}+\mathbf{S}.
	\end{aligned}
	\label{eq: high-light formula}
\end{equation}
Amplified noise and missing structure have different causes but mainly affect high-frequency residuals, while illumination are more stable. Both cases can be written as:
\begin{equation}
	\mathbf{X} = \frac{\mathbf{Y}}{r} + \mathbf{D},
	\label{eq: unified}
\end{equation}
where $r$ is the gain and $\mathbf{D}$ is $-\tilde{\mathbf{N}}$ for low-light denoising or $\mathbf{S}$ for high-light recovery.

\subsection{Motivation and Framework Architecture}
\label{sec: motivation}

Fig.~\ref{fig:fft comparison} illustrates the frequency behavior. It motivates two coupled modules. LICN estimates $r$ using channel-dependent RAW responses: Green has higher sensitivity and SNR~\cite{yang2015automated,zhang2024deep,zou2023rawhdr}, while Red/Blue often retain highlight responses. FEN restores $\mathbf{D}$ by adapting low-/high-frequency responses to amplified noise or missing texture.

The architecture of the proposed RAW Image Exposure Correction and Restoration (RIECR) framework is illustrated
in Fig.~\ref{fig:framework}. A normalized Bayer image $\mathbf{Y}\in\mathbb{R}^{2H\times2W}$ is packed as $\mathbf{I}_{rgbg}\in\mathbb{R}^{H\times W\times4}$ and mapped by:
\begin{equation}
	\mathbf{I}=f(\mathbf{I}_{rgbg};\theta),
\end{equation}
where $f$ and $\theta$ denote RIECR and its parameters. LICN produces a linearly corrected $\tilde{\mathbf{I}}$; local and global FEN branches then predict the residual and output $\mathbf{I}$.

\subsection{Linear Illumination Correction Network (LICN)}
\label{sec: LICN}

LICN contains an exposure-label estimator and channel-aware gain guidance (Fig.~\ref{fig:framework}(b)).

\noindent{\textbf{Illumination Label Estimation.}} 
A $1\times1$ convolution mixes packed channels, a $5\times5$ depth-wise convolution aggregates local illumination, and global pooling with a linear projection:
\begin{equation}
	\mathbf{L} = \text{Sigmoid}(\text{Linear}(\text{Avg.}(\text{Convs.}(\mathbf{I}_{rgbg})))),
\end{equation}
where $\mathbf{L}=[\mathbf{L}_{high},\mathbf{L}_{low}]$. The exposure-time ratio $f_{ratio}$ and maximum factors $f_h,f_l$ define severity-aware targets $\hat{\mathbf{L}}=[f_{ratio}/f_h,0]$ for high-light and $[0,f_{ratio}/f_l]$ for low-light, supervised by:
\begin{equation}
	\mathcal{L}_{label} = -\hat{\mathbf{L}}\cdot\text{log}(\mathbf{L})-(1-\hat{\mathbf{L}})\cdot\text{log}(1-\mathbf{L}),
\end{equation}
which supervises exposure direction and magnitude.

\noindent{\textbf{Channel-Aware Guidance.}} 
We split $\mathbf{I}_{rgbg}$ into Green $\mathbf{I}_{g}$ and Red/Blue $\mathbf{I}_{rb}$. AdaIN~\cite{Huang_2017_ICCV} normalizes each cue before fusion with the shared feature $\mathbf{F}_{rgbg}$:
\begin{equation}
	\mathbf{F}_{t} = \text{Convs}(\text{Concat}[\mathbf{F}_{rgbg}, \text{AdaIN}(\mathbf{I}_{t})]),
\end{equation}
where $t\in\{rb,g\}$. The label-weighted gains are:
\begin{align}
	R_{rb} &= \text{max}(\text{Sigmoid}(\text{Avg.}(\mathbf{F}_{rb}\cdot \mathbf{L}_{high}))\cdot f_h, 1), \\
	R_{g} &= \text{max}(\text{Sigmoid}(\text{Avg.}(\mathbf{F}_{g}\cdot \mathbf{L}_{low}))\cdot f_l, 1),
\end{align}
where $R_{rb}$ and $R_g$ are darkening and brightening ratios. The adjusted image is:
\begin{equation}
	\tilde{\mathbf{I}} = \begin{cases}
				\mathbf{I}_{rgbg} / R_{rb}, & \text{if } \mathbf{L}_{high} > \mathbf{L}_{low} \\
				\mathbf{I}_{rgbg} \cdot R_{g}, & \text{if } \mathbf{L}_{high} \le \mathbf{L}_{low}
			\end{cases}
\end{equation}
so the predicted state selects the RAW cue and linear operator before residual restoration.

\begin{table*}[htbp]
  \centering
  \small
  \caption{Comparison of datasets for exposure correction and restoration tasks.}
    \label{tab:dataset comparison}
  \renewcommand{\arraystretch}{1.0}
	\setlength\tabcolsep{4.7pt}
	\resizebox{\textwidth}{!}{%
    \begin{tabular}{c|c|c|c|c|c|c|c|c|c}
    \toprule
    \multirow{2}{*}{Task}            & \multirow{2}{*}{Dataset}            & \multirow{2}{*}{Format}    & \multirow{2}{*}{Type} & \multirow{2}{*}{Size} & \multirow{2}{*}{High} & \multirow{2}{*}{\makecell[c]{High-light \\ Exposure Value}}  & \multirow{2}{*}{Low} & \multirow{2}{*}{\makecell[c]{Low-light \\ Exposure Value}} &  \multirow{2}{*}{\makecell[c]{Normal-light \\ Ground-Truth}}\\
     &             &                            &                       &             &             &          &                &              \\
    \midrule
    \multirow{3}{*}{Under-Exposure Correction} & LOL~\cite{wei2018deep}& sRGB                       & Real                  & 789   & \ding{56}    & \emph{N/A}     & \ding{51}     & \emph{N/A}   & \ding{51}    \\ 
    & LOL-v2~\cite{wei2018deep}& sRGB                       & Real                  & 500   & \ding{56}    & \emph{N/A}     & \ding{51}     & \emph{N/A}    & \ding{51}    \\                  
    (RAW Image Denoising) & SID~\cite{chen2018learning}                  & RAW                  & Real                   & 2,697    & \ding{56}     &   \emph{N/A}     & \ding{51}     &    Ratio=$\frac{1}{100}$,$\frac{1}{250}$,$\frac{1}{300}$   & \ding{51}   \\ \midrule
    Over-Exposure Correction & RPO~\cite{fu2023raw}                  & RAW                  & Real                   & 2,600    & \ding{51}     & Ratio=3, 5, 8, 10       & \ding{56}     &    \emph{N/A}   & \ding{51}   \\ \midrule
    HDR Reconstruction & SIHDR~\cite{hanji2022sihdr} & RAW & Real & 905 & \ding{51} & \emph{N/A} & \ding{51} & \emph{N/A}   & \ding{56}  \\ \midrule
	\multirow{3}{*}{Multiple-Exposure Correction} & SICE~\cite{cai2018learning}& sRGB                       & Real                & 4,413   & \ding{51}    & EV=0.5,0.7,1,2,3     & \ding{51}     & EV=-0.5,-0.7,-1,-2,-3  & \ding{51}       \\
	& ME~\cite{afifi2021learning} & sRGB                       & Syn.        & 24,330    & \ding{51}     & EV=1,1.5  & \ding{51}     & EV=-1,-1.5          & \ding{51}            \\ \rowcolor{graycolor}
	& \textbf{CMER (Ours)}                  & RAW                  & Real             & 1,784    & \ding{51}        & Ratio=2,4,8        & \ding{51}     & Ratio=$\frac{1}{2}$,$\frac{1}{4}$,$\frac{1}{8}$,$\frac{1}{16}$,$\frac{1}{32}$    & \ding{51}   \\\bottomrule
    \end{tabular}}
\end{table*}

\subsection{Frequency-Based Enhancement Network (FEN)}
\label{sec: FEN}

FEN uses local and global branches with FDC (Fig.~\ref{fig:framework}(c)). The local branch exploits spatial self-similarity, while the global branch supplies context for severely degraded regions.

\noindent{\textbf{Frequency Dynamic Convolution.}} 
Fixed convolution cannot simultaneously denoise and recover detail across spatially varying degradation~\cite{hu2022bi,chen2024frequency}. FDC decomposes a base kernel $\mathbf{K}_s$ in the spatial domain:
\begin{equation}
	\begin{aligned}
		\mathbf{K}_l &= \text{Avg}(\mathbf{K}_s), \\
		\mathbf{K}_h &= \mathbf{K}_s - \mathbf{K}_l,
	\end{aligned}
\end{equation}
where $\text{Avg}(\cdot)$ is $3\times3$ pooling. $\mathbf{K}_l$ captures stable structure and $\mathbf{K}_h$ captures residual detail. Pixel-wise gates $\mathbf{M}_l,\mathbf{M}_h$:
\begin{equation}
	\hat{\mathbf{F}} = \mathbf{M}_h \odot \text{MC}(\mathbf{K}_h, \mathbf{F}) + \mathbf{M}_l \odot \text{MC}(\mathbf{K}_l, \text{Avg}(\mathbf{F})),
\end{equation}
where $\text{MC}$ is the primitive convolution and $\odot$ denotes multiplication. The gates attenuate noise-exposed high frequencies or strengthen contextual texture near saturated structures.

\noindent{\textbf{Local Convolution Branch.}} A U-shaped CNN uses residual blocks:
\begin{equation}
	\hat{\mathbf{F}} = \{\text{LReLU}(\text{Conv}_{3\times3}(\mathbf{F}))\}_{\times2} + \text{Conv}_{1\times1}(\mathbf{F}),
\end{equation}
where $\text{LReLU}$ is the activation. Down/up-sampling and two skip connections predict a residual added to $\tilde{\mathbf{I}}$, yielding $\mathbf{I}_{local}$.

\noindent{\textbf{Global Attention Branch.}} A U-shaped transformer supplies long-range context~\cite{cai2023retinexformer}. Its $i$-th attention block is:
\begin{equation}
	\begin{aligned}
		\hat{\mathbf{F}}_{i} &= \text{CMCA}(\text{LN}(\mathbf{F}_{i-1})) + \mathbf{F}_{i-1}, \\
		\mathbf{F}_{i} &= \text{FFN}(\text{LN}(\hat{\mathbf{F}}_{i})) + \hat{\mathbf{F}}_{i},
	\end{aligned}
\end{equation}
where CMCA, LN~\cite{ba2016layer}, and FFN denote channel attention, layer normalization, and feed-forward layers. Multi-scale FDC blocks let reliable regions guide distant degradation; convolution fuses $\mathbf{I}_{local}$ and $\mathbf{I}_{global}$ into $\mathbf{I}$.

\subsection{Learning Details}

We scale each reference by its exposure ratio to obtain the LICN target $\mathbf{I}_l$:
\begin{equation}
	\mathcal{L}_{illu} = \|\tilde{\mathbf{I}}-\mathbf{I}_l\|_{1}.
\end{equation}

The total loss is:
\begin{equation}
	\mathcal{L} = \mathcal{L}_{r} + \gamma_{1}\mathcal{L}_{illu} + \gamma_{2}\mathcal{L}_{label},
\end{equation}
where $\mathcal{L}_{r}=\|\mathbf{I}-\mathbf{X}\|_1$ and $\gamma_1,\gamma_2$ balance the illumination and label terms.

\section{Multiple Exposure RAW Dataset}
\label{sec:dataset}

Existing RAW datasets target low-light denoising, high-light recovery, or HDR reconstruction, while multi-exposure datasets are mainly sRGB~\cite{chen2018learning,fu2023raw,hanji2022sihdr,afifi2021learning,zhu2025trim}. CMER provides paired RAW data for both directions (Table~\ref{tab:dataset comparison}).

\begin{figure}[t]
    \includegraphics[width=1\linewidth]{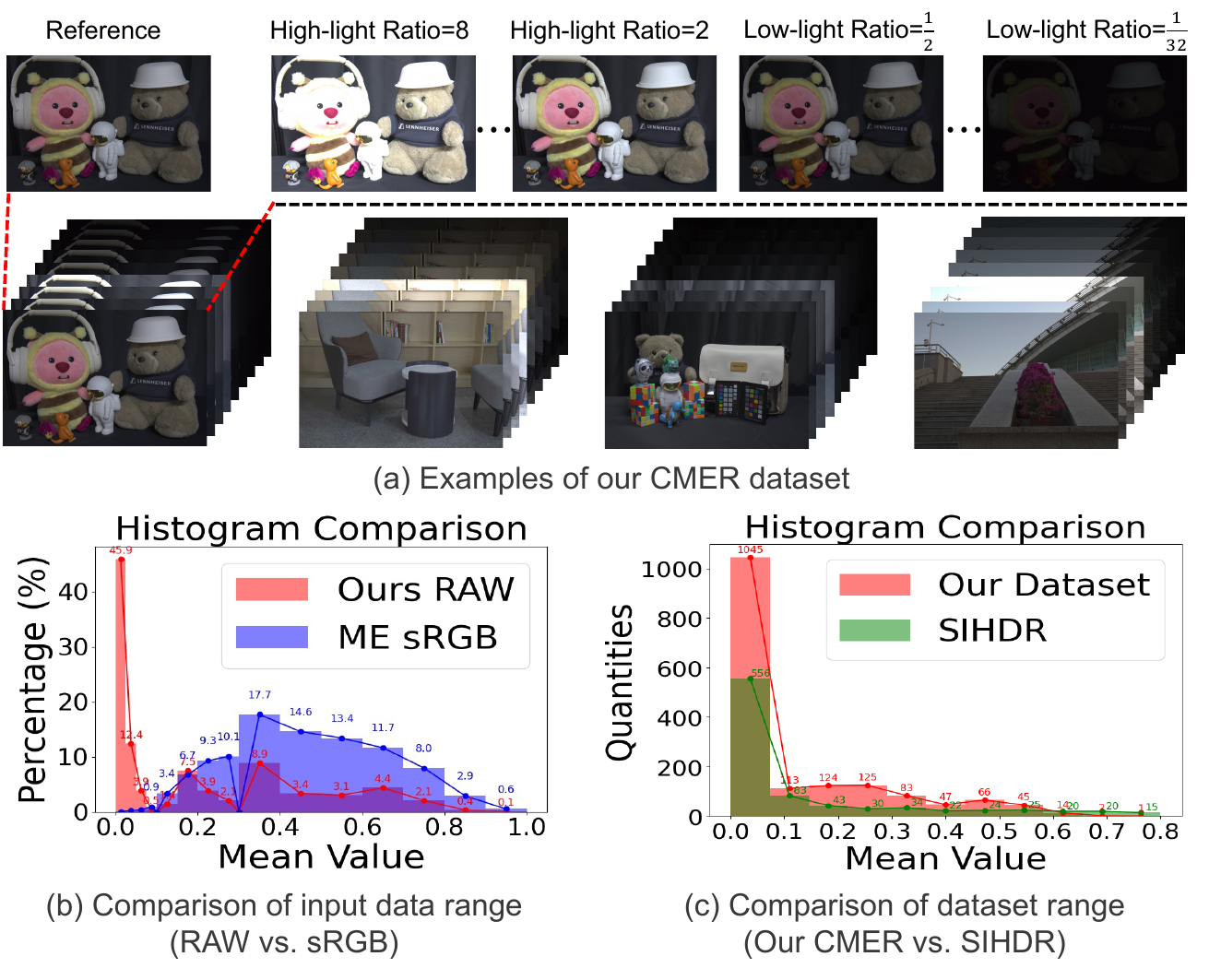}
    \caption{Overview and analysis of the proposed CMER dataset. (a) Paired Samples: Representative scenes demonstrating rigorous alignment across extreme exposure ratios (from $1/32$ to $8$) and their normal-light ground truth. (b) Data Integrity: Histogram comparison confirming that the RAW data preserves the original linear response compared to processed sRGB formats~\cite{afifi2021learning}. (c) Dataset Diversity: Statistical comparison showing CMER provides broader scene coverage than the existing SIHDR~\cite{hanji2022sihdr}.}
    \label{fig:data}
\end{figure}

\textit{Complex Multiple Exposure RAW (CMER) Dataset:}
We capture 223 static indoor/outdoor scenes with a Canon EOS 5D Mark IV and save 14-bit RAW images. The scenes include dark smooth regions, strong light sources, reflective materials, and fine structures. A tripod and remote control fix the viewpoint; after acquiring a well-exposed reference, we fix aperture and vary only exposure time. Captures with motion, illumination variation, or misalignment are discarded, preserving pixel alignment.
Daytime images use ISO 100; nighttime inputs additionally use ISO 1,600--6,400 while references remain at ISO 100. We record over-exposure ratios \{2,4,8\} and under-exposure ratios \{$1/2$,$1/4$,$1/8$,$1/16$,$1/32$\}~\cite{reinhard2010high}. CMER contains 1,784 pairs at $6720\times4480$: 669 over-exposed and 1,115 under-exposed. A scene-disjoint split uses 213 scenes (1,704 pairs) for training and 10 scenes (80 pairs) for testing.
Data samples and comparative data analysis are shown in Fig.~\ref{fig:data}.

\begin{table*}[t]
    \centering
    \renewcommand{\arraystretch}{1.0} 
    \setlength{\tabcolsep}{3.2pt}     
    \small                       
    
    \caption{Results on the disjoint CMER test split under two training protocols. Parameters and MACs are reported in M and G, respectively, with MACs calculated on a $256\times256$ RAW patch. Best and second-best restoration scores are \red{bold} and \blue{underlined}.}
    \label{tab: results on dataset}

    \resizebox{\textwidth}{!}{%
    \begin{tabular}{l | cccccc | cccccc | cc}
        \toprule
        \multirow{3}{*}{\textbf{Methods}} & \multicolumn{6}{c|}{\textbf{CMER (train) $\rightarrow$ CMER (test)}} & \multicolumn{6}{c|}{\textbf{SMIT(pre) $\rightarrow$ CMER(ft) $\rightarrow$ CMER(test)}} & \multirow{3}{*}{\makecell{\textbf{Params.}\\\textbf{(M)}}} & \multirow{3}{*}{\makecell{\textbf{MACs}\\\textbf{(G)}}} \\
        \cmidrule(lr){2-7} \cmidrule(lr){8-13}
        & \multicolumn{2}{c}{High-light} & \multicolumn{2}{c}{Low-light} & \multicolumn{2}{c|}{Average} & \multicolumn{2}{c}{High-light} & \multicolumn{2}{c}{Low-light} & \multicolumn{2}{c|}{Average} & & \\
        & PSNR$\uparrow$ & SSIM$\uparrow$ & PSNR$\uparrow$ & SSIM$\uparrow$ & PSNR$\uparrow$ & SSIM$\uparrow$ & PSNR$\uparrow$ & SSIM$\uparrow$ & PSNR$\uparrow$ & SSIM$\uparrow$ & PSNR$\uparrow$ & SSIM$\uparrow$ & & \\
        \midrule
        	\rowcolor{lightgraycolor}
        	\multicolumn{15}{l}{\textit{sRGB-based Model}} \\
        
        MSEC~\cite{afifi2021learning}\scriptsize\gray{[CVPR'21]} & 
        26.72 & 0.875 & 25.23 & 0.824 & 25.79 & 0.843 & 
        25.72 & 0.857 & 26.46 & 0.836 & 26.18 & 0.844 & 7.02 & 10.57 \\ 
        
        ECCM~\cite{nsampi2021learning}\scriptsize\gray{[BMVC'21]} & 
        28.81 & 0.913 & \blue{31.84} & 0.920 & 30.07 & 0.917 & 
        30.92 & 0.943 & 31.65 & 0.926 & 31.38 & 0.933 & 5.39 & 9.55 \\
        
        ECLNet~\cite{huang2022exposure-consistency}\scriptsize\gray{[MM'22]} & 
        26.76 & 0.899 & 26.89 & 0.876 & 26.76 & 0.885 & 
        29.08 & 0.929 & 26.99 & 0.890 & 27.77 & 0.905 & 0.02 & 1.67 \\
        
        ENC~\cite{huang2022exposure}\scriptsize\gray{[CVPR'22]} & 
        30.27 & 0.940 & 31.30 & 0.910 & 30.91 & 0.921 & 
        30.74 & \blue{0.956} & 31.38 & 0.924 & 31.14 & 0.936 & 0.62 & 14.24 \\
        
%
        
        FECNet~\cite{huang2022deep}\scriptsize\gray{[ECCV'22]} & 
        \blue{30.47} & 0.908 & 31.35 & 0.915 & \blue{31.02} & 0.912 & 
        30.62 & 0.923 & 31.84 & 0.917 & 31.38 & 0.919 & 0.15 & 5.82 \\
        
        RF~\cite{cai2023retinexformer}\scriptsize\gray{[ICCV'23]} & 
        29.94 & 0.939 & 30.09 & 0.906 & 30.03 & 0.919 & 
        \blue{32.69} & 0.950 & \blue{32.51} & 0.924 & \blue{32.58} & 0.933 & 1.61 & 20.38 \\
        
        CSEC~\cite{li2024color}\scriptsize\gray{[CVPR'24]} & 
        28.92 & 0.908 & 29.80 & 0.850 & 29.47 & 0.872 & 
        29.41 & 0.927 & 30.43 & 0.864 & 30.05 & 0.887 & 6.06 & 30.62 \\
        
        ES~\cite{jung2025exposure}\scriptsize\gray{[CVPR'25]} & 
        28.87 & 0.853 & 28.92 & 0.888 & 28.91 & 0.875 & 
        30.75 & 0.923 & 30.62 & 0.903 & 30.67 & 0.911 & 1.91 & 14.34 \\
        
        	\midrule
        	\rowcolor{lightgraycolor}
        	\multicolumn{15}{l}{\textit{RAW-based Model}} \\
        
        SID~\cite{chen2018learning}\scriptsize\gray{[CVPR'18]} & 
        25.09 & 0.805 & 25.32 & 0.824 & 25.23 & 0.817 & 
        25.18 & 0.818 & 25.79 & 0.844 & 25.65 & 0.834 & 7.76 & 13.73 \\
        
        RED~\cite{lamba2021restoring}\scriptsize\gray{[CVPR'21]} & 
        25.43 & 0.859 & 24.37 & 0.798 & 24.77 & 0.806 & 
        26.06 & 0.865 & 24.61 & 0.813 & 25.16 & 0.832 & 0.78 & 0.97 \\
        
        ABF~\cite{dong2022abandoning}\scriptsize\gray{[CVPR'22]} & 
        25.74 & 0.819 & 30.94 & 0.920 & 29.00 & 0.898 & 
        29.00 & 0.910 & 30.88 & 0.922 & 30.18 & 0.917 & 15.01 & 22.56 \\
        
        CGNet~\cite{fu2023raw}\scriptsize\gray{[TCSVT'23]} & 
        25.46 & 0.858 & 27.80 & 0.867 & 26.92 & 0.863 & 
        27.99 & 0.852 & 29.65 & 0.834 & 29.03 & 0.841 & 17.67 & 46.24 \\
        
        RawHDR~\cite{zou2023rawhdr}\scriptsize\gray{[ICCV'23]} & 
        27.03 & 0.899 & 26.90 & 0.879 & 26.94 & 0.886 & 
        27.81 & 0.913 & 27.72 & 0.890 & 27.76 & 0.898 & 10.49 & 50.81 \\
        
        SMG~\cite{xu2023low}\scriptsize\gray{[CVPR'23]} & 
        29.07 & 0.867 & 30.07 & 0.898 & 29.69 & 0.886 & 
        28.01 & 0.860 & 30.82 & 0.909 & 29.77 & 0.891 & 15.42 & 42.64 \\
        
        ADU~\cite{huang2024real}\scriptsize\gray{[TOMM'24]} & 
        28.16 & 0.893 & 27.33 & 0.839 & 27.64 & 0.859 & 
        28.92 & 0.895 & 28.49 & 0.861 & 28.65 & 0.874 & 0.64 & 0.71 \\
        		
        	RRM~\cite{chen2025retinex}\scriptsize\gray{[TCSVT'25]} &
        	29.74 & \blue{0.941} & 30.22 & \blue{0.933} & 30.05 & \blue{0.936} &
        	30.49 & 0.942 & 32.08 & \blue{0.935} & 31.48 & \blue{0.938} & 7.65 & 32.06 \\ \rowcolor{graycolor}
        
        \textbf{RIECR (Ours)} & 
        \red{31.23} & \red{0.950} & \red{33.45} & \red{0.950} & \red{32.63} & \red{0.950} & 
        \red{33.65} & \red{0.965} & \red{33.98} & \red{0.941} & \red{33.85} & \red{0.951} & 2.56 & 31.91 \\
        
        \bottomrule
    \end{tabular}%
    }
\end{table*}

\begin{figure*}
\scriptsize
\centering
  \setlength{\tabcolsep}{0.9mm}
  \renewcommand{\arraystretch}{1}
\begin{tabular}{ccccccccc}
   \includegraphics[width=0.1\linewidth]{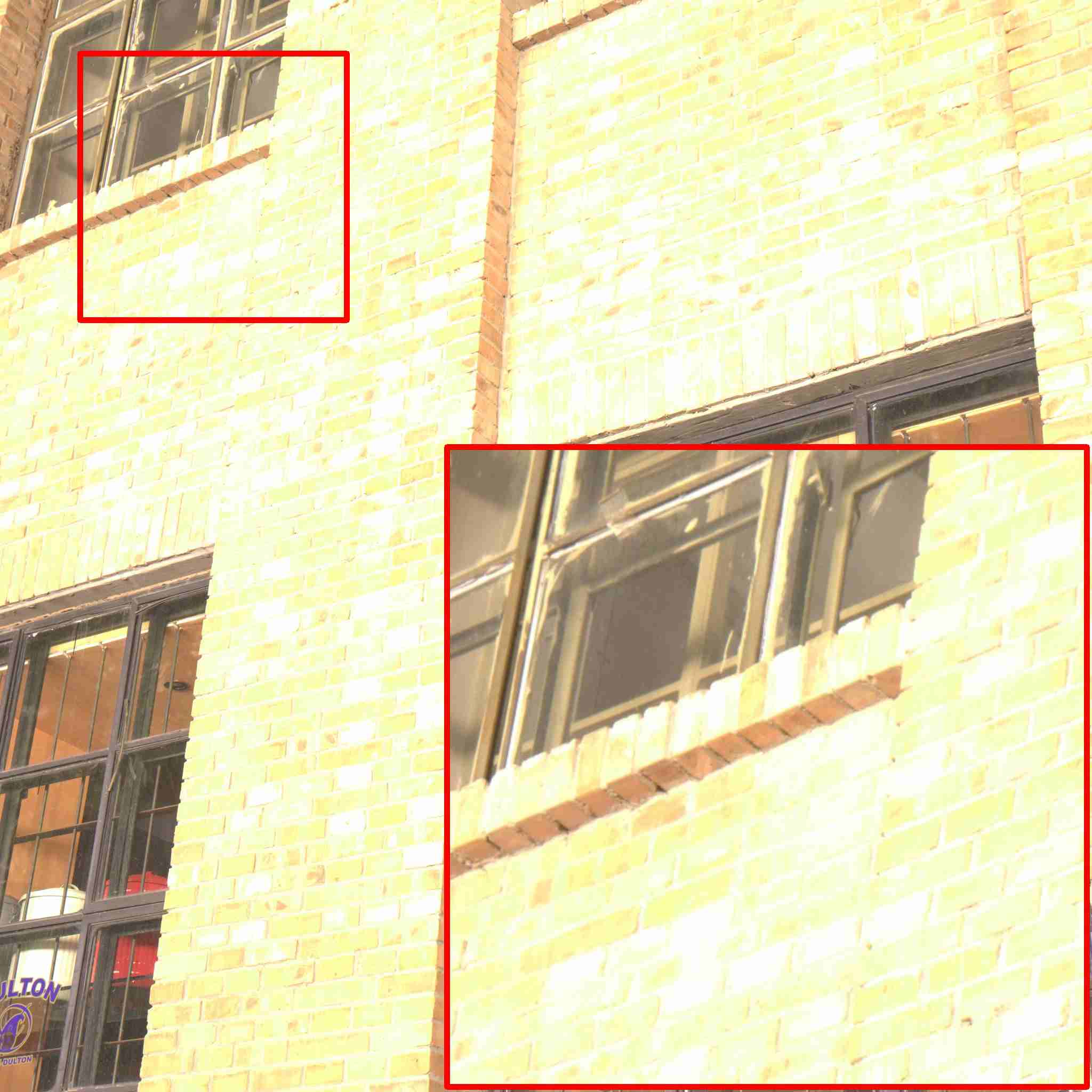}
  & \includegraphics[width=0.1\linewidth]{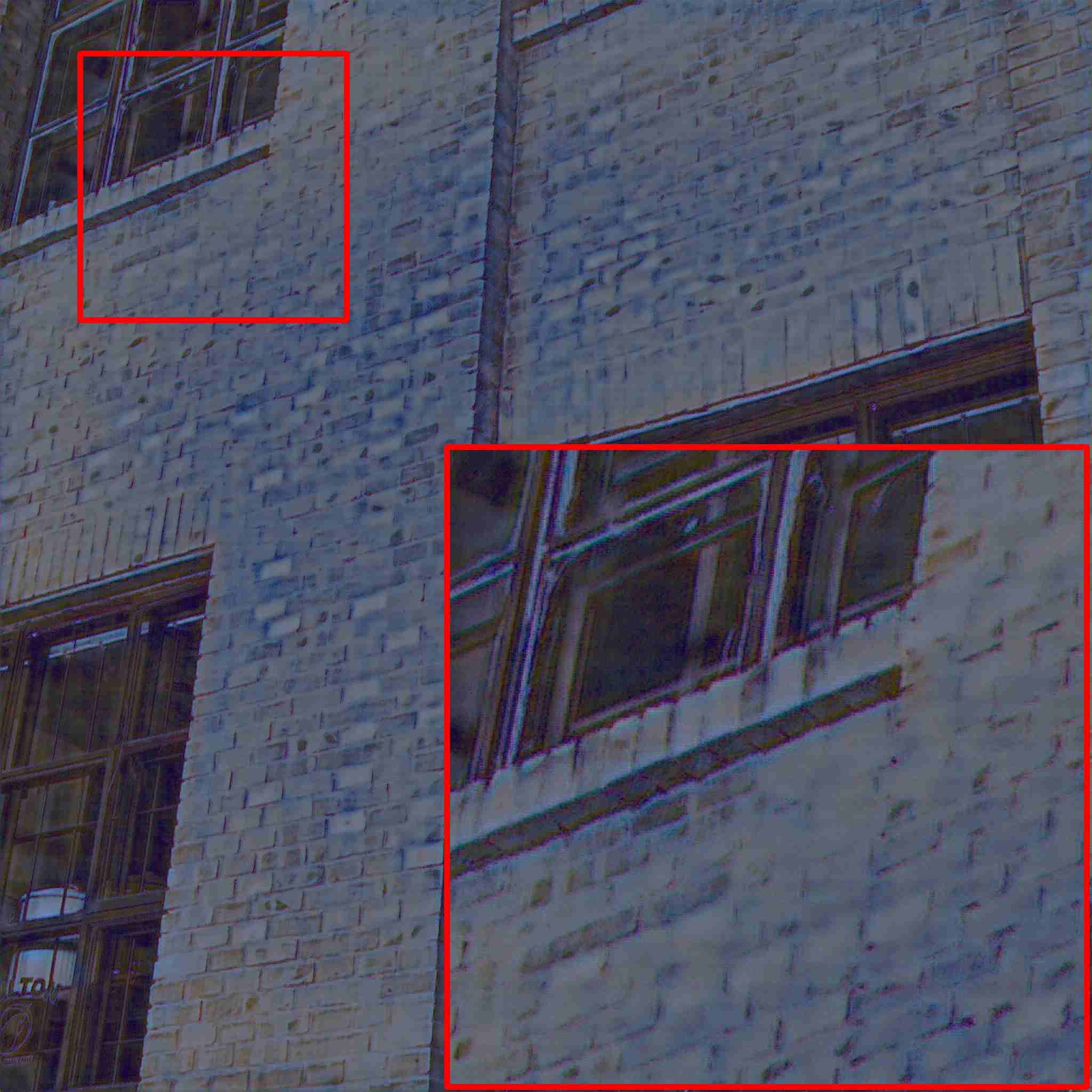}
  & \includegraphics[width=0.1\linewidth]{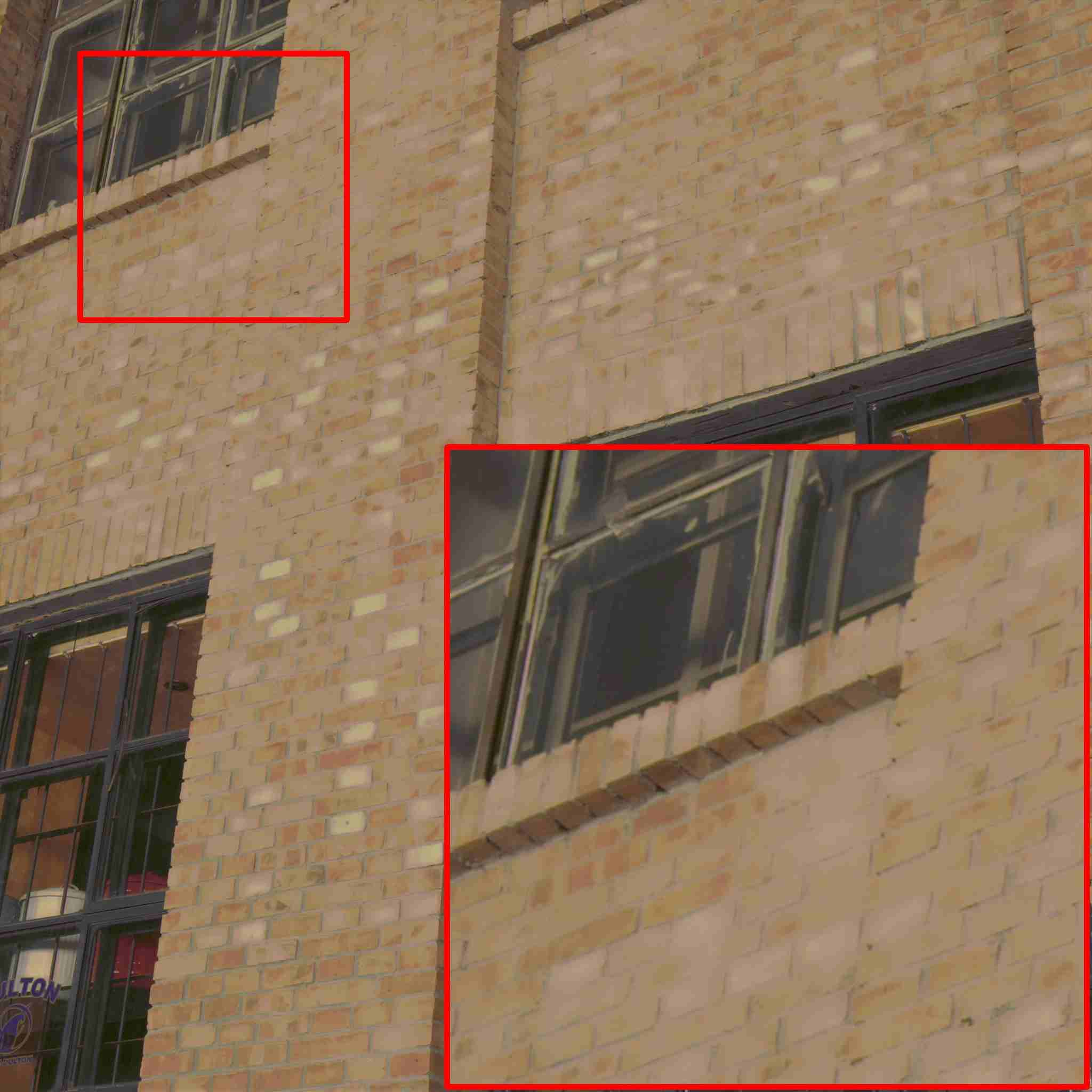}
  & \includegraphics[width=0.1\linewidth]{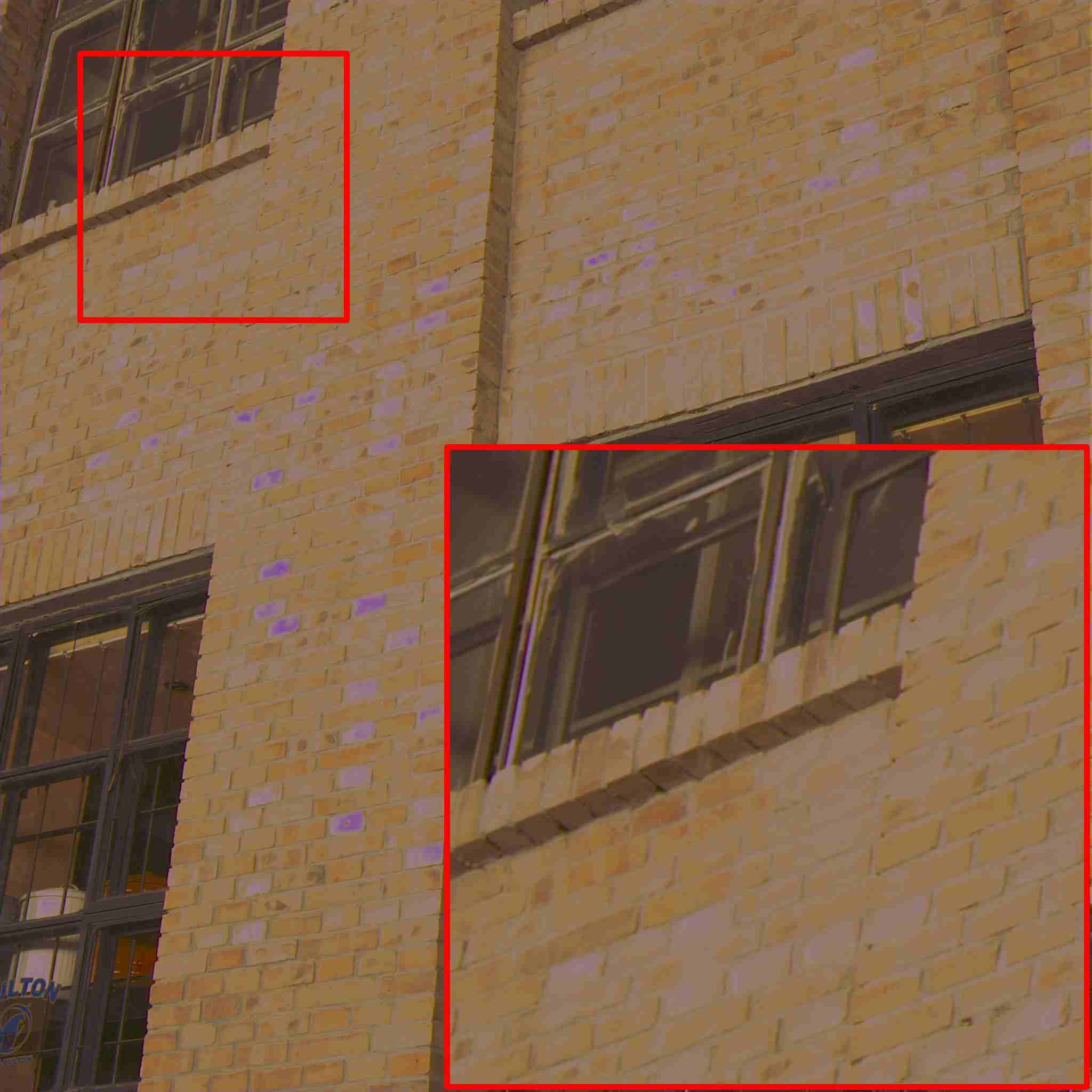}
  & \includegraphics[width=0.1\linewidth]{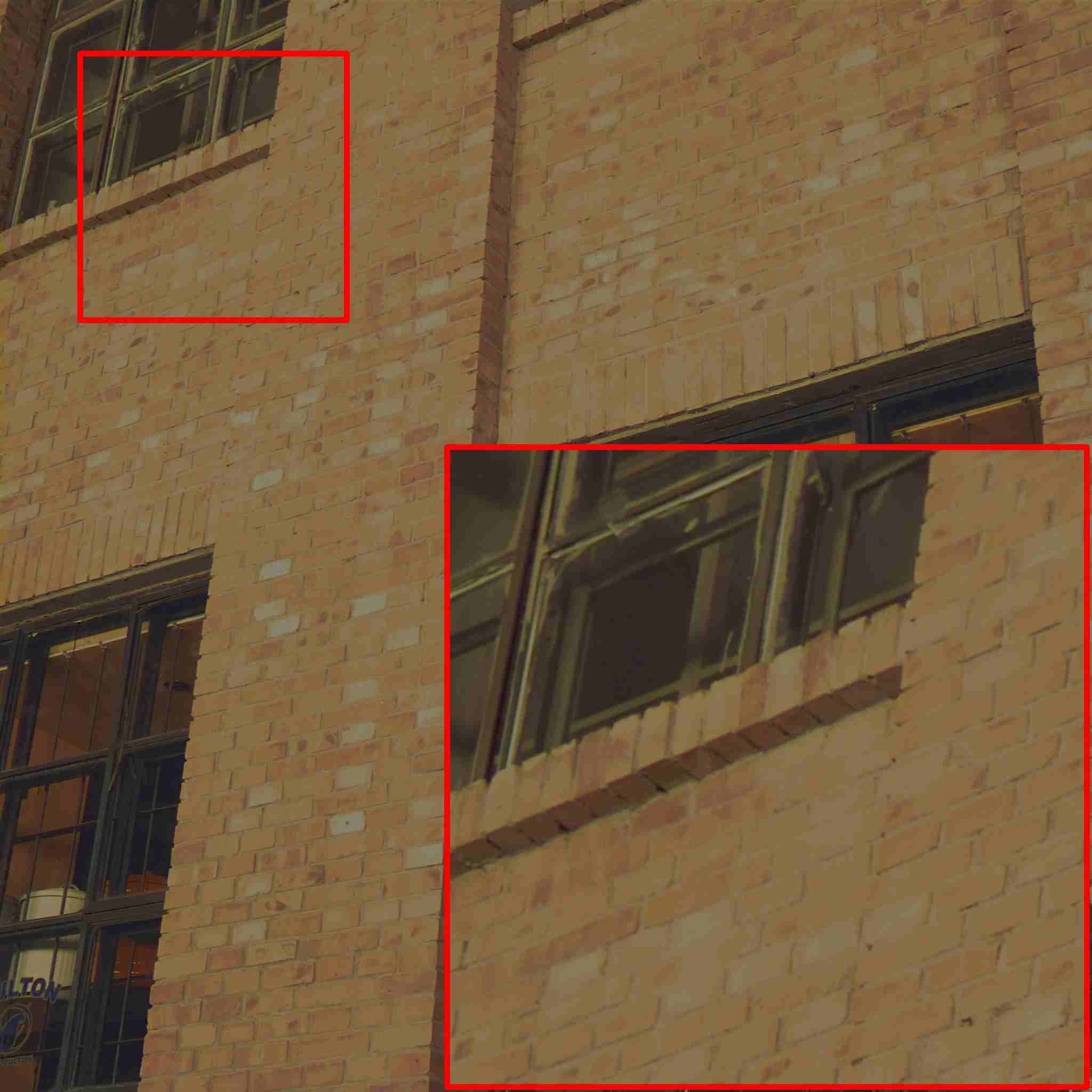}
  & \includegraphics[width=0.1\linewidth]{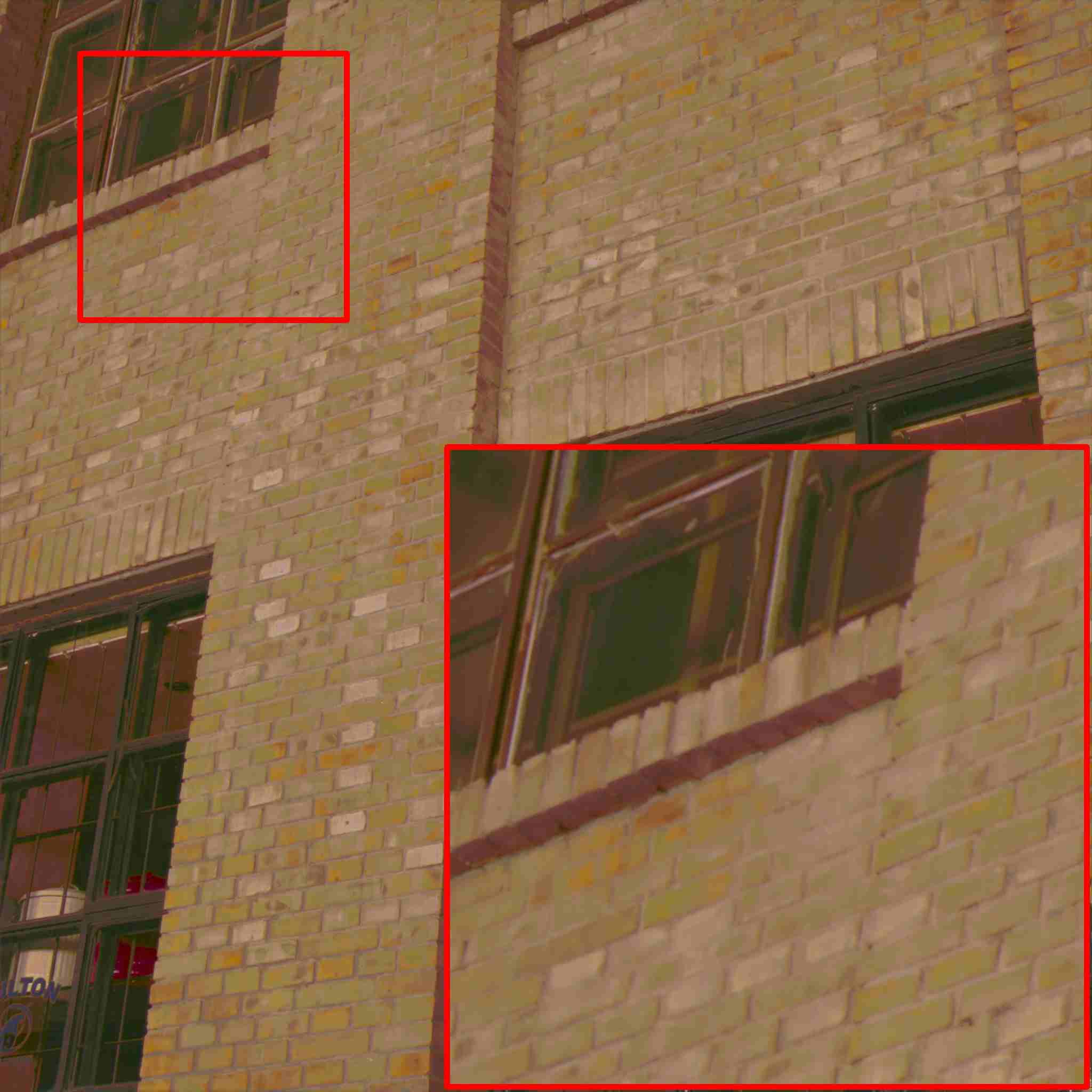}
  & \includegraphics[width=0.1\linewidth]{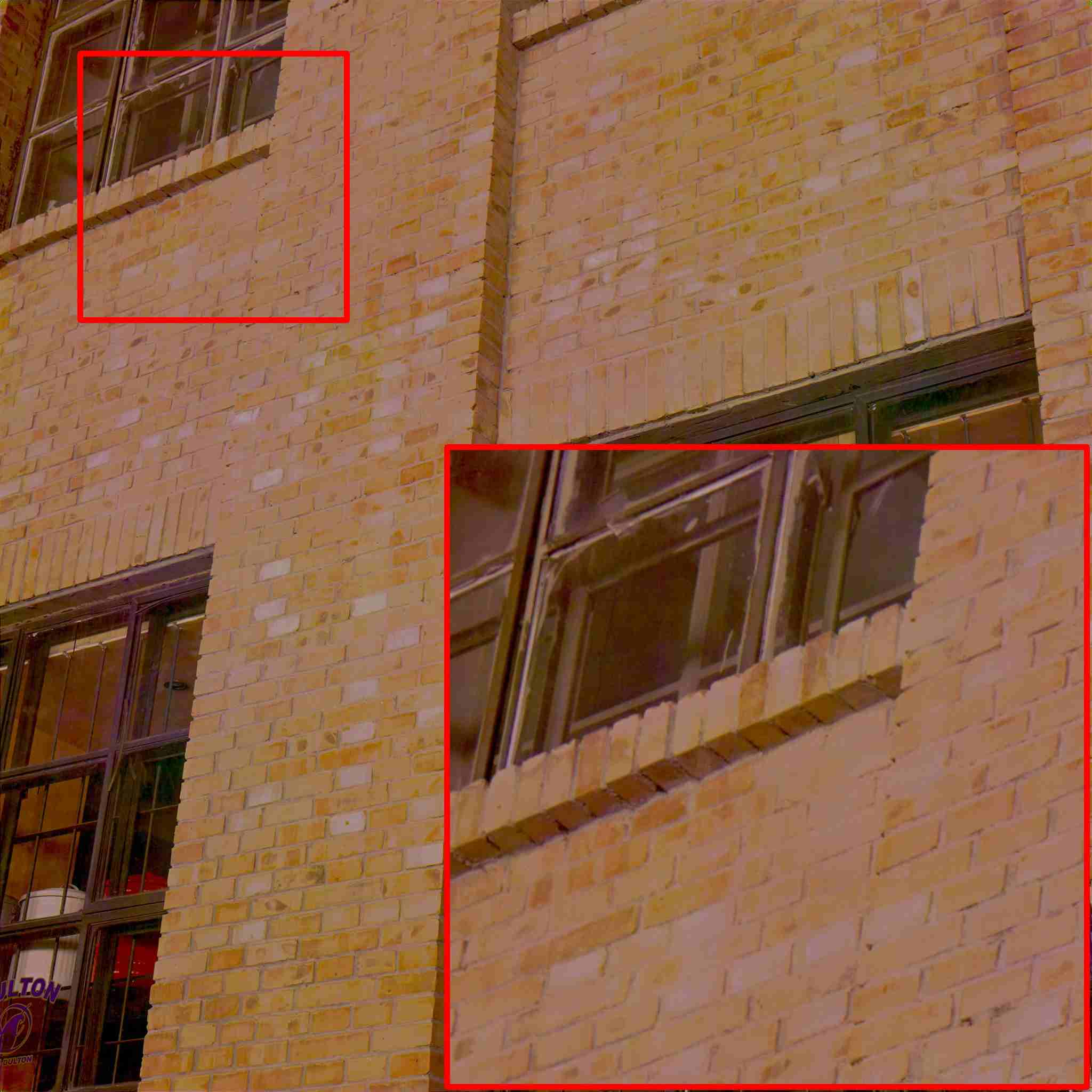}
  & \includegraphics[width=0.1\linewidth]{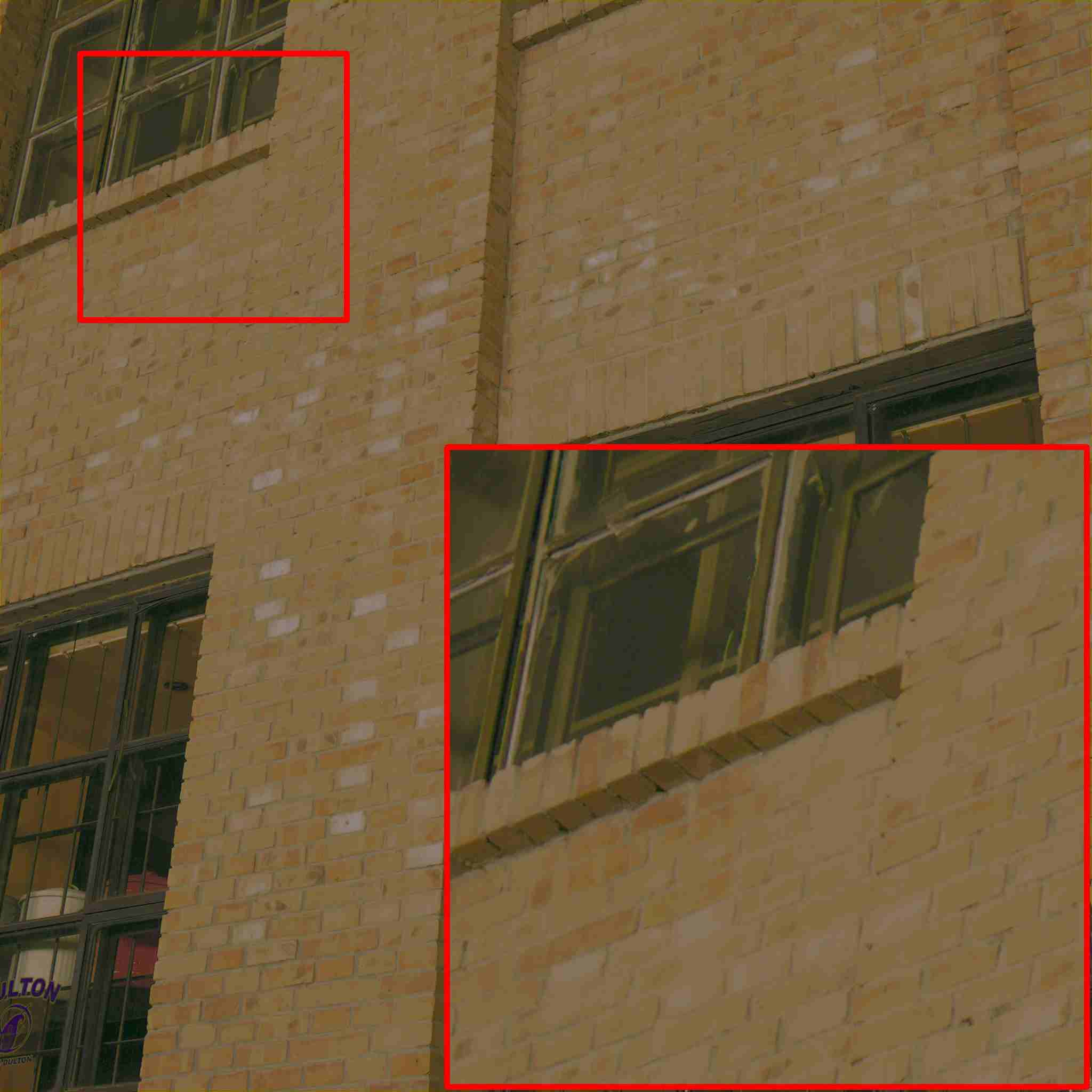}
  & \includegraphics[width=0.1\linewidth]{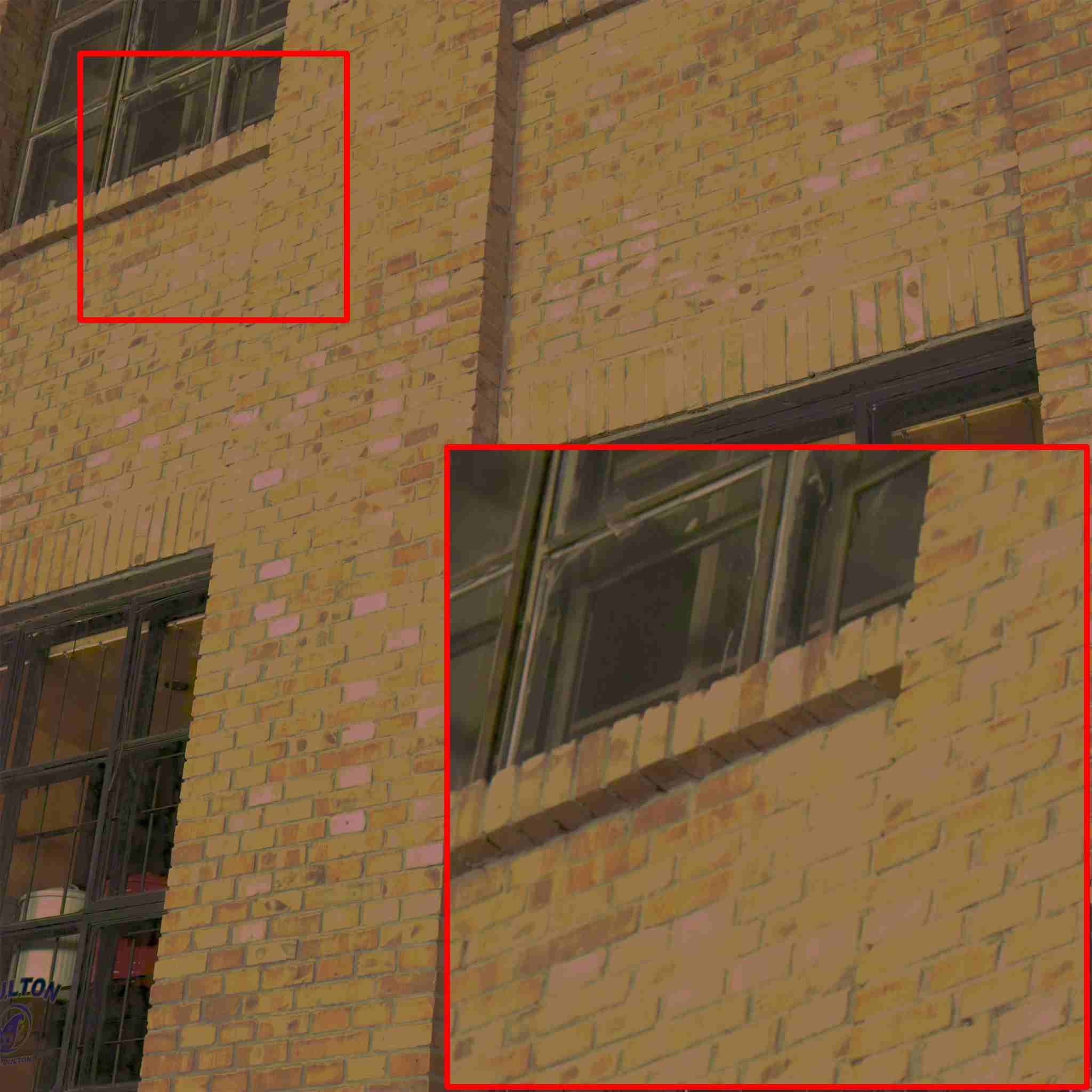}
   \\

Input  & MSEC~\cite{afifi2021learning} & ECCM~\cite{nsampi2021learning} & ECLNet~\cite{huang2022exposure-consistency}  & ENC-DRBN~\cite{huang2022exposure} & ENC-SID~\cite{huang2022exposure} & FECNet~\cite{huang2022deep} & RF~\cite{cai2023retinexformer} & CSEC~\cite{li2024color}  \\

  \includegraphics[width=0.1\linewidth]{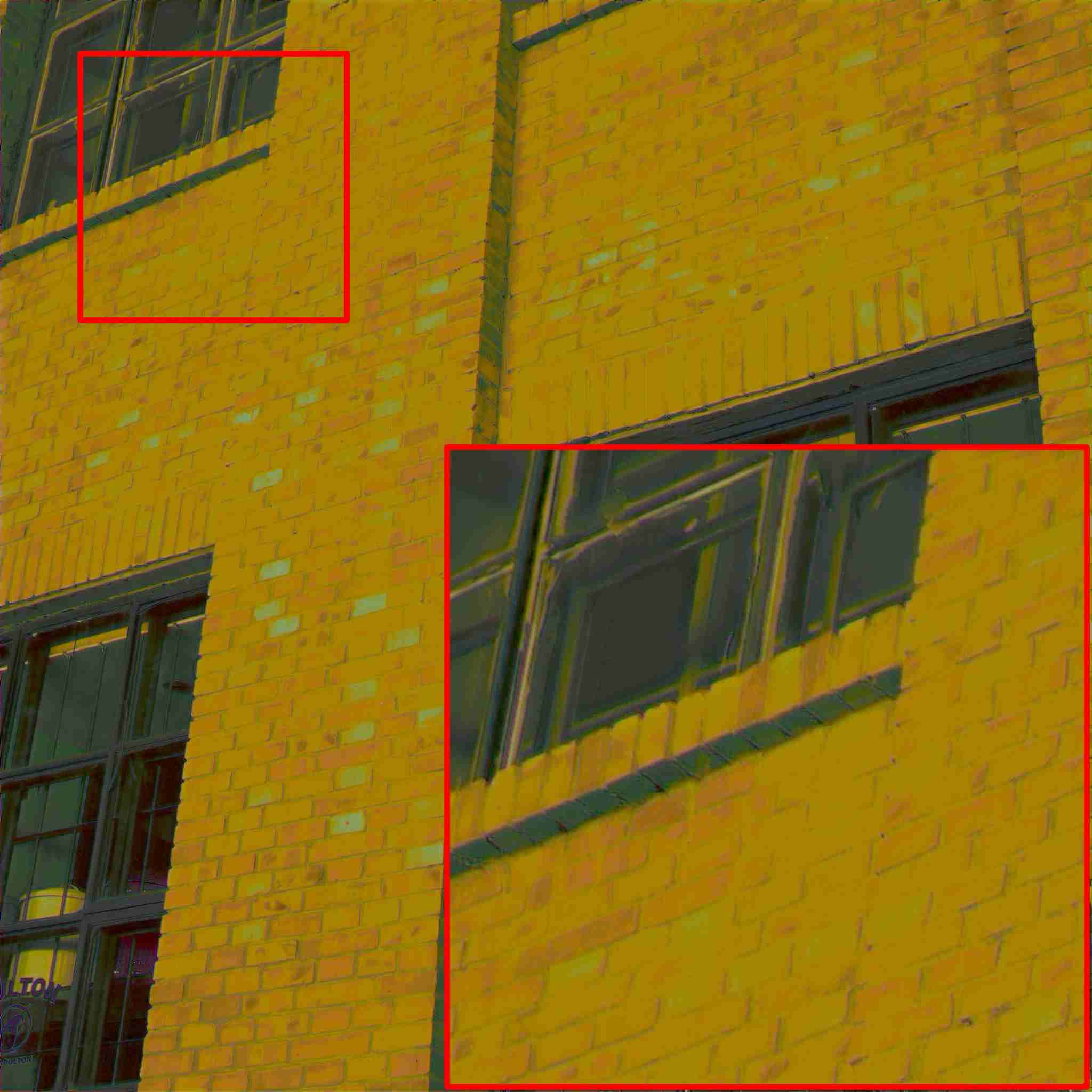}
  & \includegraphics[width=0.1\linewidth]{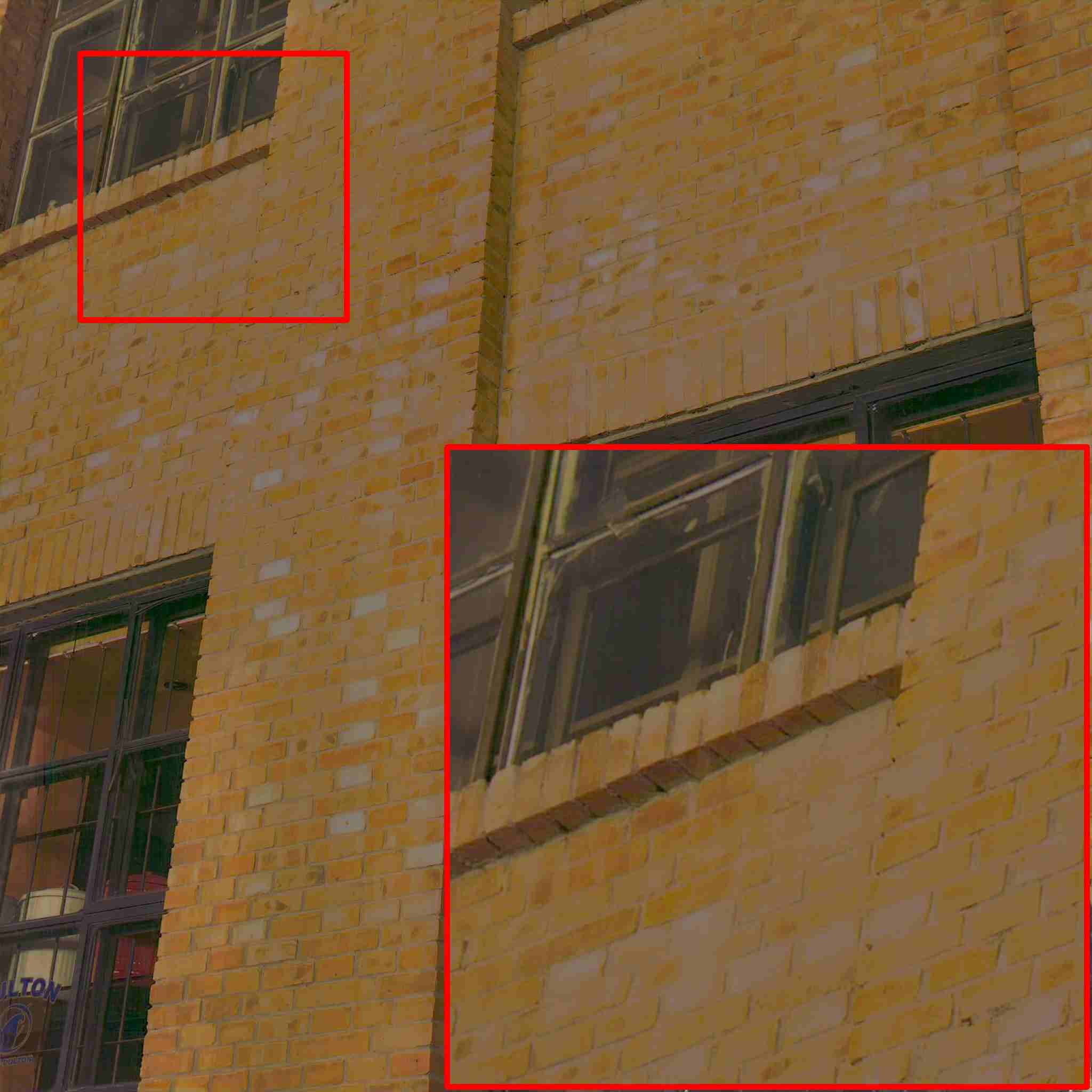}
  & \includegraphics[width=0.1\linewidth]{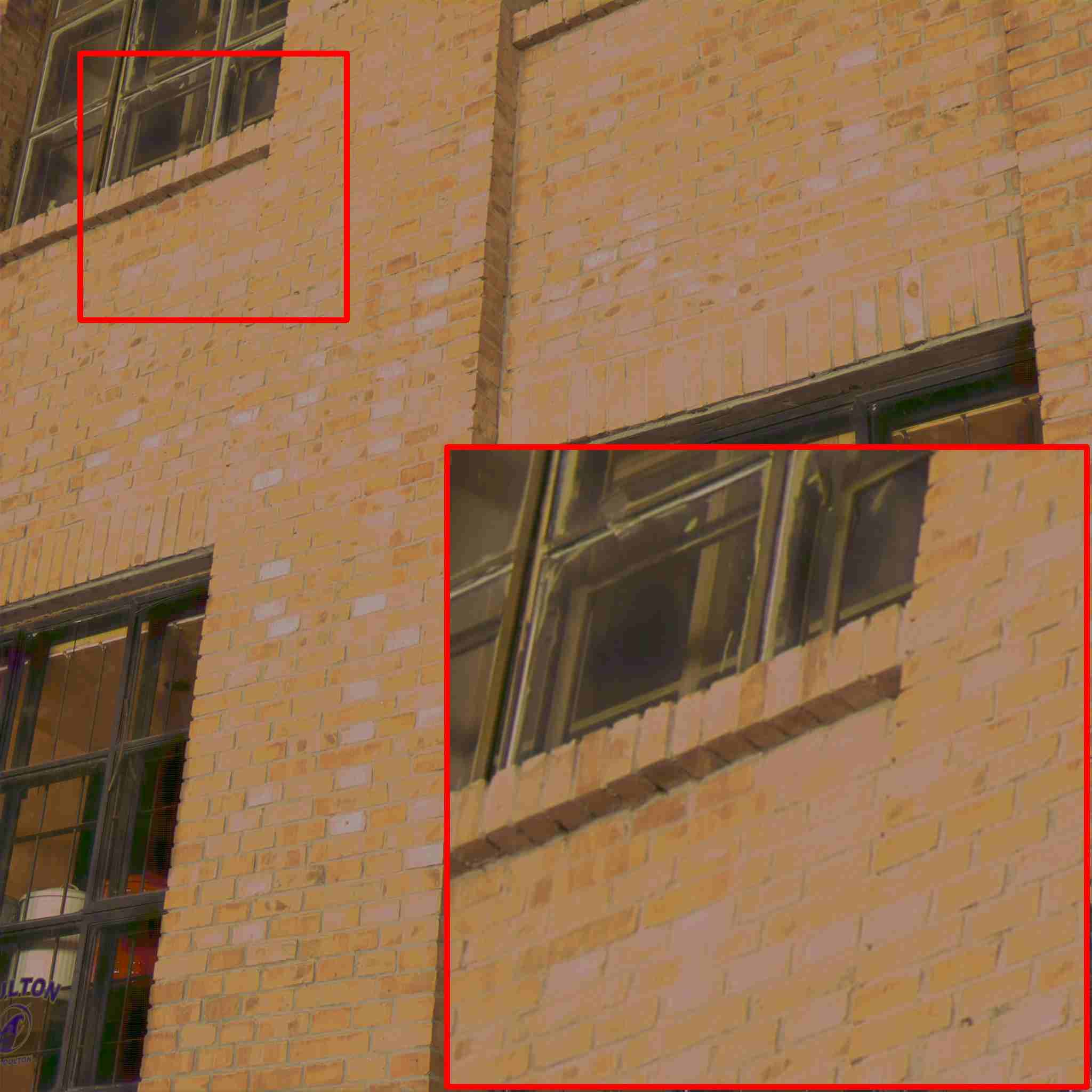}
  & \includegraphics[width=0.1\linewidth]{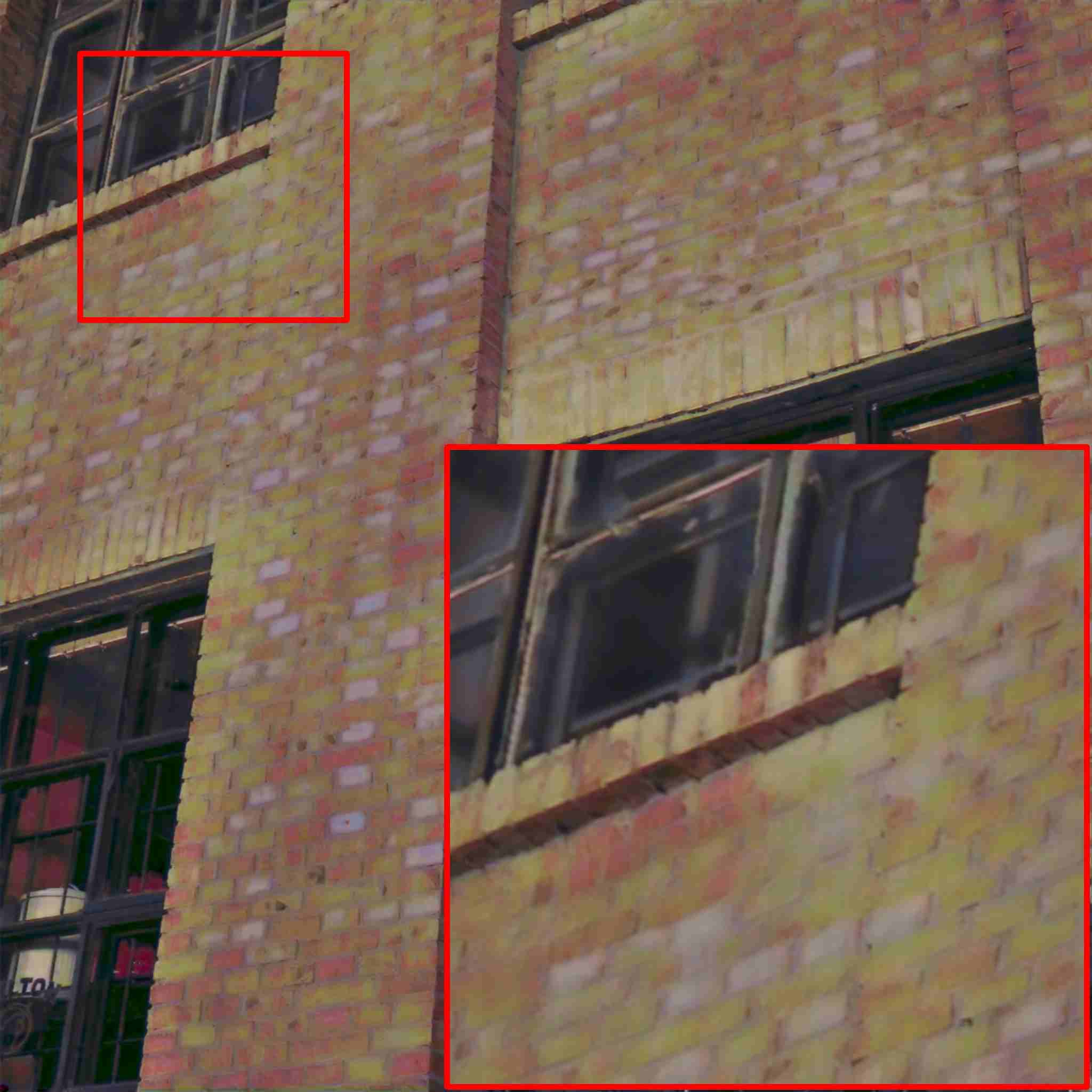}
  & \includegraphics[width=0.1\linewidth]{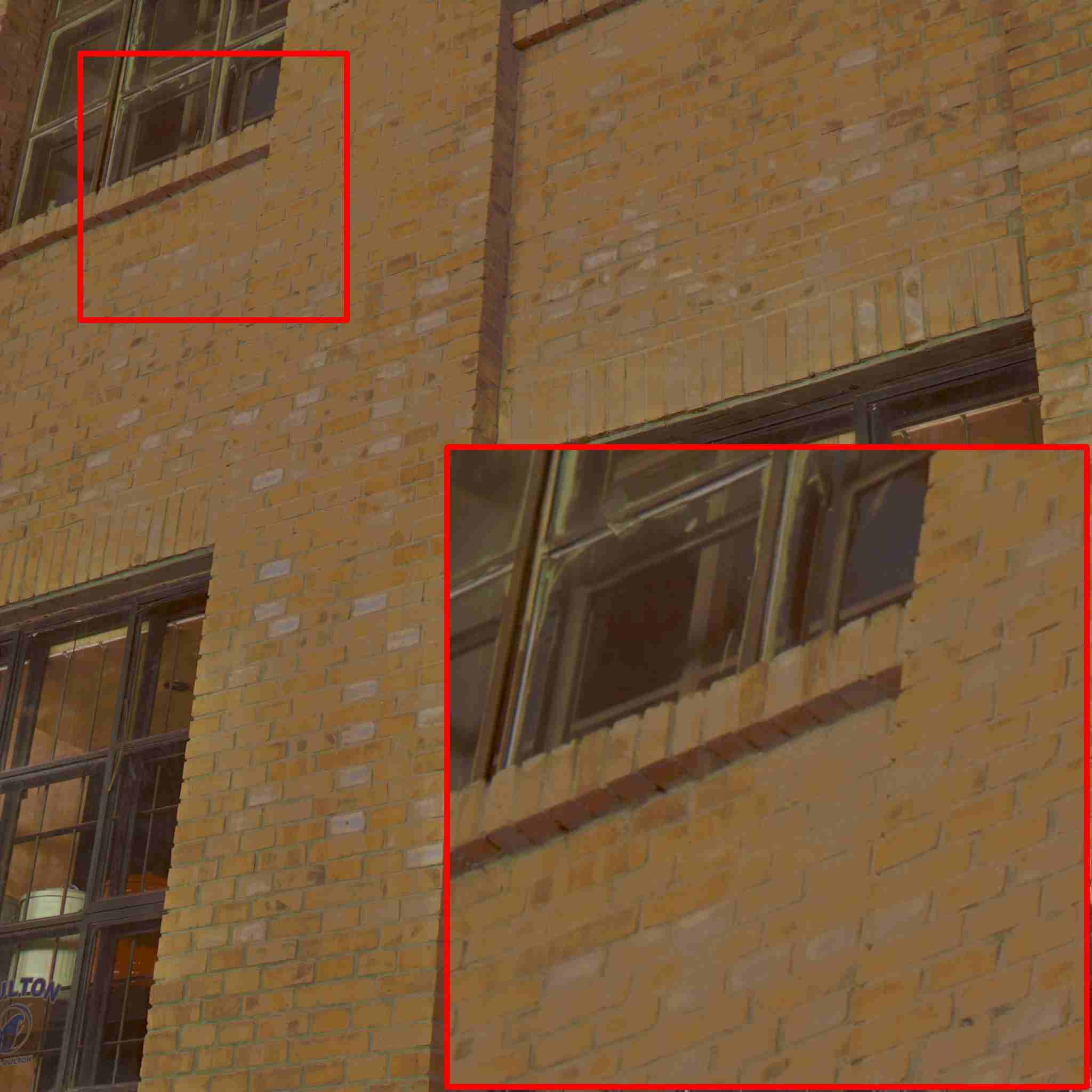}
  & \includegraphics[width=0.1\linewidth]{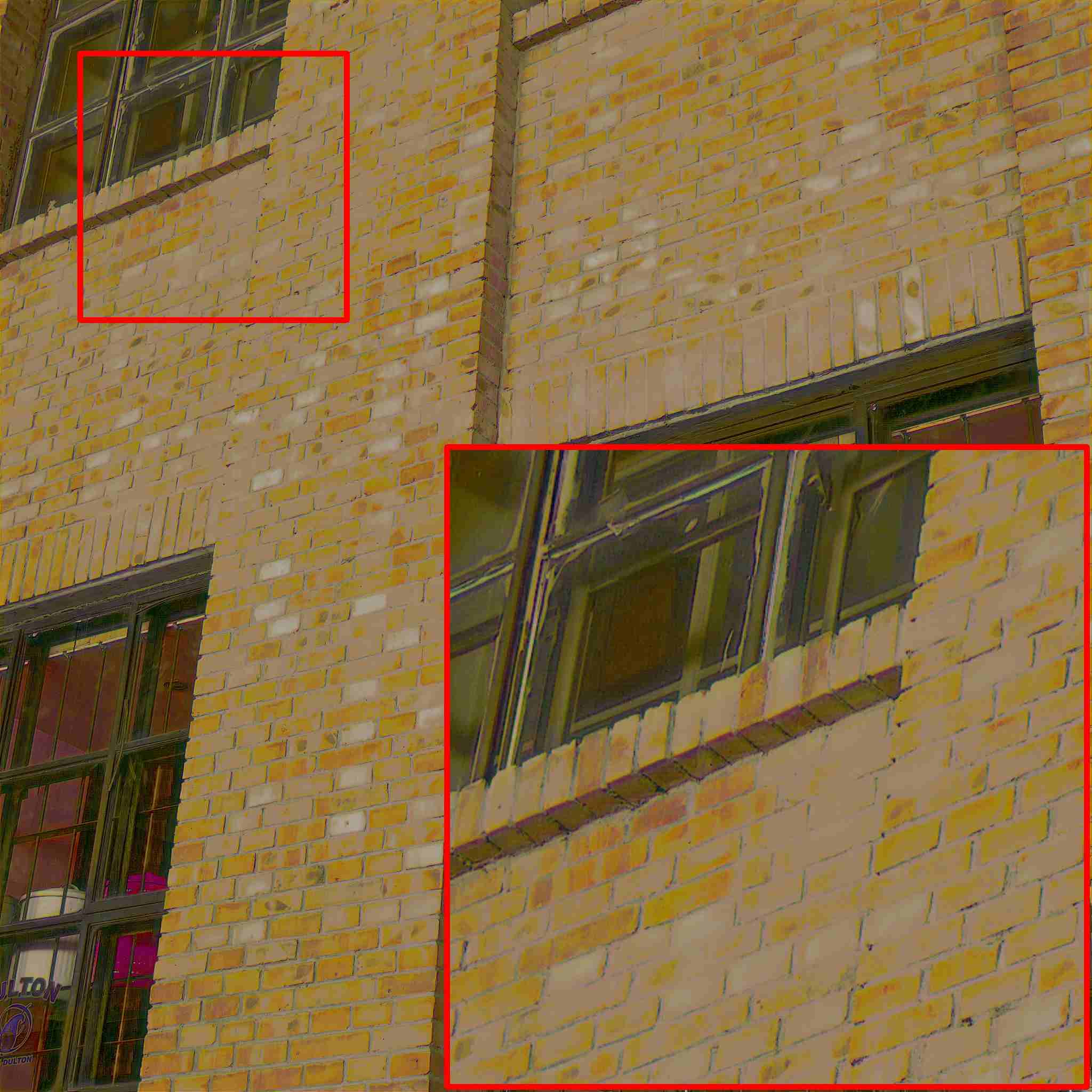}
  & \includegraphics[width=0.1\linewidth]{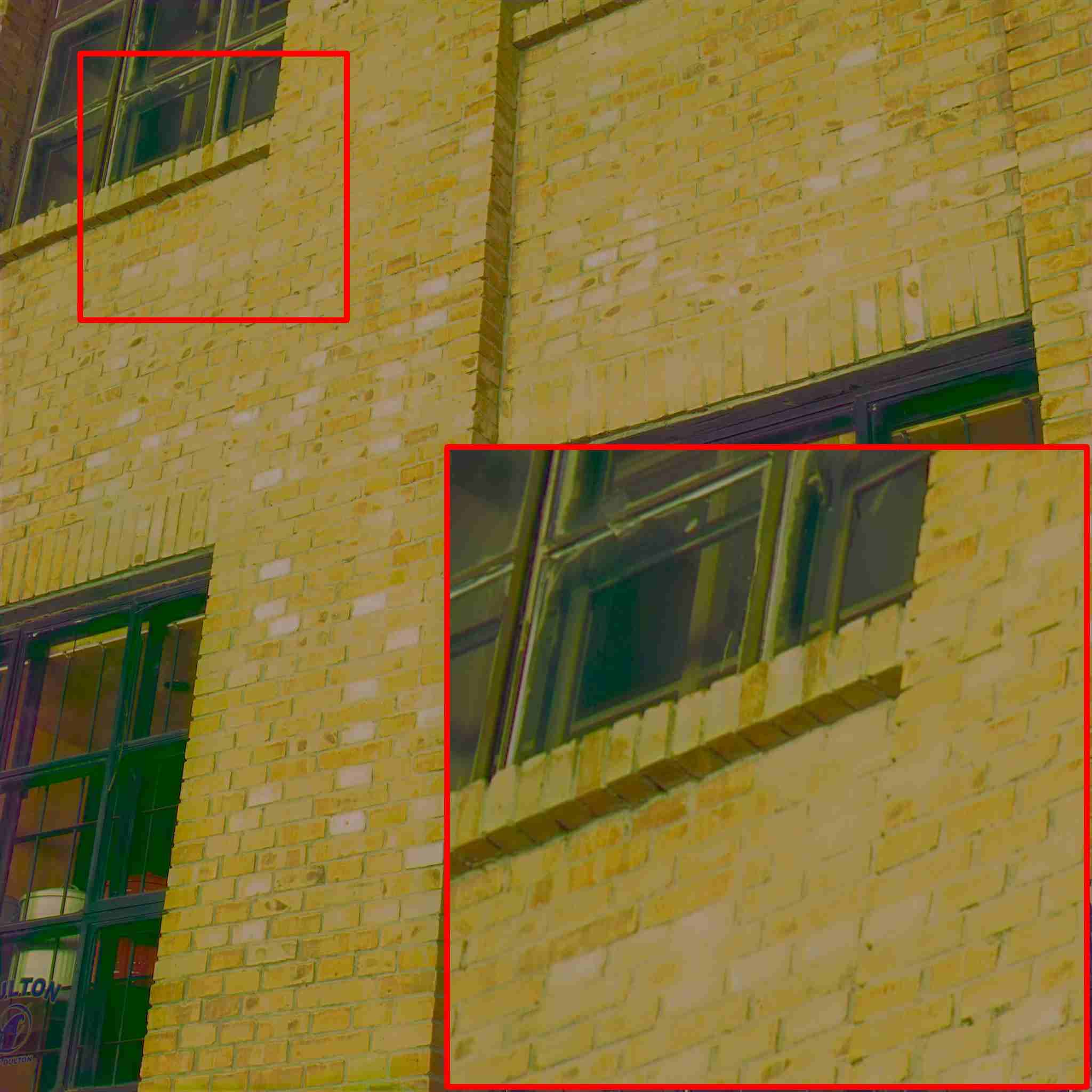}
  & \includegraphics[width=0.1\linewidth]{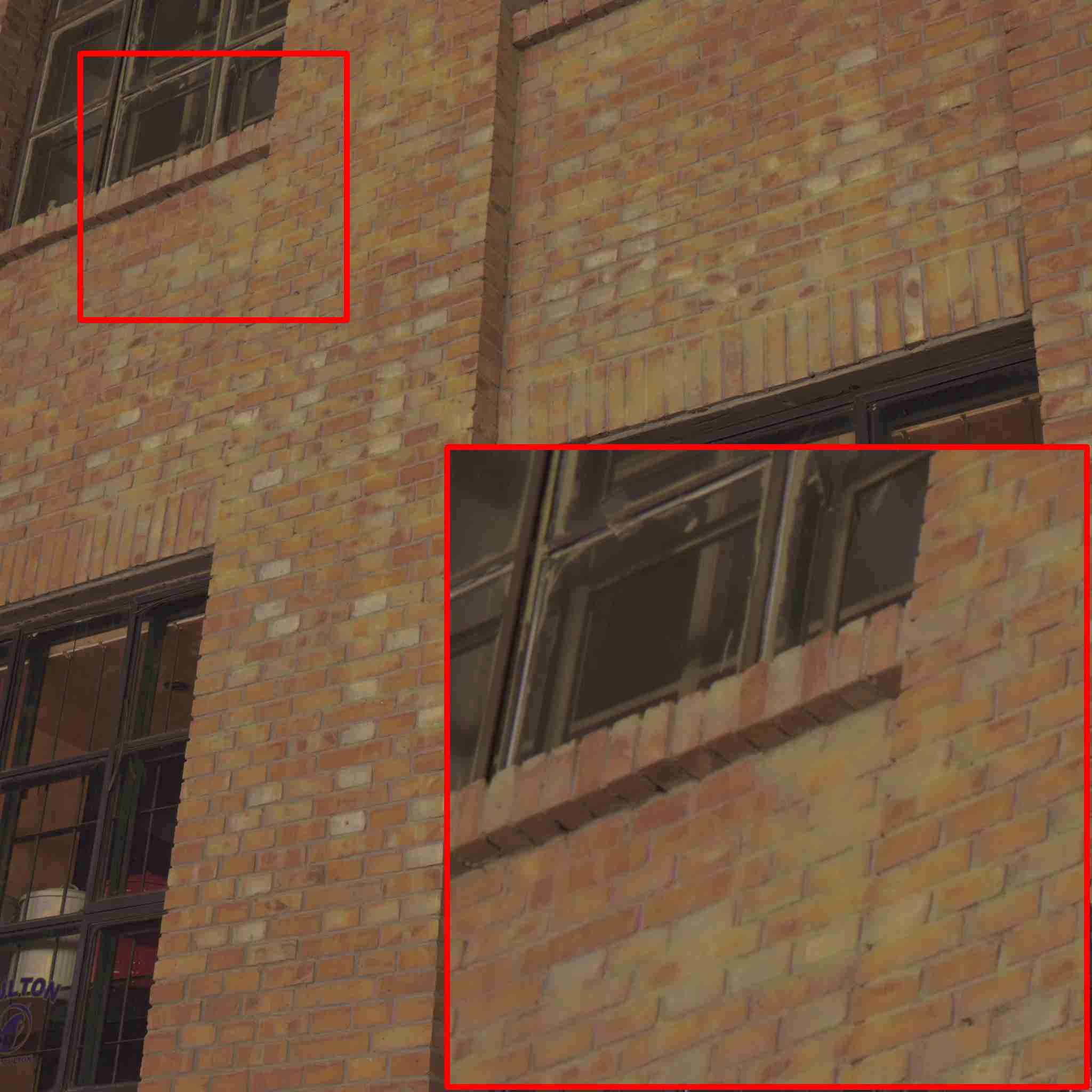}
  & \includegraphics[width=0.1\linewidth]{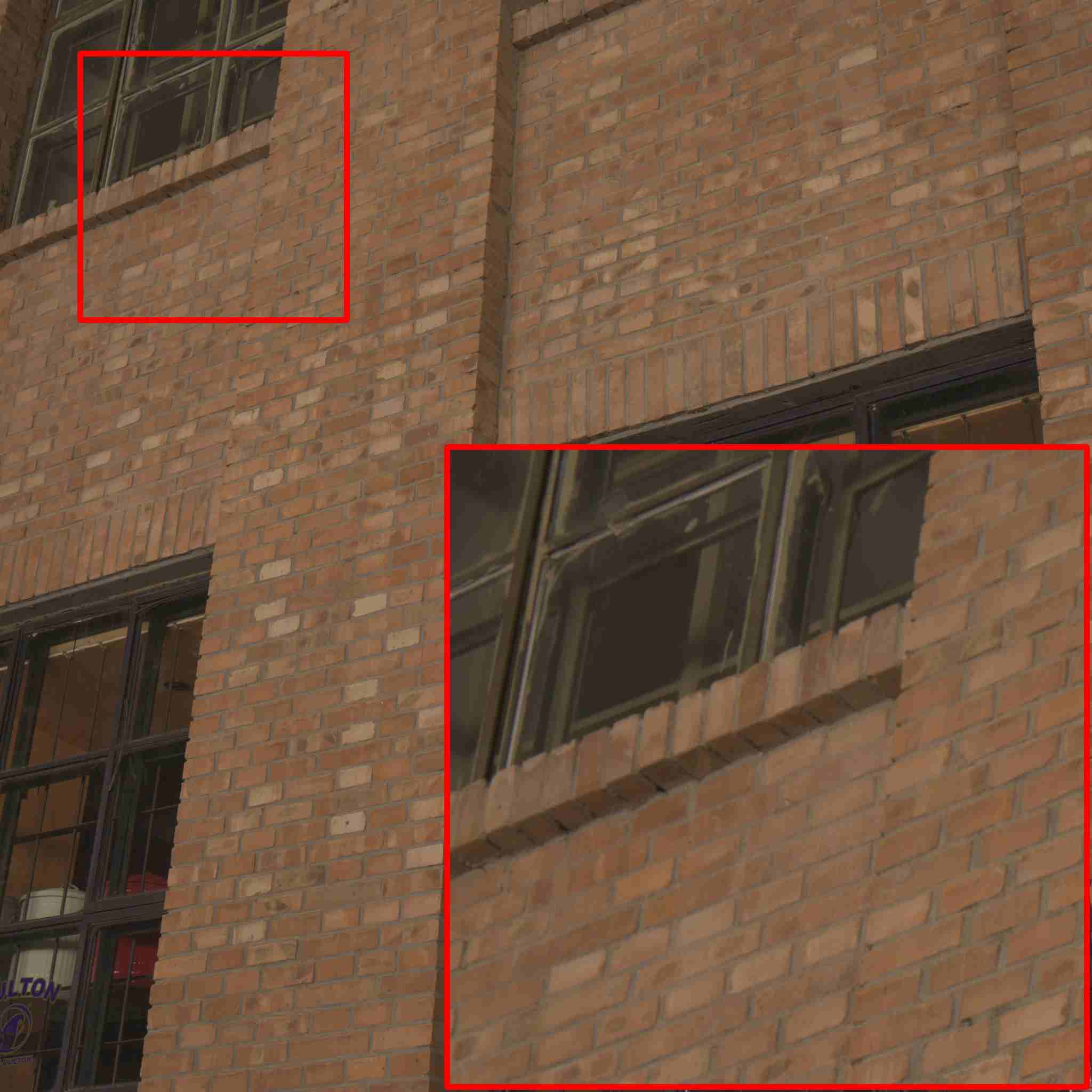} \\

 ES~\cite{jung2025exposure} & RED~\cite{lamba2021restoring} & ABF~\cite{dong2022abandoning} & CGNet~\cite{fu2023raw} & RawHDR~\cite{zou2023rawhdr} & SMG~\cite{xu2023low} & ADU~\cite{huang2024real} & RIECR (Ours) & Reference \\ \midrule

	\includegraphics[width=0.1\linewidth]{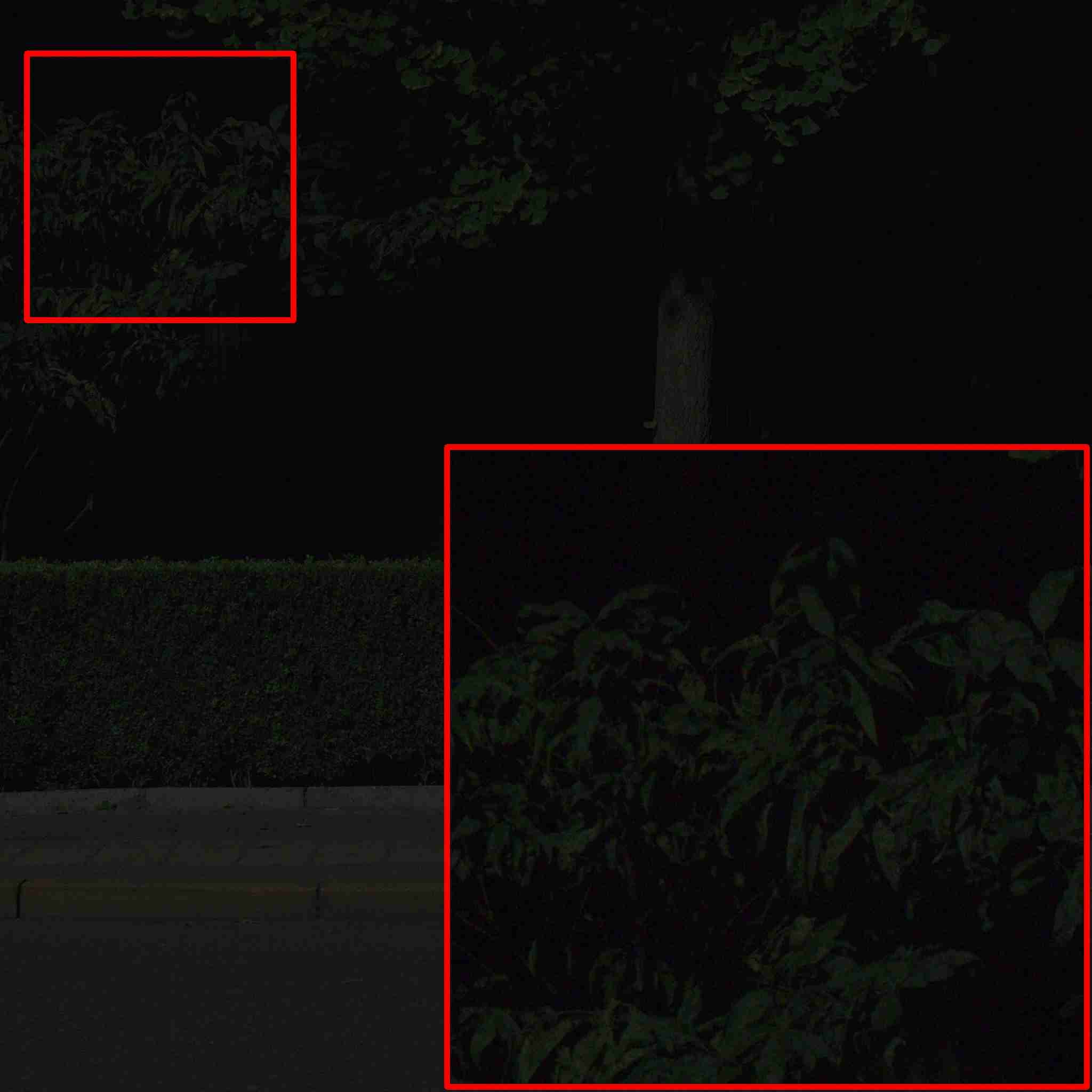}
	& \includegraphics[width=0.1\linewidth]{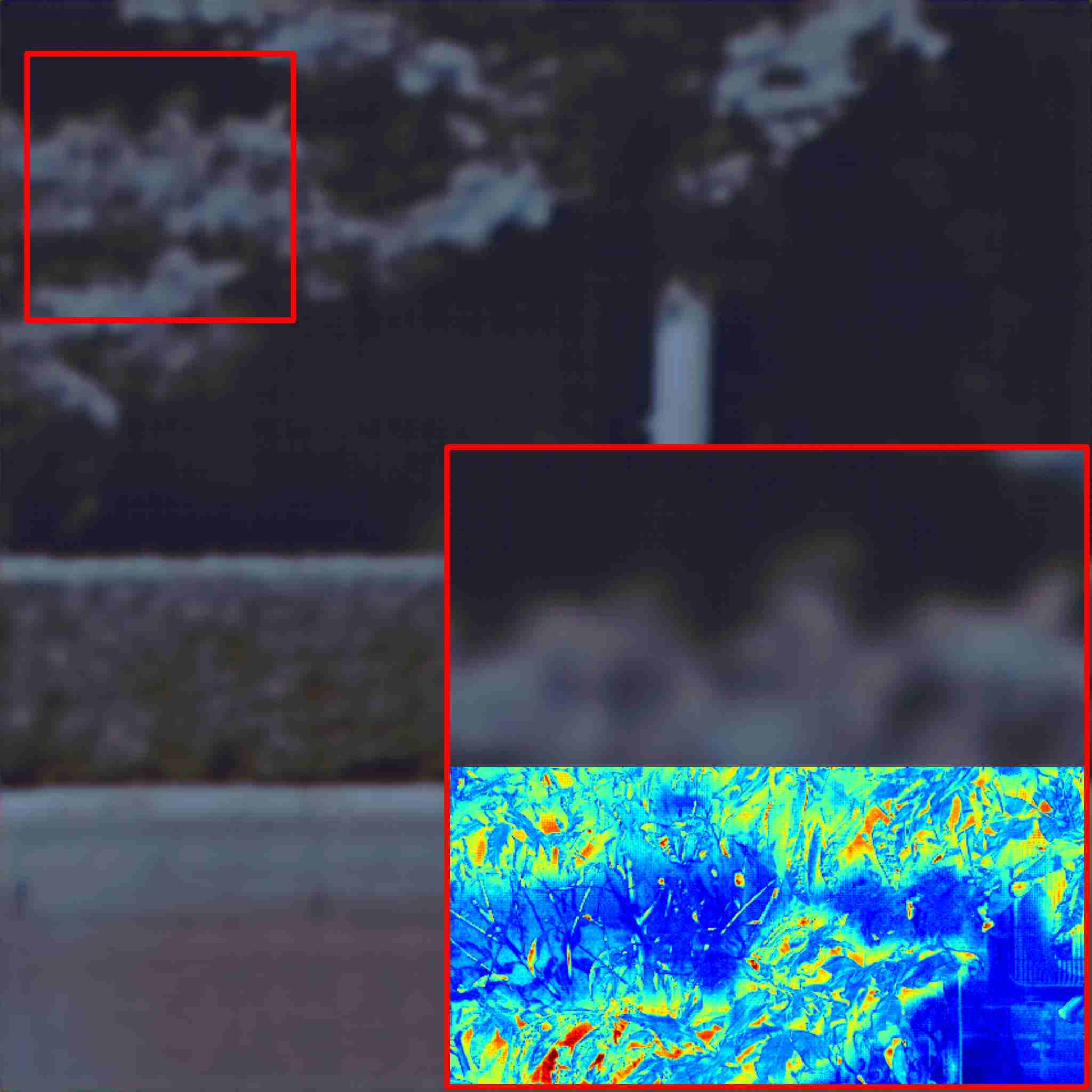}
  & \includegraphics[width=0.1\linewidth]{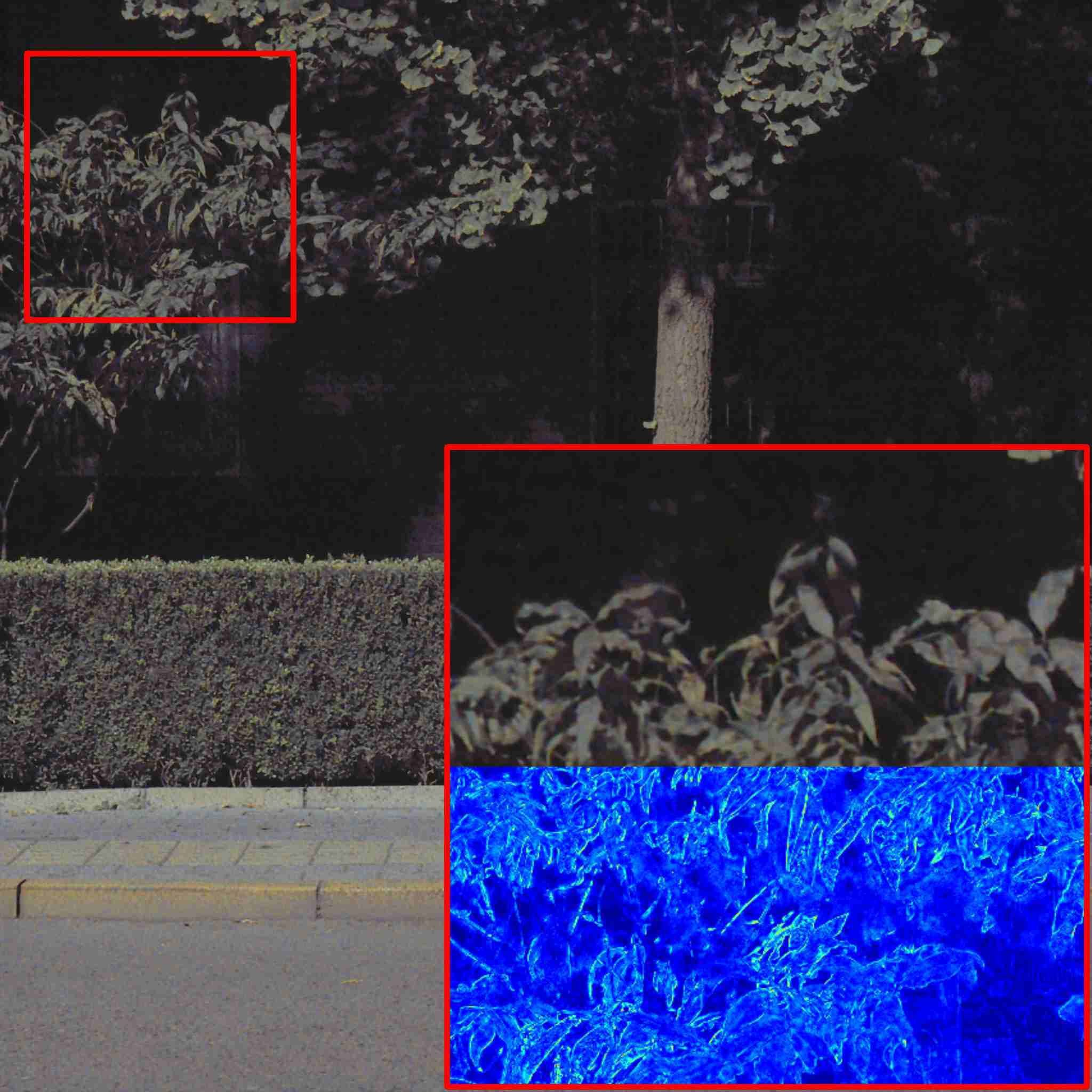}
  & \includegraphics[width=0.1\linewidth]{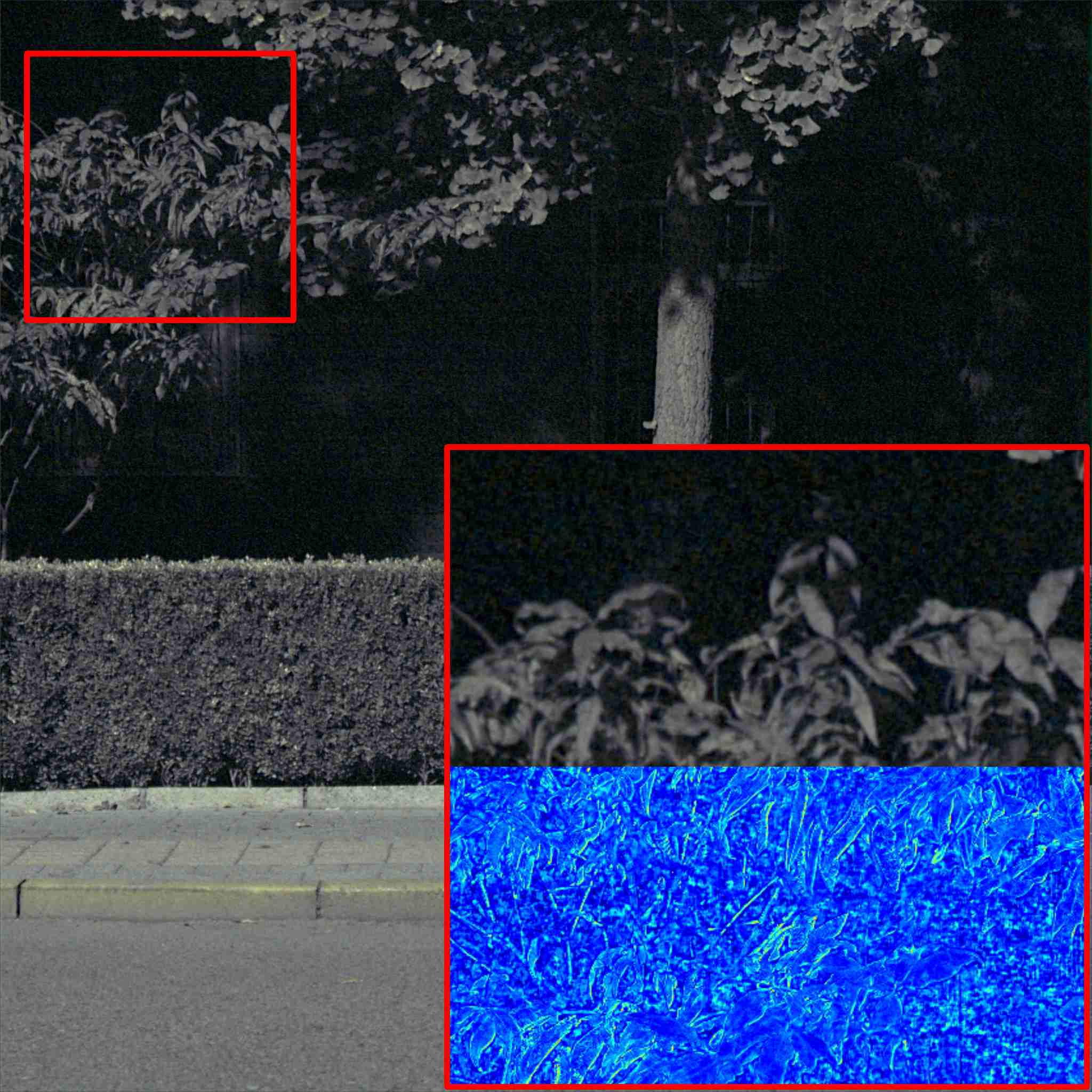}
  & \includegraphics[width=0.1\linewidth]{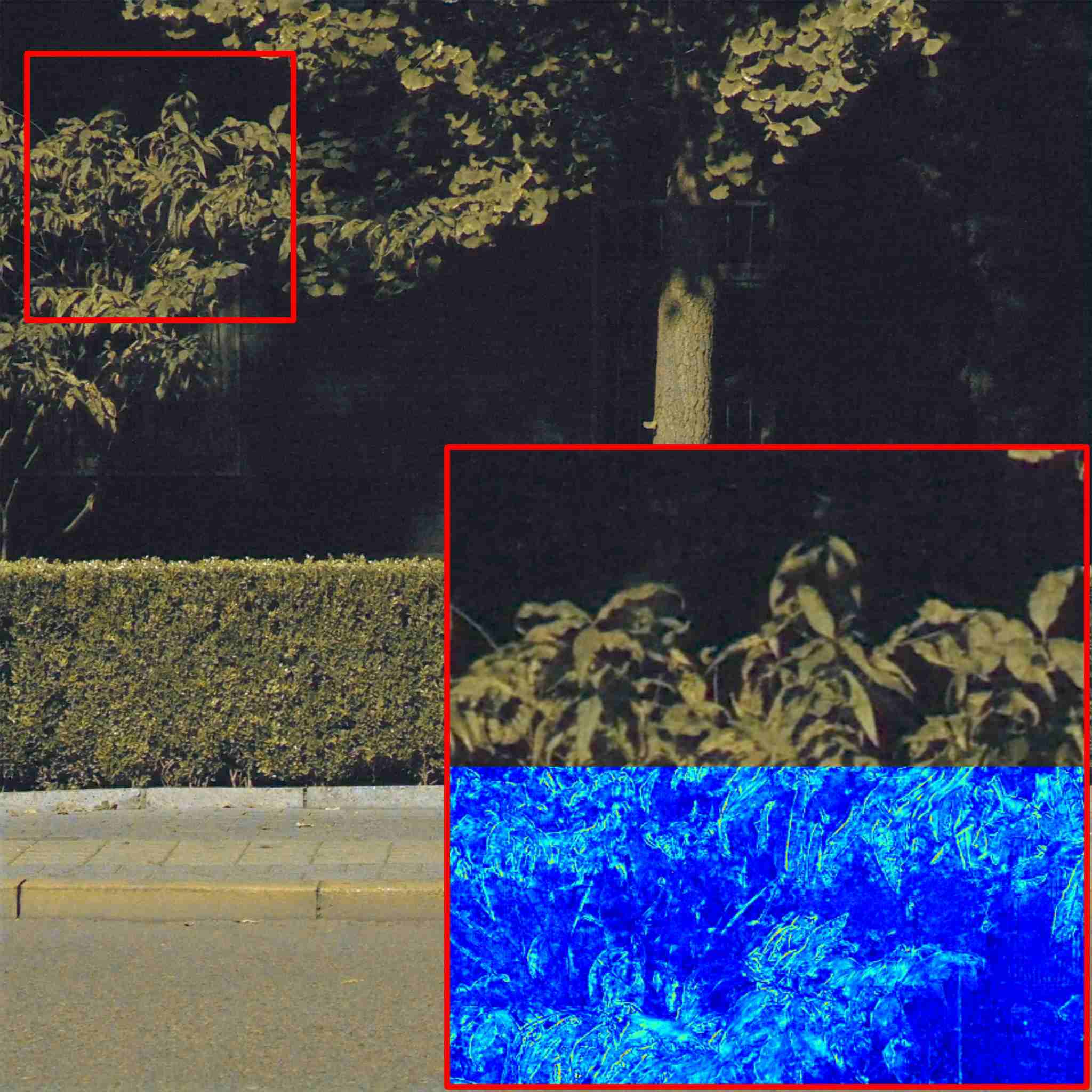}
  & \includegraphics[width=0.1\linewidth]{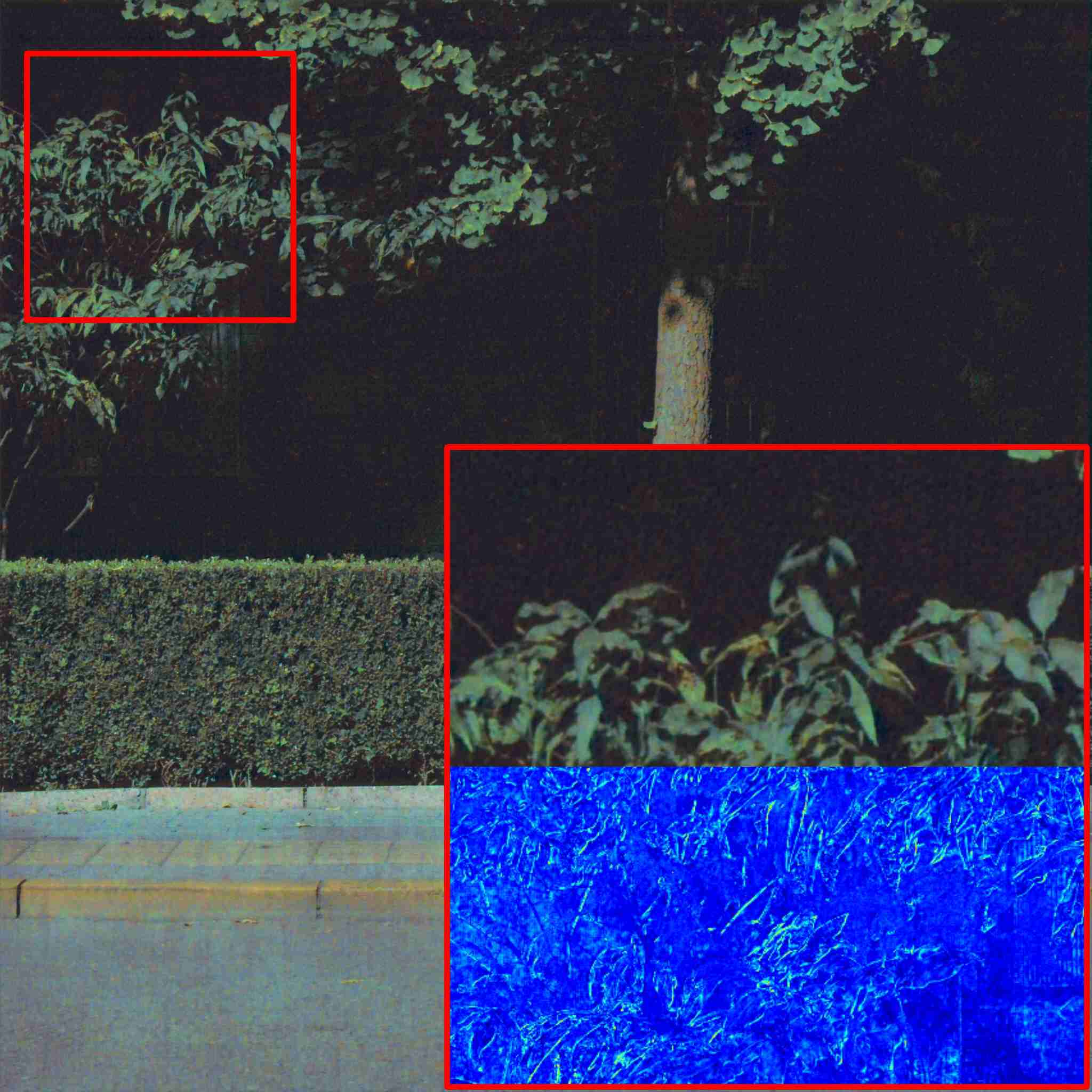}
  & \includegraphics[width=0.1\linewidth]{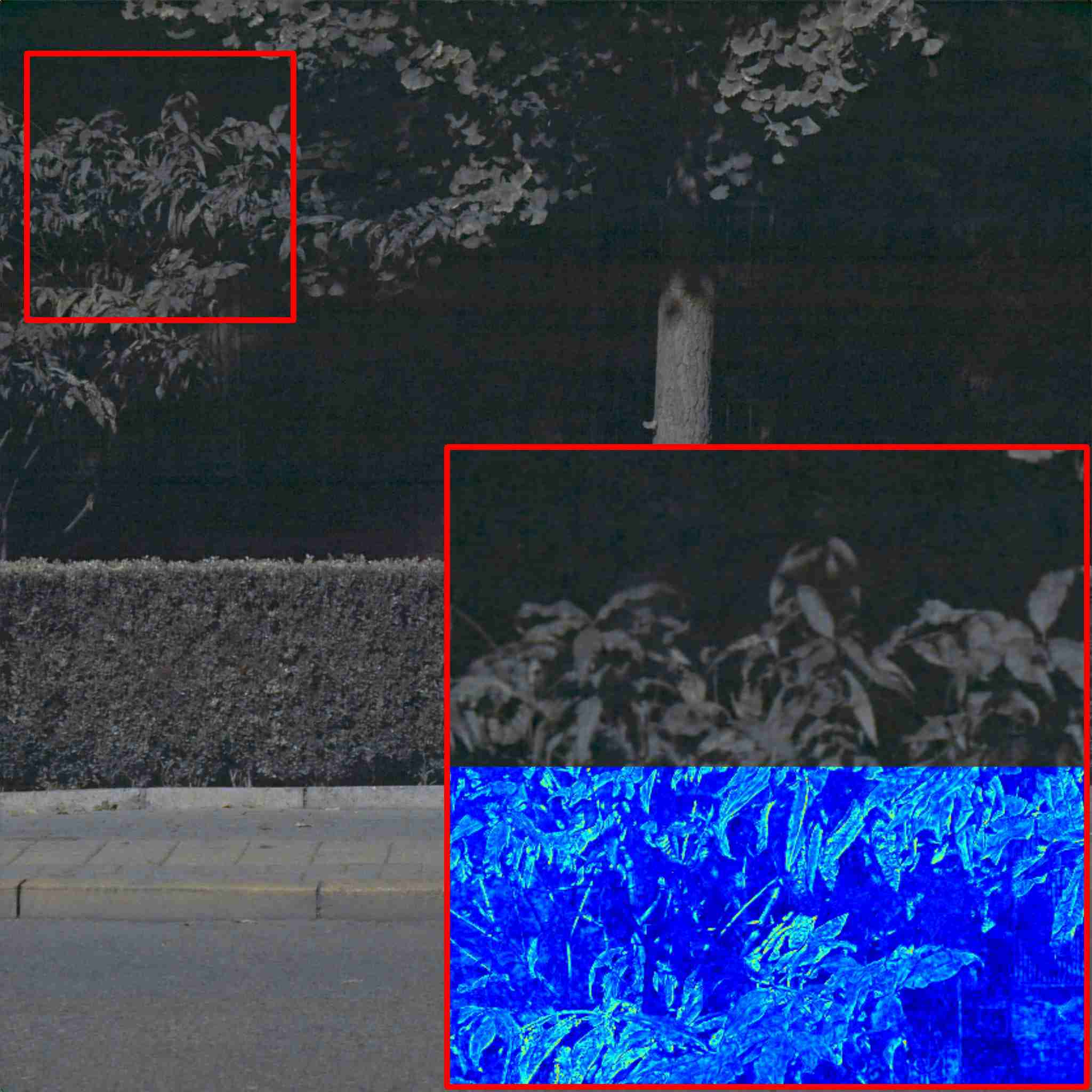}
  & \includegraphics[width=0.1\linewidth]{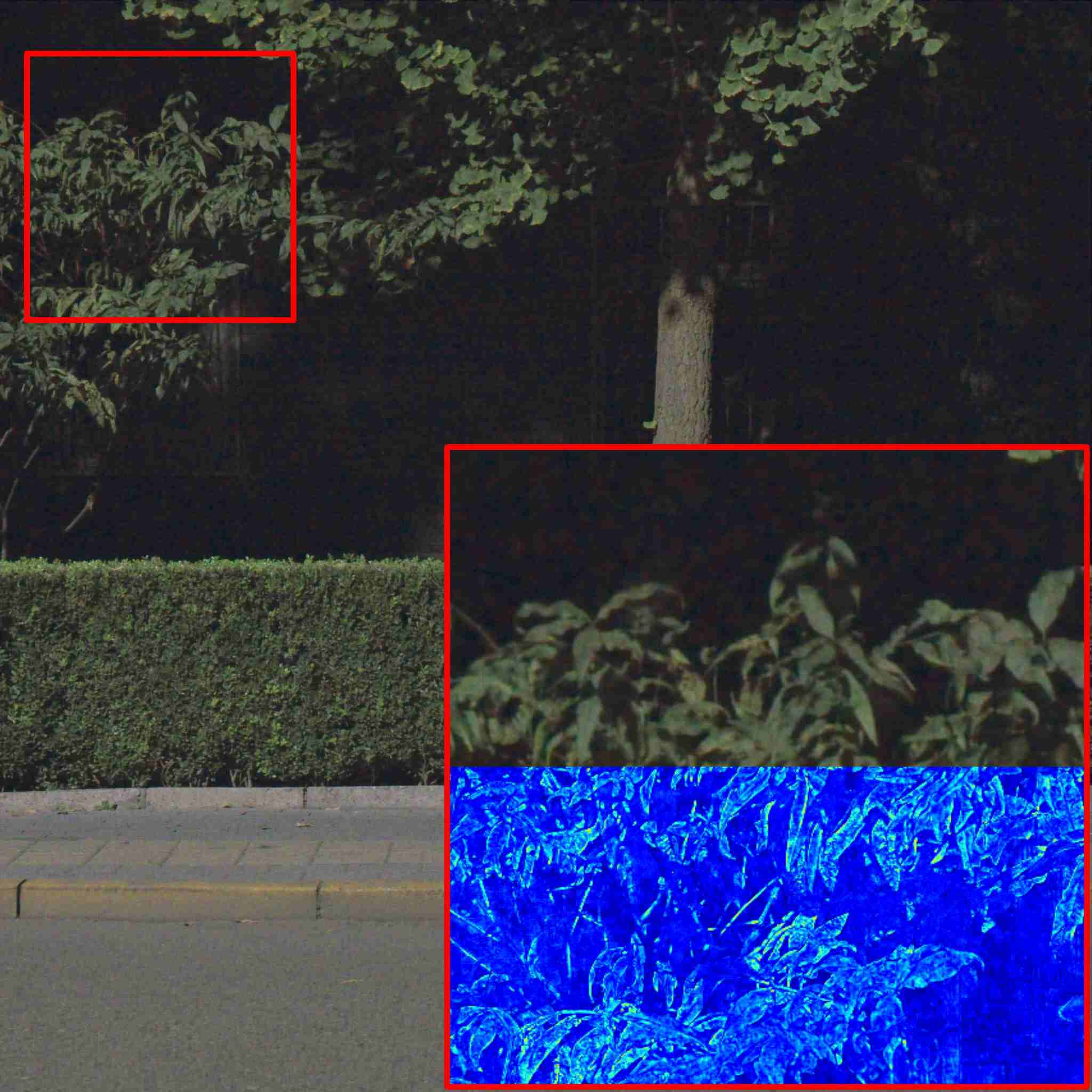}
  & \includegraphics[width=0.1\linewidth]{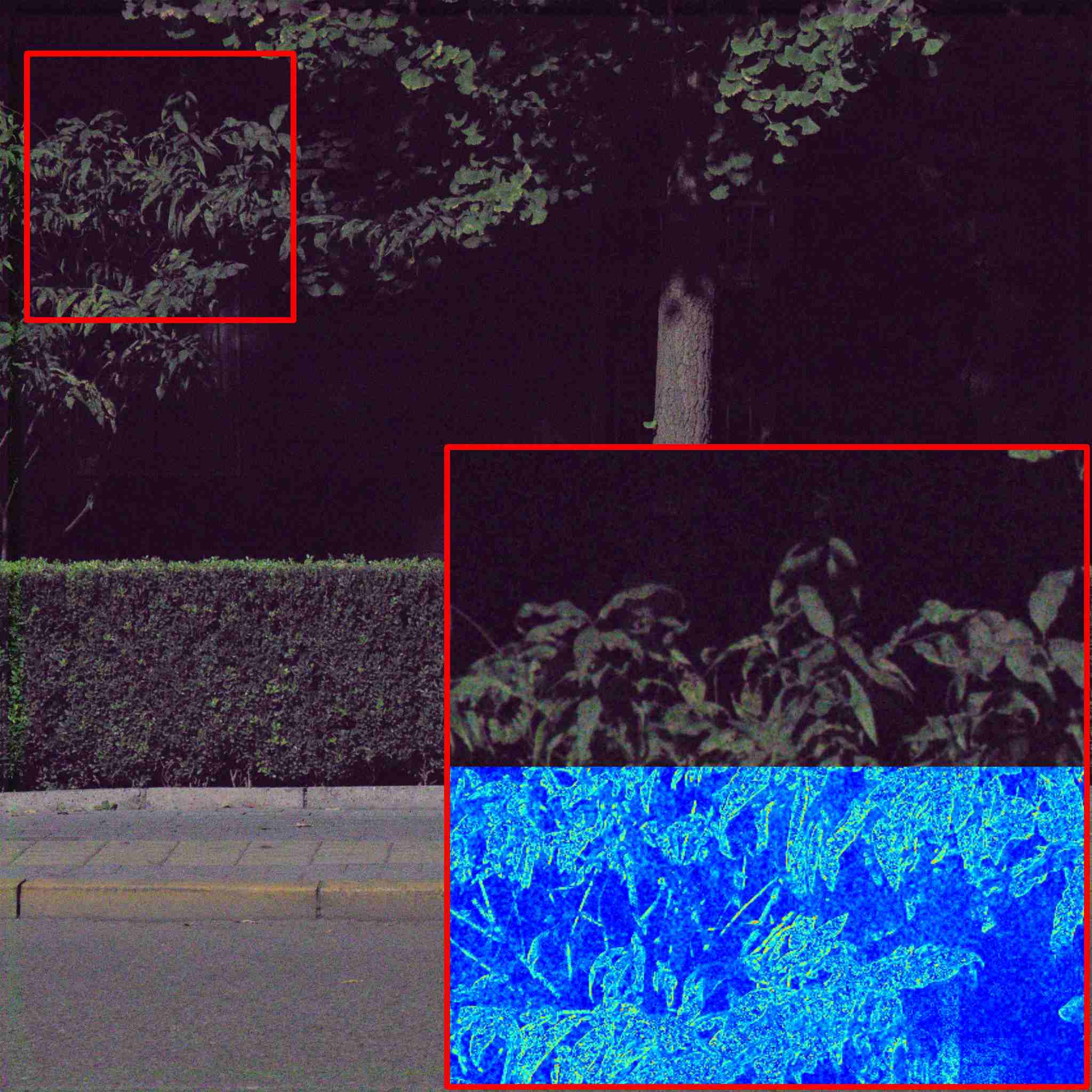}
   \\
   
Input  & MSEC~\cite{afifi2021learning} & ECCM~\cite{nsampi2021learning} & ECLNet~\cite{huang2022exposure-consistency}  & ENC-DRBN~\cite{huang2022exposure} & ENC-SID~\cite{huang2022exposure} & FECNet~\cite{huang2022deep} & RF~\cite{cai2023retinexformer} & CSEC~\cite{li2024color}  \\

   \includegraphics[width=0.1\linewidth]{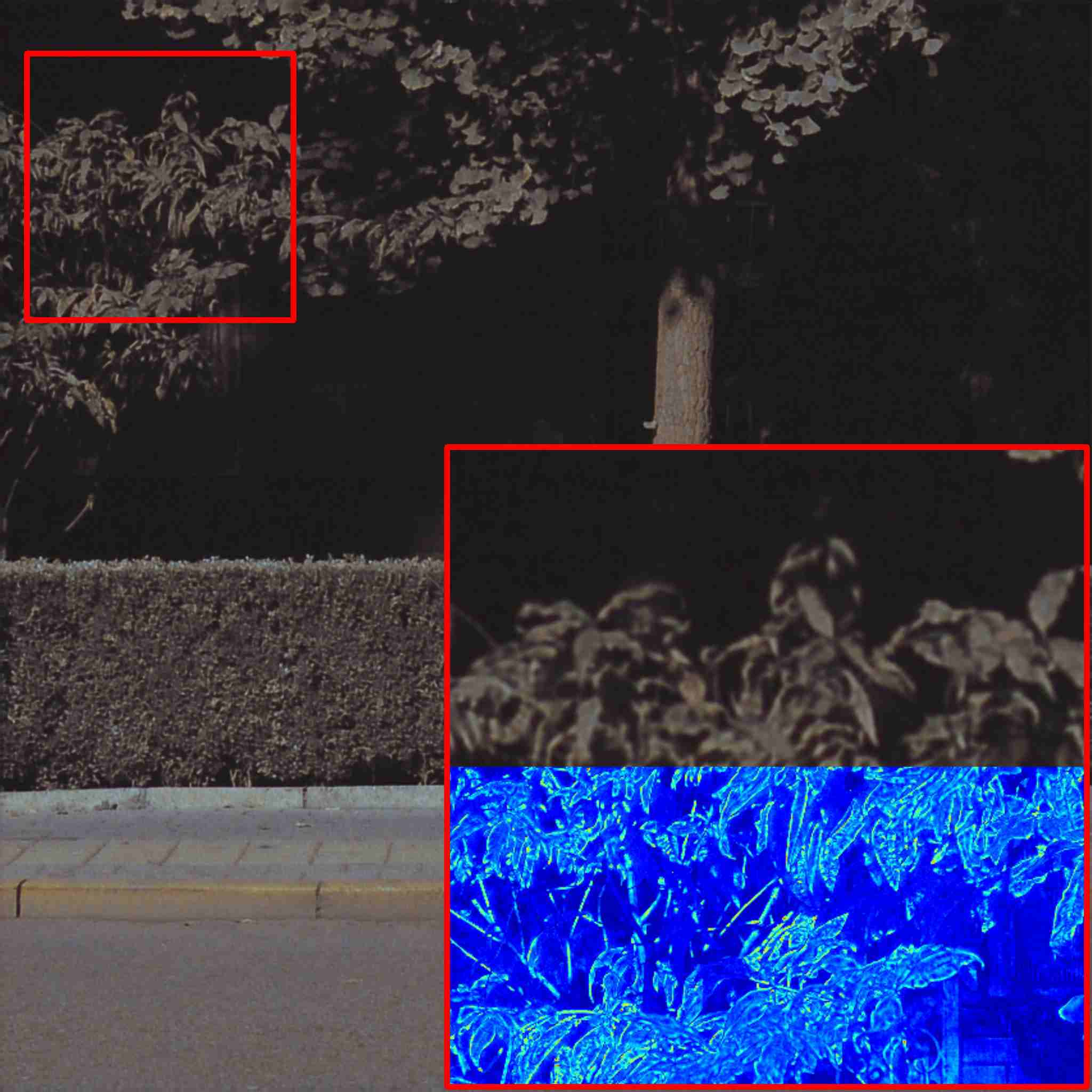}
  & \includegraphics[width=0.1\linewidth]{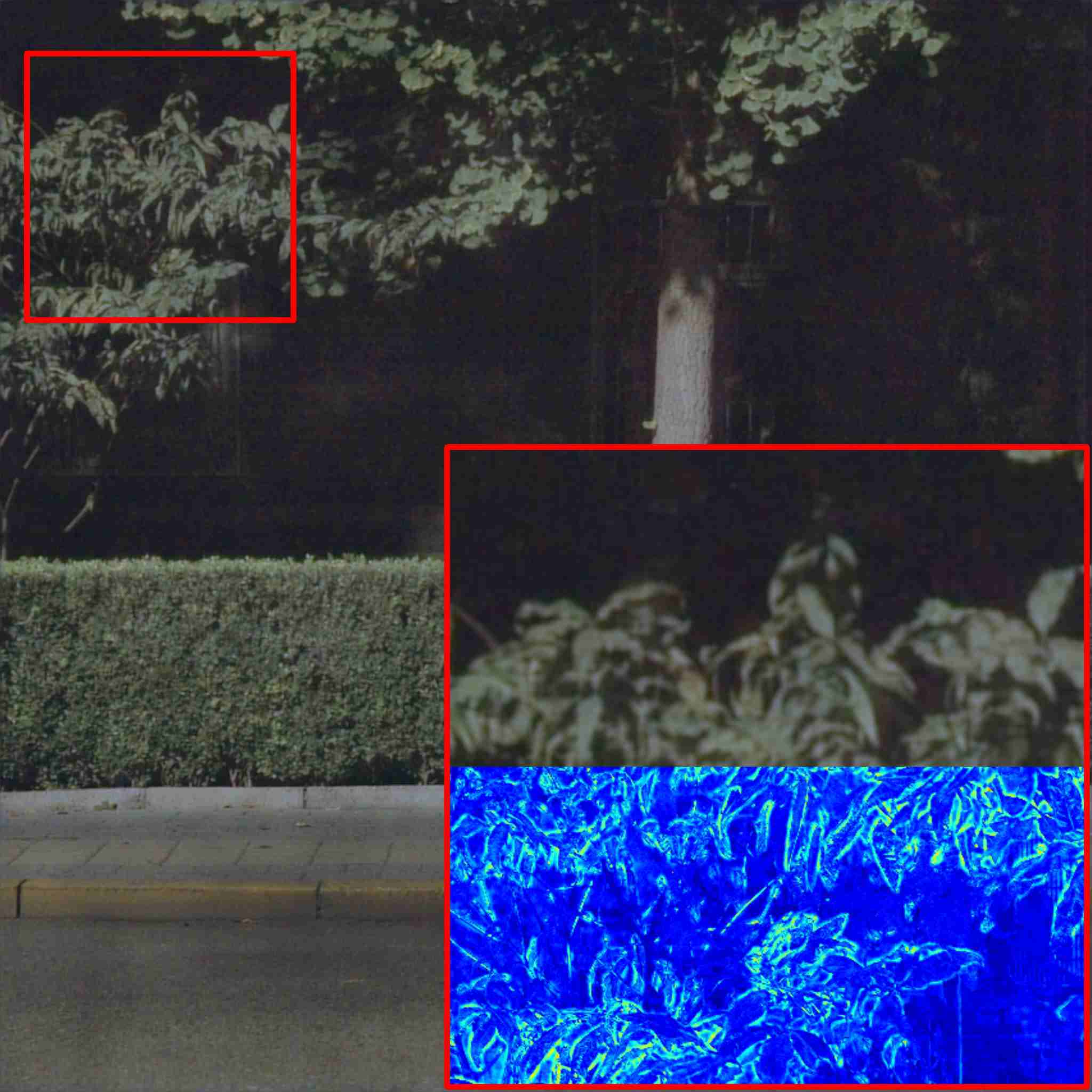}
  & \includegraphics[width=0.1\linewidth]{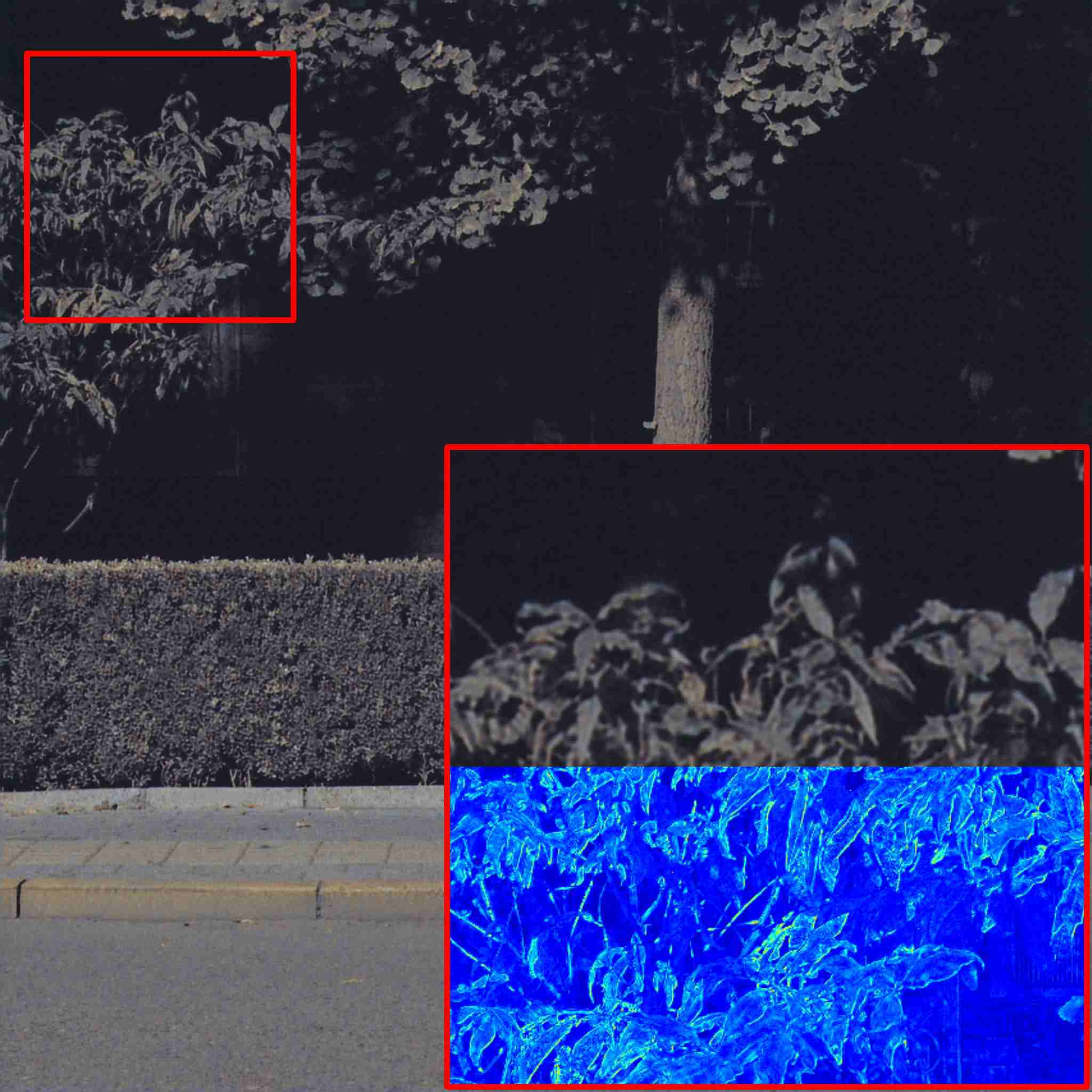}
  & \includegraphics[width=0.1\linewidth]{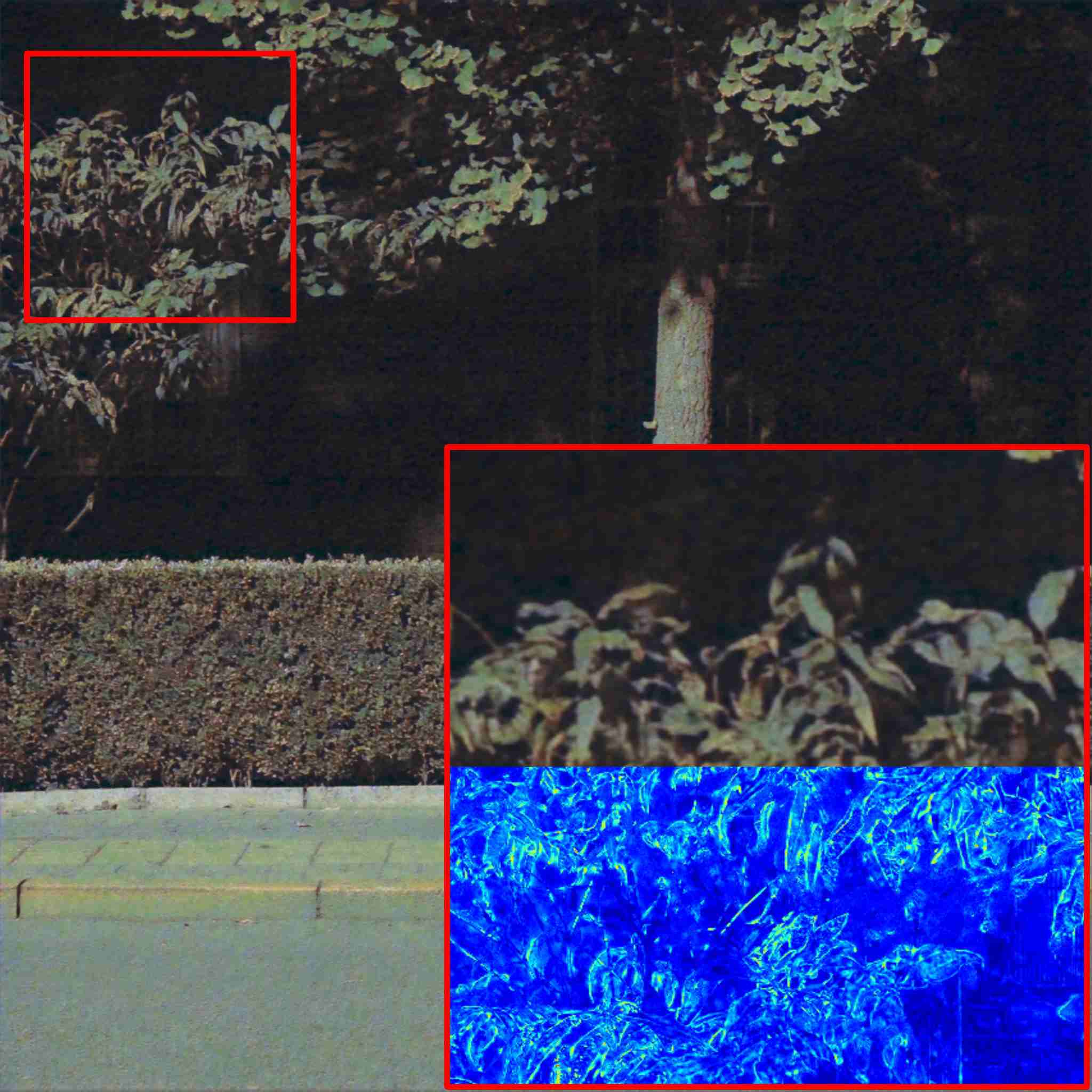}
  & \includegraphics[width=0.1\linewidth]{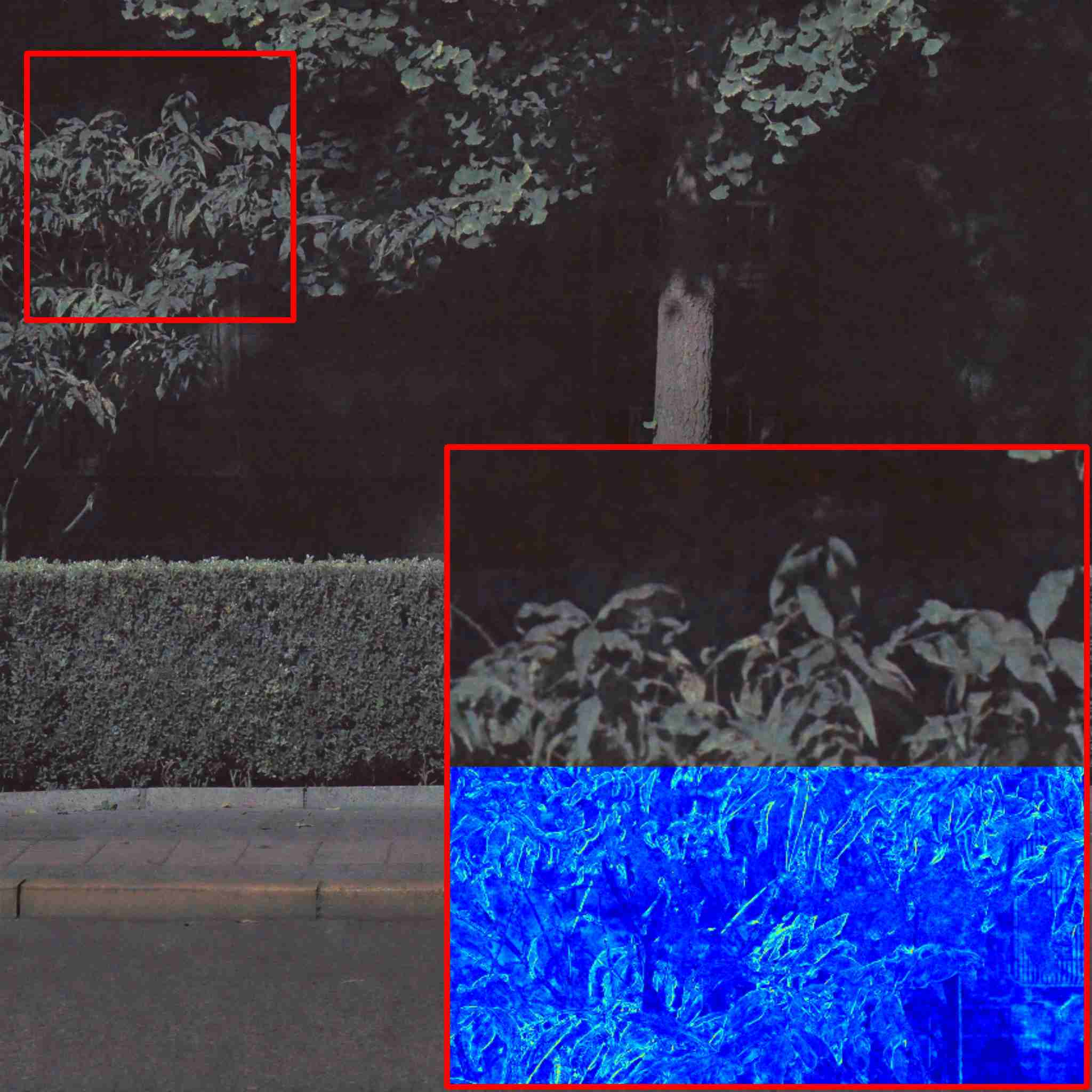}
  & \includegraphics[width=0.1\linewidth]{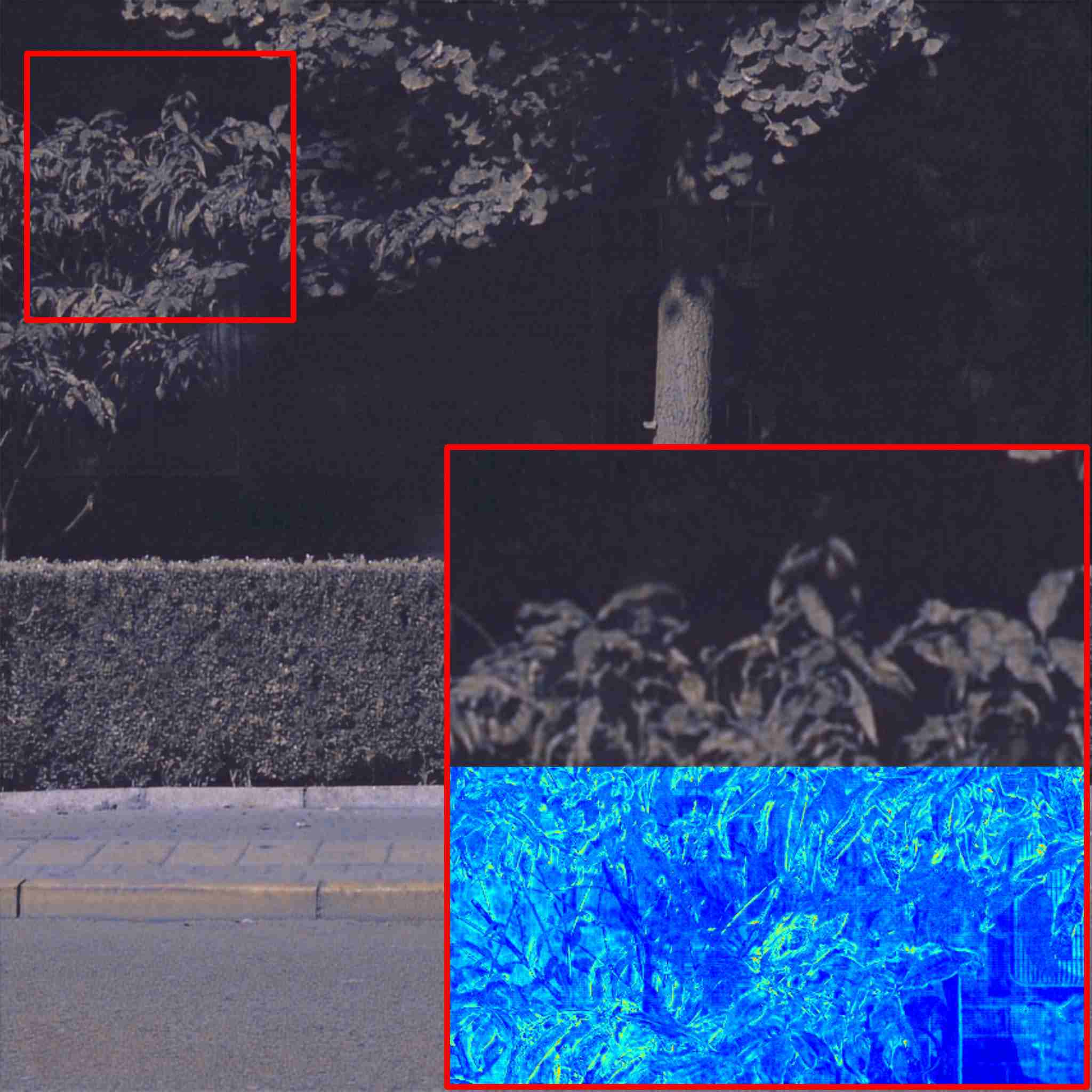}
  & \includegraphics[width=0.1\linewidth]{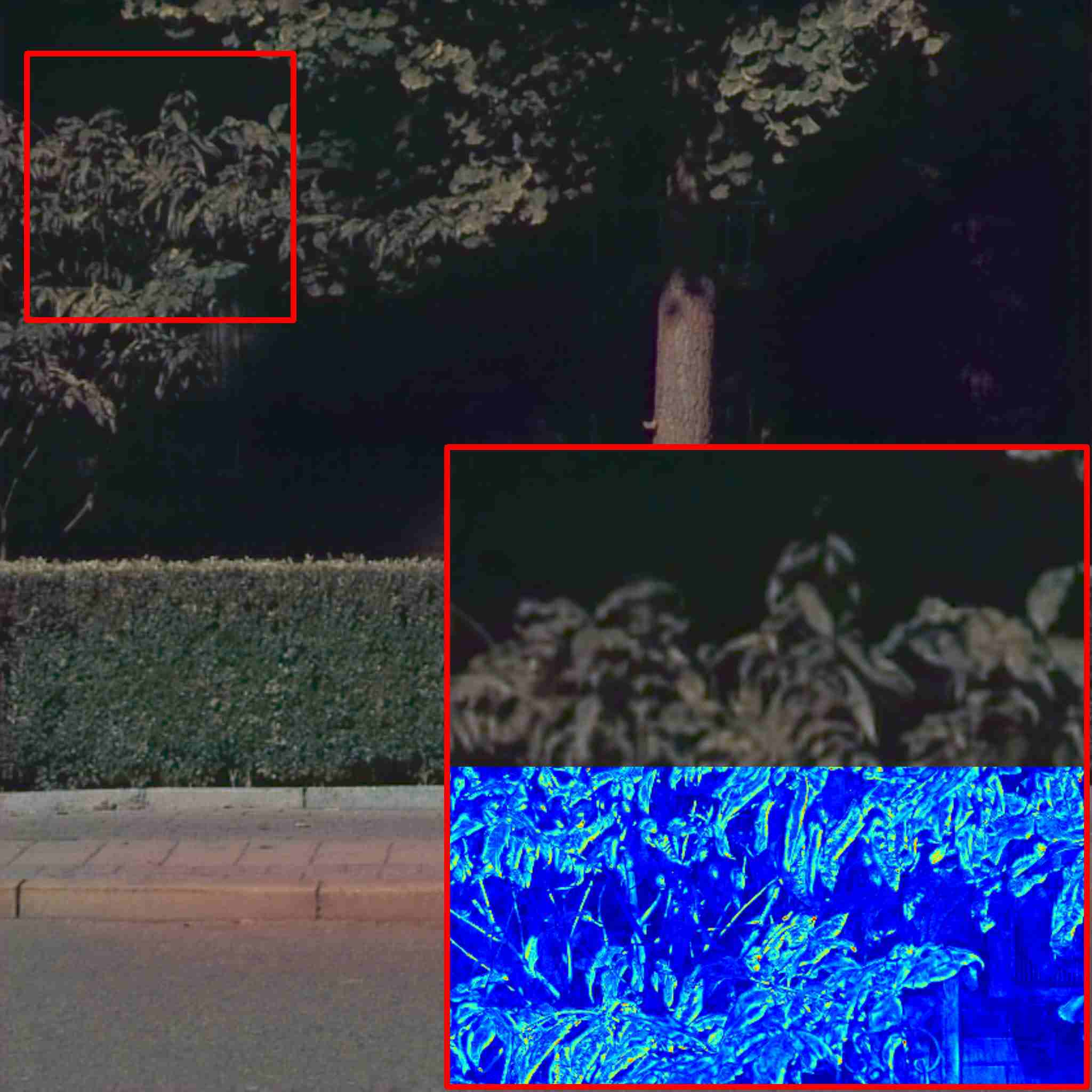}
  & \includegraphics[width=0.1\linewidth]{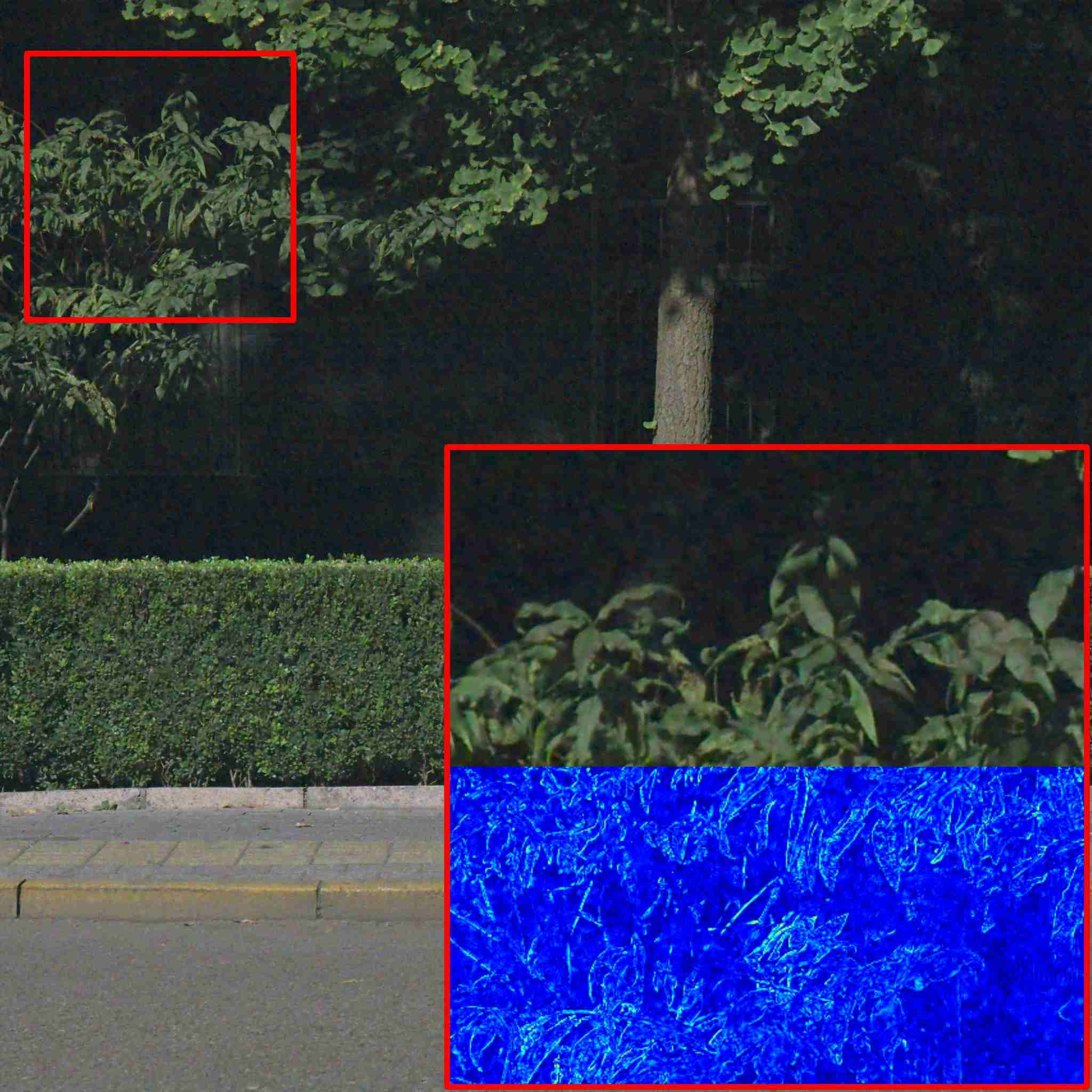} 
  & \includegraphics[width=0.1\linewidth]{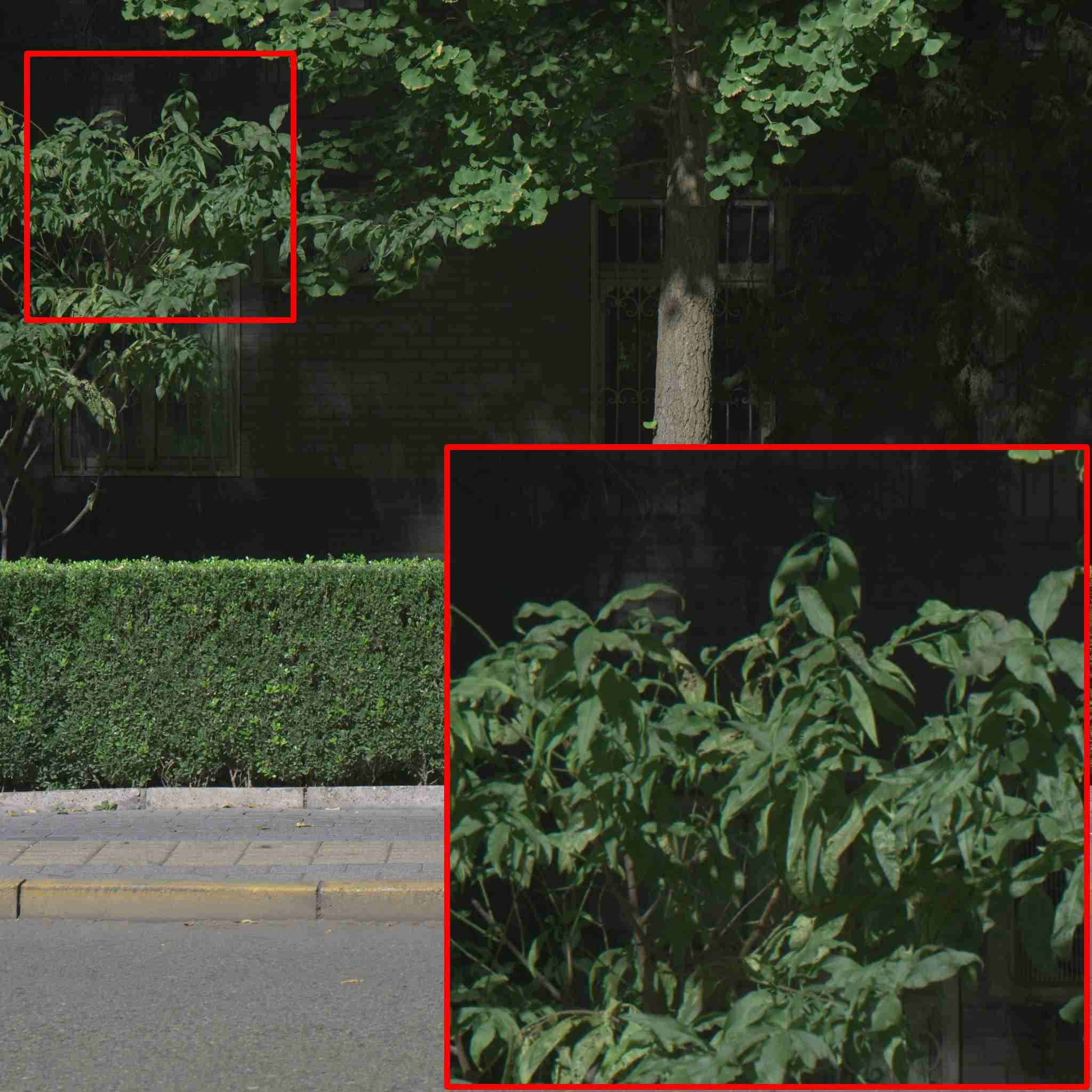} \\

ES~\cite{jung2025exposure} & RED~\cite{lamba2021restoring} & ABF~\cite{dong2022abandoning} & CGNet~\cite{fu2023raw} & RawHDR~\cite{zou2023rawhdr} & SMG~\cite{xu2023low} & ADU~\cite{huang2024real} & RIECR (Ours) & Reference \\

  \end{tabular}
 \caption{CMER visual comparison for over-exposure (top) and under-exposure (bottom), displayed using the same RAW-to-sRGB ISP.}
 \label{fig:visual quality comparison}
\end{figure*}

\textit{Synthetic RAW Image Dataset:}
The synthetic dataset, named SMIT, applies the same eight ratios to 1,500 MIT-Adobe FiveK RAW references~\cite{bychkovsky2011learning}. Linear amplification plus clipping simulates over-exposure; attenuation plus sensor noise simulates under-exposure~\cite{afifi2021learning,fu2023raw,wei2020physics}, yielding 12,000 pre-training pairs.

\section{Experiments}
\label{sec:experiments}

We evaluate RIECR against representative sRGB- and RAW-domain methods, followed by component, domain, efficiency, and cross-camera analyses.

\subsection{Experimental Settings}

\noindent{\textbf{Compared Methods.}}
We compare eight sRGB methods: MSEC~\cite{afifi2021learning}, ECCM~\cite{nsampi2021learning}, ECLNet~\cite{huang2022exposure-consistency}, ENC~\cite{huang2022exposure}, FECNet~\cite{huang2022deep}, Retinexformer (RF)~\cite{cai2023retinexformer}, CSEC~\cite{li2024color}, and ES~\cite{jung2025exposure}; and eight RAW methods: SID~\cite{chen2018learning}, RED~\cite{lamba2021restoring}, ABF~\cite{dong2022abandoning}, CGNet~\cite{fu2023raw}, RawHDR~\cite{zou2023rawhdr}, SMG~\cite{xu2023low}, ADU~\cite{huang2024real}, and RRM~\cite{chen2025retinex}. We adapt sRGB input/output layers to packed four-channel RAW and retrain all methods on identical splits and protocols using official implementations when available.

\noindent{\textbf{Evaluation Metrics.}}
We report PSNR and SSIM~\cite{wang2004image} on normalized RAW outputs for high-light, low-light, and their average; higher values indicate better restoration. All visualizations use the same RAW-to-sRGB ISP.

\begin{figure*}[t]
\small
\centering
  \setlength{\tabcolsep}{0.9mm}
  \renewcommand{\arraystretch}{1}
\begin{tabular}{cccccccc}
	
	\includegraphics[width=0.155\linewidth]{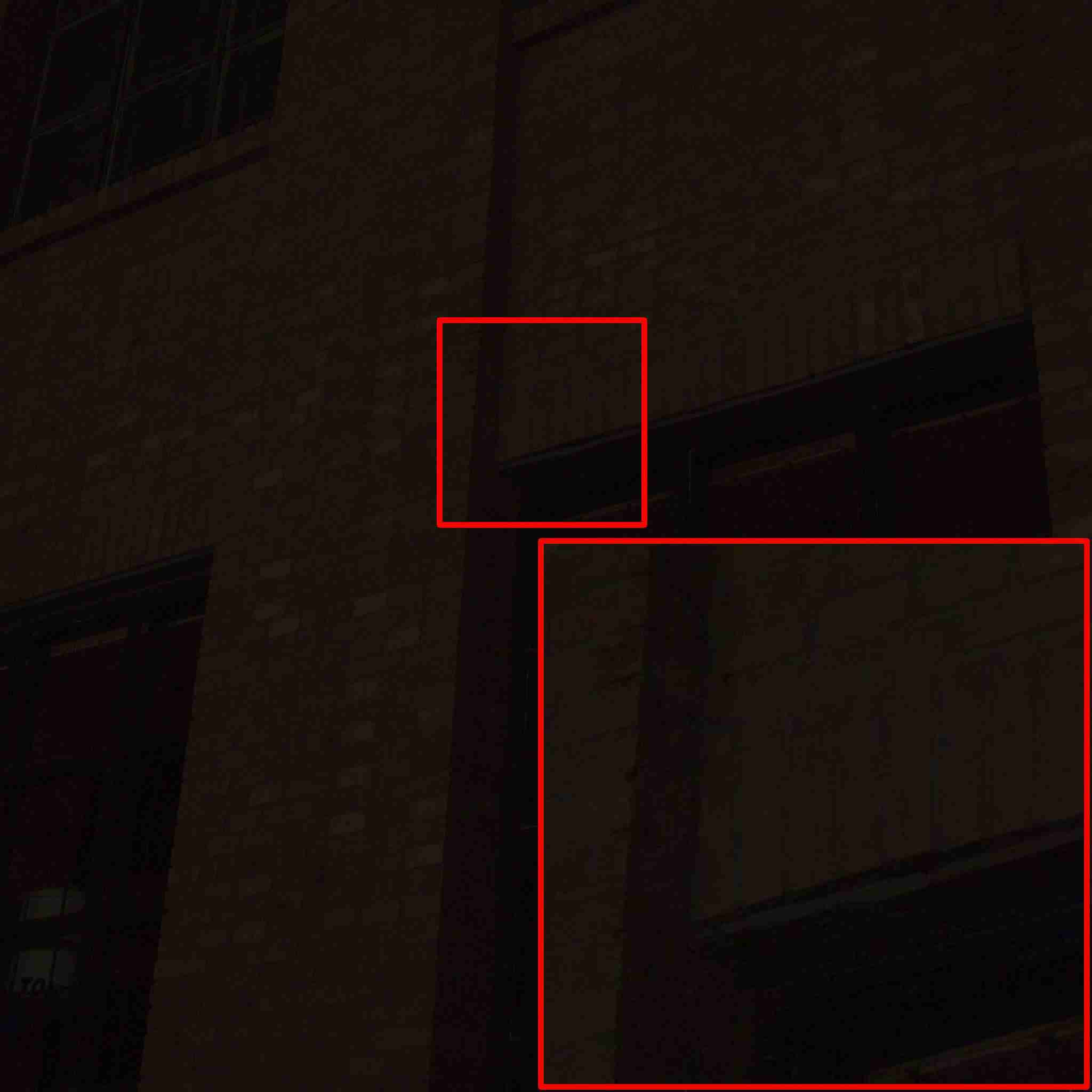} &
    \includegraphics[width=0.155\linewidth]{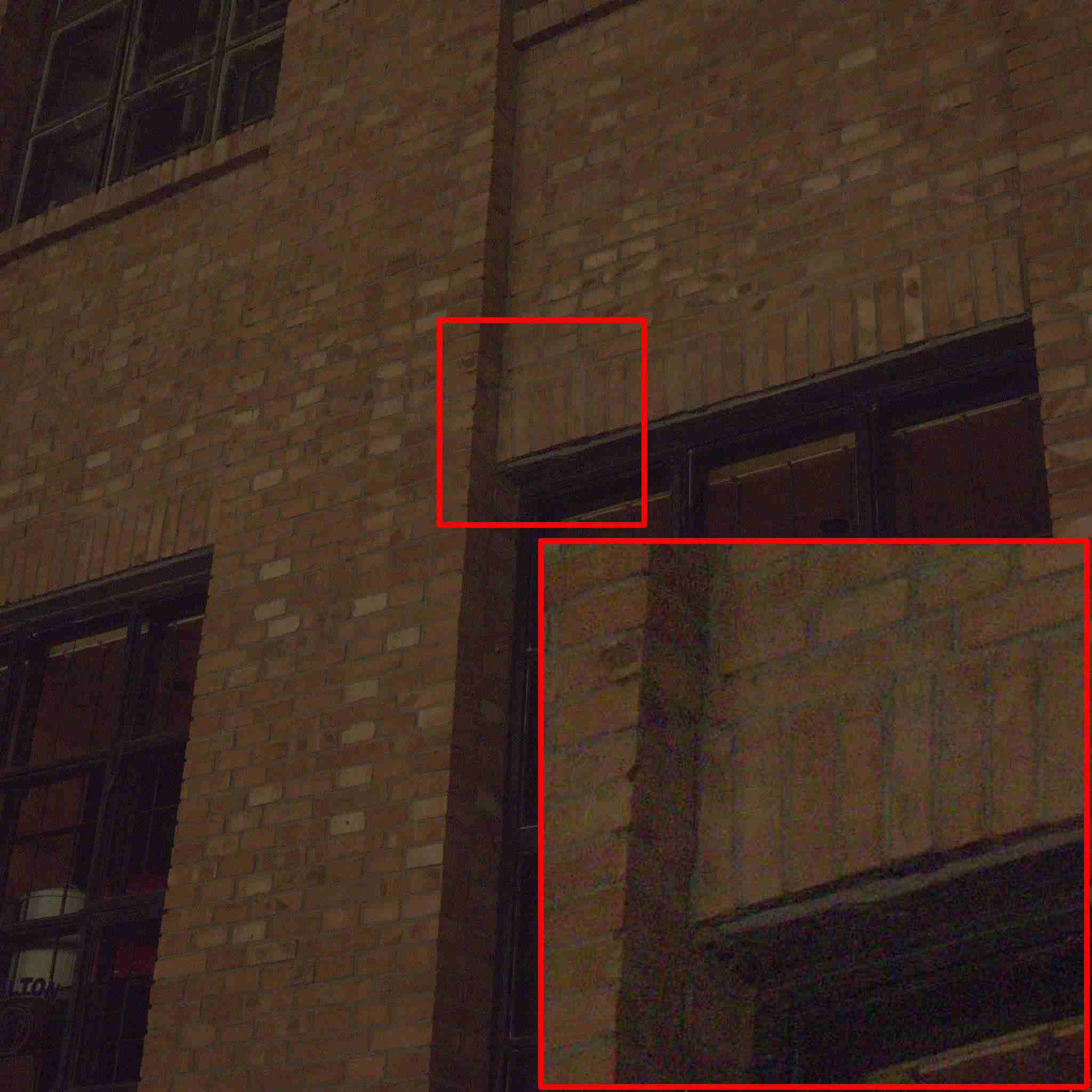} &
    \includegraphics[width=0.155\linewidth]{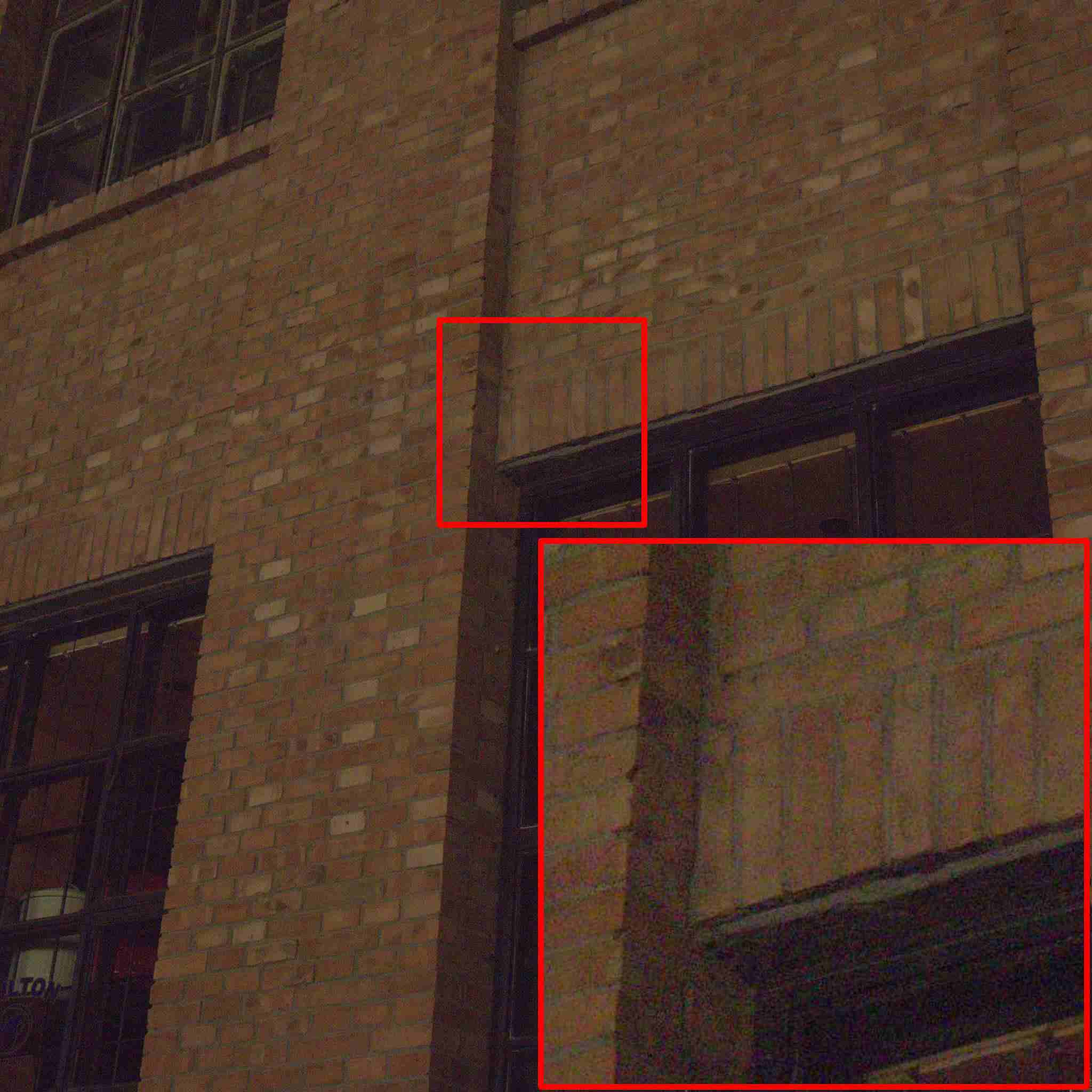}  &
    \includegraphics[width=0.155\linewidth]{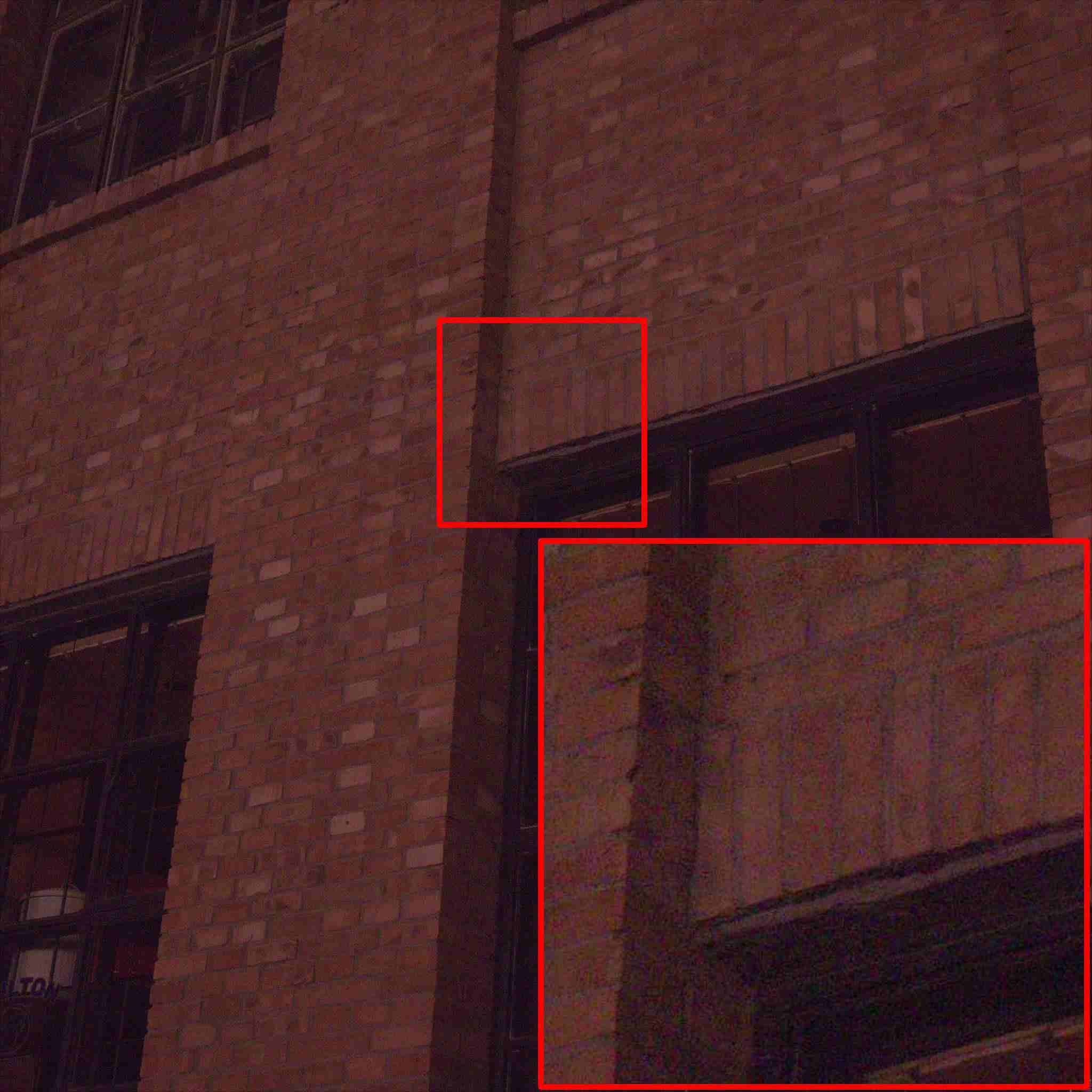} &
    \includegraphics[width=0.155\linewidth]{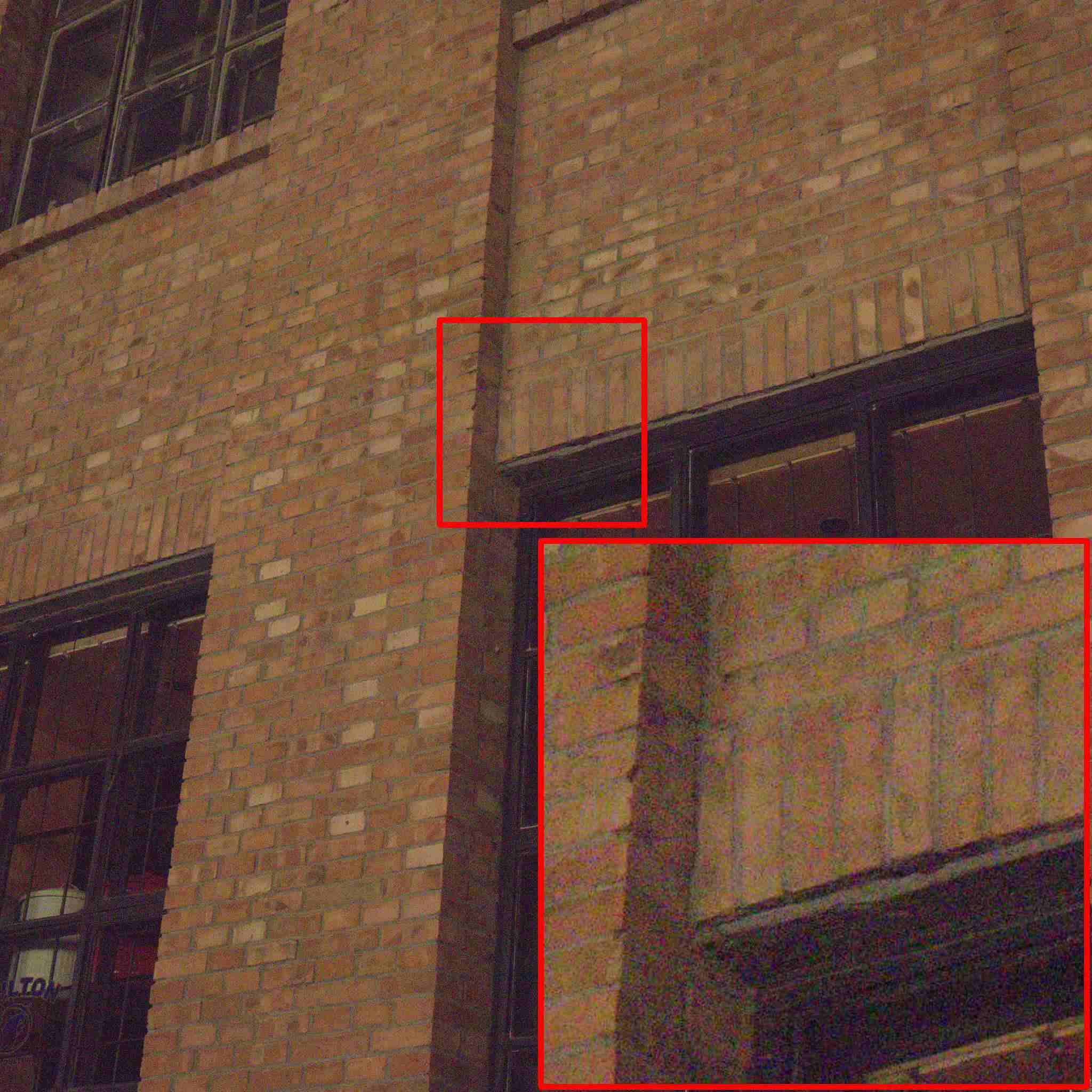} &
    \includegraphics[width=0.155\linewidth]{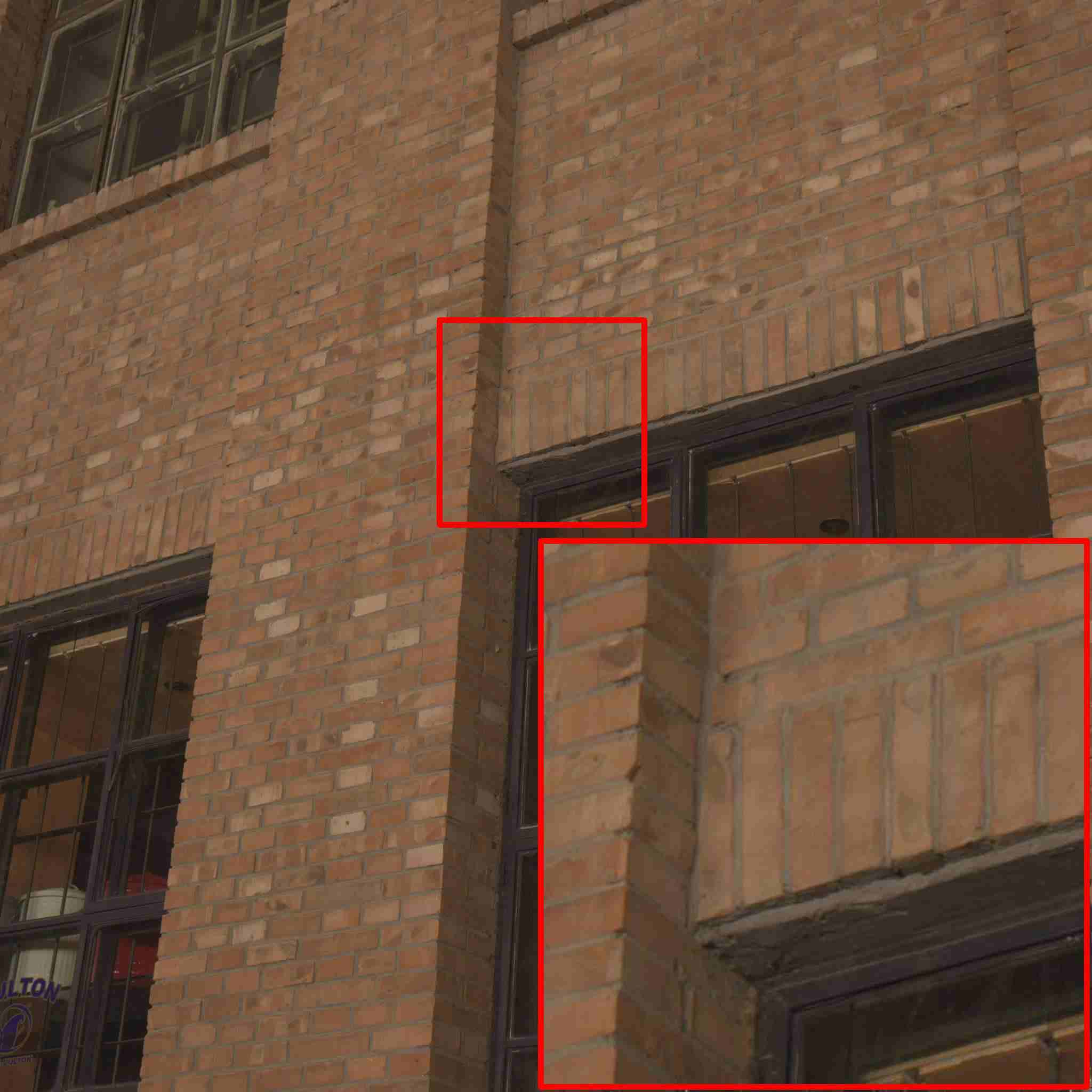} \\
    Input & w/o Guidance &  w/o Label & w/o Linear &  Ours &  Reference \\
  
  \end{tabular}
 \caption{Comparison of illumination correction by the LICN in our model. The term \textit{w/o Guidance} indicates adjustment without separating the color channels, \textit{w/o Label} means adjustment without estimating the illumination label, \textit{w/o Linear} means adjustment without linear operation, instead of an illumination map that is added to the original image.}
 \label{fig:illumination comparison}
\end{figure*}

\begin{table}[t]
    \renewcommand{\arraystretch}{1.0}
    \setlength{\tabcolsep}{5.3 pt}
       \centering
       \caption{Ablation of the proposed RIECR framework.}
    \label{tab: ablation design}
       \small
       \resizebox{\linewidth}{!}{%
        \begin{tabular}{cc|cc|cc|cc}
        \hline\hline
		\multirow{2}{*}{\raisebox{-0.5ex}{\textbf{LICN}}} & \multirow{2}{*}{\raisebox{-0.5ex}{\textbf{FEN}}} & \multicolumn{2}{c|}{High-light} & \multicolumn{2}{c|}{Low-light} & \multicolumn{2}{c}{Average} \\ 
			\cline{3-8} 
			& & PSNR$\uparrow$ & SSIM$\uparrow$ & PSNR$\uparrow$ & SSIM$\uparrow$ & PSNR$\uparrow$ & SSIM$\uparrow$
		\\
        \hline
        	\ding{56} & \ding{56} & 26.35 & 0.894 & 27.61 & 0.825 & 27.13 & 0.851\\
        \ding{51} & \ding{56} & 29.90 & 0.950 & 30.67 & 0.808 & 30.38 & 0.861\\
        	\ding{56} & \ding{51} & 30.07 & 0.939 & 31.37 & 0.931 & 30.88 & 0.934 \\ \rowcolor{graycolor}
        \ding{51} & \ding{51} & \textbf{31.23} & \textbf{0.950} & \textbf{33.45} & \textbf{0.950} & \textbf{32.63} & \textbf{0.950} \\
        \hline\hline
        \end{tabular}}
\end{table}

\noindent{\textbf{Implementation Details.}}
We train on random $256\times256$ crops using PyTorch~\cite{paszke2019pytorch} and Adam~\cite{kinga2015method} ($\beta_1=0.9,\beta_2=0.99$), with batch size 4 and learning rate $10^{-4}$. LICN is initialized for 50 epochs with $(\gamma_1,\gamma_2)=(1,0.1)$ to learn linear correction. Joint training uses $(0.5,0.01)$ and halves the rate every 50 epochs. Training lasts 50 epochs on SMIT and 500 on CMER using an NVIDIA GeForce 4090.

\subsection{Comparison with State-of-the-Art Methods}

\noindent{\textbf{Quantitative Results.}}  
Table~\ref{tab: results on dataset} reports CMER-only training and SMIT pre-training followed by CMER fine-tuning under identical real-data supervision. RIECR achieves the best average PSNR/SSIM under both protocols and exposure directions: 32.63~dB/0.950 and 33.85~dB/0.951, exceeding the respective second-best averages by 1.61/1.27~dB and 0.014/0.013 SSIM. The gain from pre-training shows that SMIT complements rather than replaces real paired supervision.

\textit{(1) Physics-Aware vs. Generic Architectures.}
Generic sRGB architectures such as FECNet~\cite{huang2022deep} and RF~\cite{cai2023retinexformer} learn strong mappings but do not explicitly model RAW-linear exposure. LICN estimates illumination from channel-specific RAW cues, while FEN handles the exposure-dependent residual, producing consistent gains over both baselines.

\textit{(2) Unified vs. Task-Specific RAW Restoration.}
SID~\cite{chen2018learning} and SMG~\cite{xu2023low} emphasize low-light recovery, whereas CGNet~\cite{fu2023raw} and RawHDR~\cite{zou2023rawhdr} target high-light/HDR restoration. Against the strongest average RAW baseline, RRM~\cite{chen2025retinex}, RIECR gains 2.58/2.37~dB under the two protocols. This supports sharing linear correction while adapting residual restoration to the exposure direction.

\textit{(3) Computational Efficiency.}
RIECR uses 2.56~M parameters and 31.91~G MACs. Relative to CGNet/RawHDR, it reduces parameters by 85.5\%/75.6\% and MACs by 31.0\%/37.2\%; versus RRM it uses 66.5\% fewer parameters at comparable MACs. Lightweight models such as ECLNet, FECNet, RED, and ADU require fewer MACs but restore substantially worse, indicating that RIECR offers a favorable accuracy--complexity trade-off without relying on an excessively large parameter count.

\noindent{\textbf{Perceptual Quality.}} 
Fig.~\ref{fig:visual quality comparison} compares both exposure directions after the same ISP. For over-exposure, several methods leave color casts or miss clipped window structure, whereas RIECR preserves tone and highlight edges. For under-exposure, ENC~\cite{huang2022exposure} and FECNet~\cite{huang2022deep} retain artifacts or over-smooth details. RIECR suppresses amplified noise while maintaining boundaries, consistent with the LICN/FDC design.

\begin{table}[t]
    \renewcommand{\arraystretch}{1.0}
    \setlength{\tabcolsep}{4.0pt}
       \centering
       \caption{Component ablation of LICN and FEN.}
    \label{tab: ablation LICN}
    \label{tab: ablation FEN}
       \small
       \resizebox{\linewidth}{!}{%
        \begin{tabular}{l|cc|cc|cc}
        \hline\hline
		\multirow{2}{*}{\raisebox{-0.5ex}{\textbf{Methods}}} & \multicolumn{2}{c|}{High-light} & \multicolumn{2}{c|}{Low-light} & \multicolumn{2}{c}{Average} \\ 
			\cline{2-7} 
			& PSNR$\uparrow$ & SSIM$\uparrow$ & PSNR$\uparrow$ & SSIM$\uparrow$ & PSNR$\uparrow$ & SSIM$\uparrow$
		\\
        \hline
        \multicolumn{7}{l}{\textit{LICN variants}} \\
        w/o Linear & 30.47 & 0.924 & 30.22 & 0.913 & 30.31 & 0.917\\
        w/o Label & \textbf{31.68} & 0.944 & 32.43 & 0.920 & 32.15 & 0.929\\
        w/o Guidance & 30.10 & 0.940 & 31.88 & 0.937 & 31.21 & 0.938 \\
        \hline
        \multicolumn{7}{l}{\textit{FEN variants}} \\
        w/o Local & 29.41 & 0.944 & 31.39 & 0.929 & 30.65 & 0.935\\
        w/o Global & 30.65 & 0.950 & 30.33 & 0.926 & 30.45 & 0.935 \\
        w/o FDC & 29.08 & 0.932 & 32.08 & 0.926 & 30.96 & 0.929\\
        \hline
        \rowcolor{graycolor}
        \textbf{RIECR (Ours)} & 31.23 & \textbf{0.950} & \textbf{33.45} & \textbf{0.950} & \textbf{32.63} & \textbf{0.950} \\
        \hline\hline
        \end{tabular}}
\end{table}

\begin{figure}[t]
\small
\centering
  \setlength{\tabcolsep}{1.0mm}
  \renewcommand{\arraystretch}{1}
\begin{tabular}{ccc}
  \includegraphics[width=0.29\linewidth]{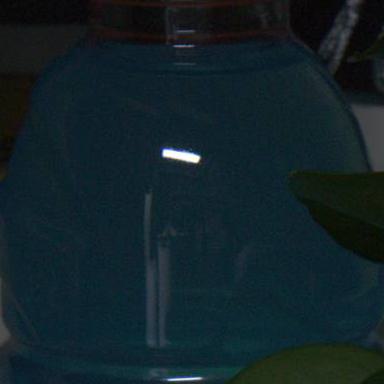}
  & \includegraphics[width=0.29\linewidth]{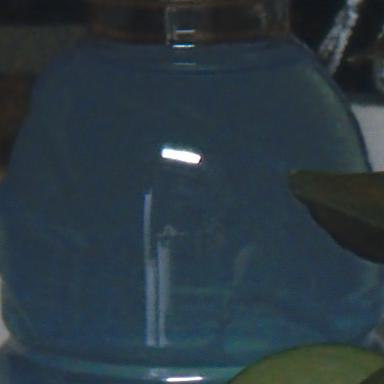}
  & \includegraphics[width=0.29\linewidth]{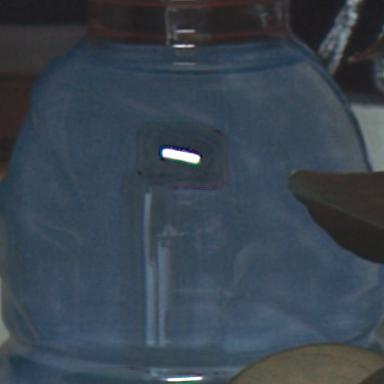} \\
Input & w/o LICN & w/o Global \\
  \includegraphics[width=0.29\linewidth]{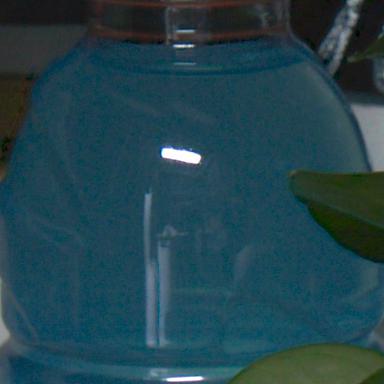}
  & \includegraphics[width=0.29\linewidth]{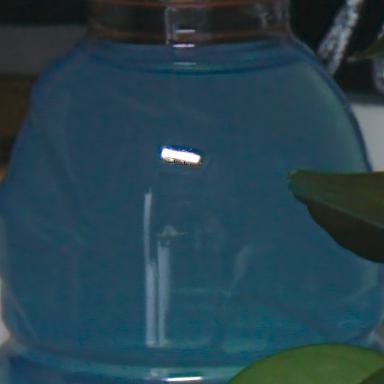}
  & \includegraphics[width=0.29\linewidth]{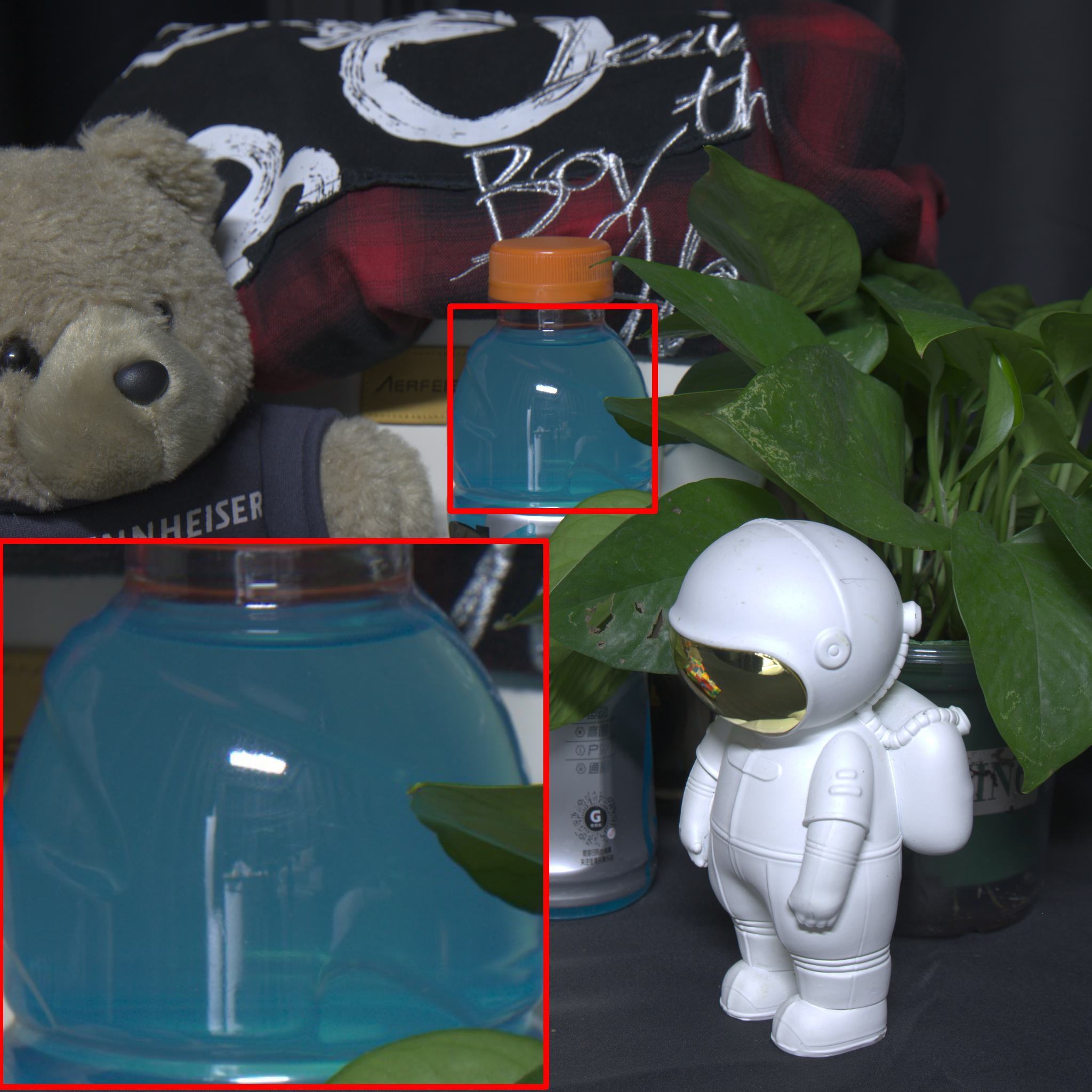} \\
w/o Local & Ours & Reference \\
  
  \end{tabular}
 \caption{Visual quality comparison of FEN. \textit{Local} and \textit{Global} mean the local convolution branch and global attention branch, respectively.}
 \label{fig: FEN}
\end{figure}

\subsection{Ablation Studies}
\label{sec:ablation}

\noindent{\textbf{Unified Framework.}} LICN and FEN raise the 27.13~dB baseline to 30.38 and 30.88~dB, respectively; together they reach 32.63~dB/0.950 and perform best on both subsets (Table~\ref{tab: ablation design}). This verifies that physically constrained illumination correction and learned residual restoration are complementary rather than interchangeable.

\noindent{\textbf{Linear Illumination Correction Network.}} 
In Table~\ref{tab: ablation LICN}, replacing linear gain with residual addition loses 2.32~dB; removing the exposure label or channel guidance also degrades performance. The variants confirm the need for exposure-aware gating and complementary Green and Red/Blue cues.
As visualized in Fig.~\ref{fig:illumination comparison}, our LICN achieves a result close to the reference image, validating the necessity of our channel-specific design.

\begin{figure}[t]
	\centering
    \small
    \setlength\tabcolsep{7 pt}
    \resizebox{\linewidth}{!}{%
    \begin{tabular}{ccc}
    \centering
    \includegraphics[width=0.23\linewidth]{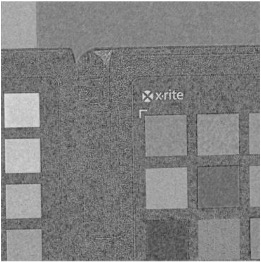} &
    \includegraphics[width=0.23\linewidth]{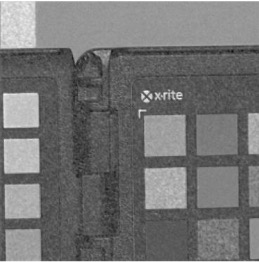} &
    \includegraphics[width=0.23\linewidth]{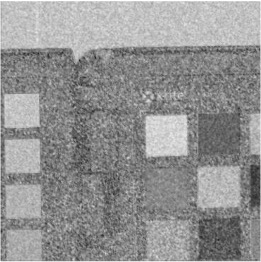} \\
    (a) FDC Input & (b) Low-frequency & (c) Standard Input \\
    \includegraphics[width=0.23\linewidth]{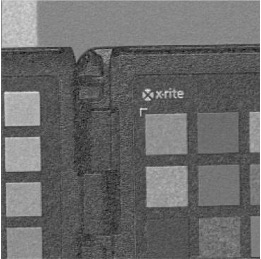} &
    \includegraphics[width=0.23\linewidth]{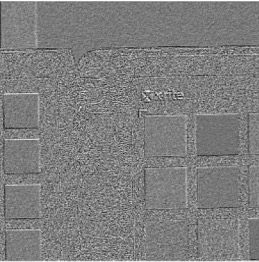} &
    \includegraphics[width=0.23\linewidth]{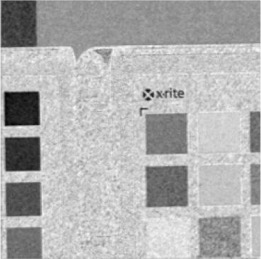} \\
    (d) FDC Output & (e) High-frequency & (f) Standard Output
    \end{tabular}}
    \caption{FDC features in the first global-branch layer: (a,d) FDC input/output, (b,e) decomposed low/high frequencies, and (c,f) standard-convolution input/output. FDC retains structure while isolating residual detail.}
    \label{fig: fdc}
\end{figure}

\noindent{\textbf{Frequency-Based Enhancement Network.}} 
Removing either FEN branch reduces performance in Table~\ref{tab: ablation FEN}. Fig.~\ref{fig: FEN} confirms that local processing suppresses noise while global attention supplies context; their combination best preserves structure.

\noindent{\textbf{Frequency Dynamic Convolution.}} 
The ``w/o FDC'' row in Table~\ref{tab: ablation FEN} drops average PSNR from 32.63 to 30.96~dB. Fig.~\ref{fig: fdc} shows that FDC retains low-frequency structure while isolating high-frequency residuals for exposure-adaptive modulation.

\begin{table}[t]
    \renewcommand{\arraystretch}{1.0}
    \setlength{\tabcolsep}{6 pt}
       \centering
       \caption{sRGB versus RAW input; metrics are evaluated in RGB.}
    \label{tab: ablation domain}
       \small
       \resizebox{\linewidth}{!}{%
        \begin{tabular}{c|cc|cc|cc}
        \hline\hline
		\multirow{2}{*}{\raisebox{-0.5ex}{\textbf{Setting}}} & \multicolumn{2}{c|}{High-light} & \multicolumn{2}{c|}{Low-light} & \multicolumn{2}{c}{Average} \\ 
			\cline{2-7} 
			& PSNR$\uparrow$ & SSIM$\uparrow$ & PSNR$\uparrow$ & SSIM$\uparrow$ & PSNR$\uparrow$ & SSIM$\uparrow$
		\\
        \hline
        sRGB & 27.14 & 0.911 & 29.00 & 0.902 & 28.35 & 0.905\\ \rowcolor{graycolor}
        RAW & \textbf{29.43} & \textbf{0.925} & \textbf{29.73} & \textbf{0.915} & \textbf{29.62} & \textbf{0.919}\\
        	\hline\hline
        \end{tabular}}
\end{table}

\begin{figure}[t]
    \small
    \setlength\tabcolsep{2.5 pt}
    \begin{tabular}{cccc}
    \centering
    \includegraphics[width=0.23\linewidth]{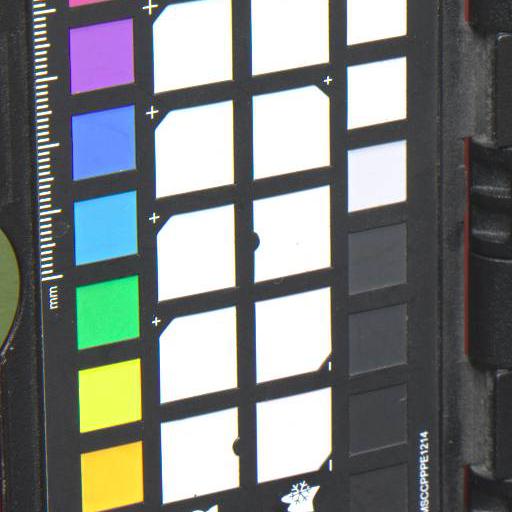} &
    \includegraphics[width=0.23\linewidth]{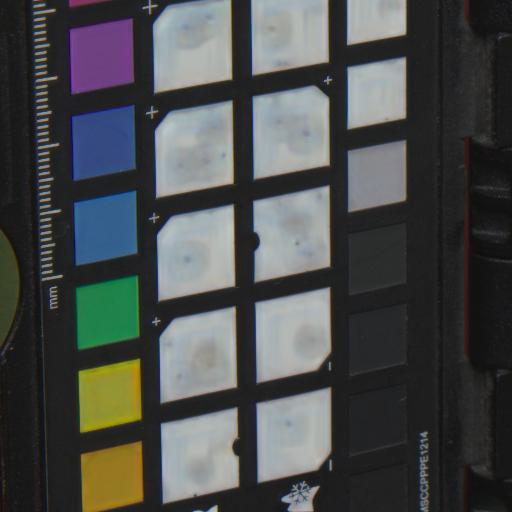} &
    \includegraphics[width=0.23\linewidth]{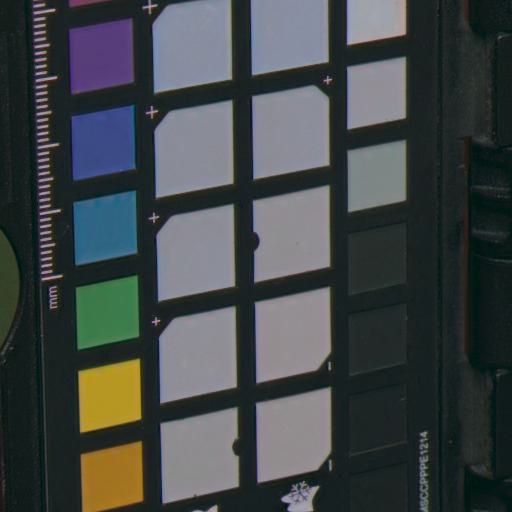} &
    \includegraphics[width=0.23\linewidth]{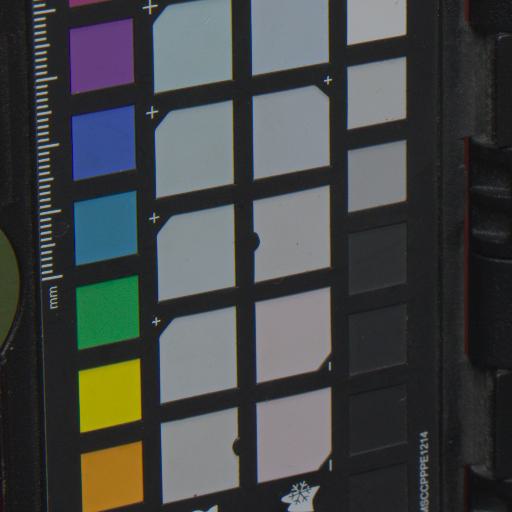} \\
    (a) Input & (b) sRGB & (c) RAW & (d) Reference
    \end{tabular}
    \caption{The visualized comparison between the sRGB and RAW domains.}
    \label{fig:domain comparison}
\end{figure}

\subsection{Discussion}
\label{sec:discussion}


\noindent{\textbf{Evaluation between RAW and sRGB.}} 
We train RIECR on CMER RAW data and sRGB counterparts generated by the same ISP~\cite{wei2021physics}, and evaluate both through that ISP. Table~\ref{tab: ablation domain} shows that RAW improves average PSNR/SSIM from 28.35~dB/0.905 to 29.62~dB/0.919. The gain occurs in both subsets because linear high-bit-depth RAW preserves highlights before tone mapping and enables physically consistent low-light scaling and noise restoration.
As shown in Fig.~\ref{fig:domain comparison}, the visual example further shows that RAW input preserves recoverable high-light details that are already compressed or clipped in sRGB. This confirms that the rich detail, higher bit depth, and linear response of RAW data are crucial for joint exposure correction and restoration.

\begin{figure}[!t]
    \small
    \setlength\tabcolsep{2.5 pt}
    \begin{tabular}{cccc}
    \centering
    \includegraphics[width=0.23\linewidth]{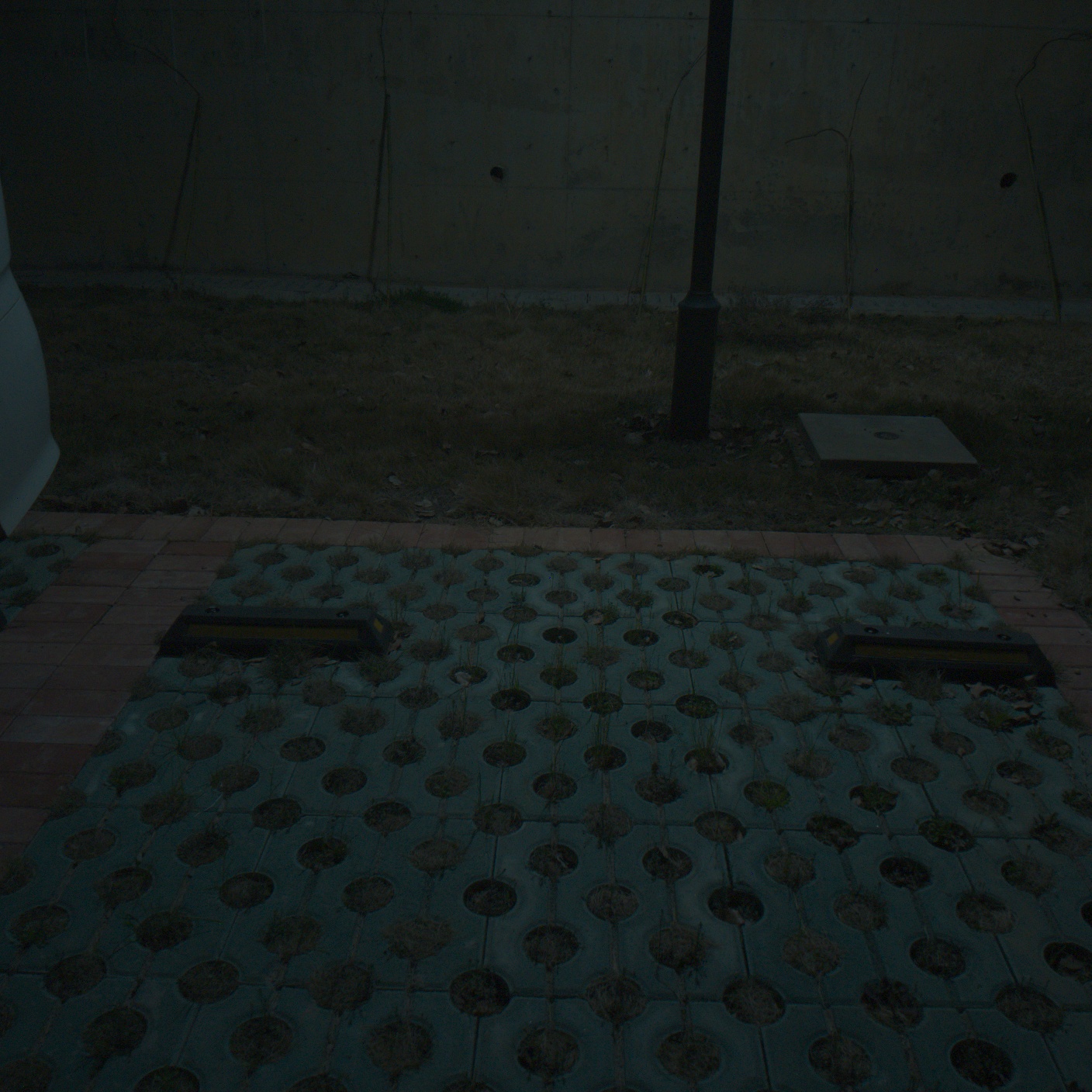} &
    \includegraphics[width=0.23\linewidth]{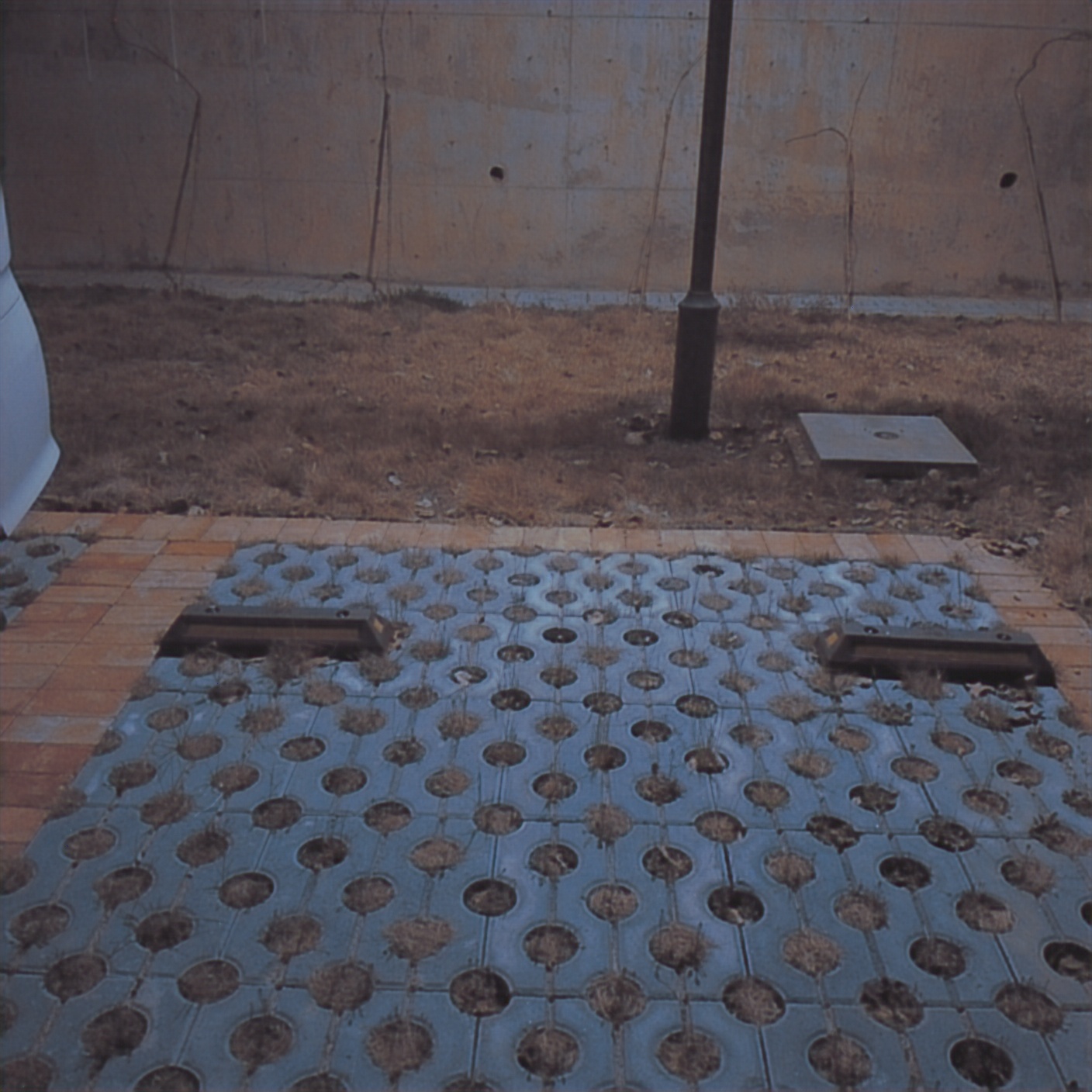} &
    \includegraphics[width=0.23\linewidth]{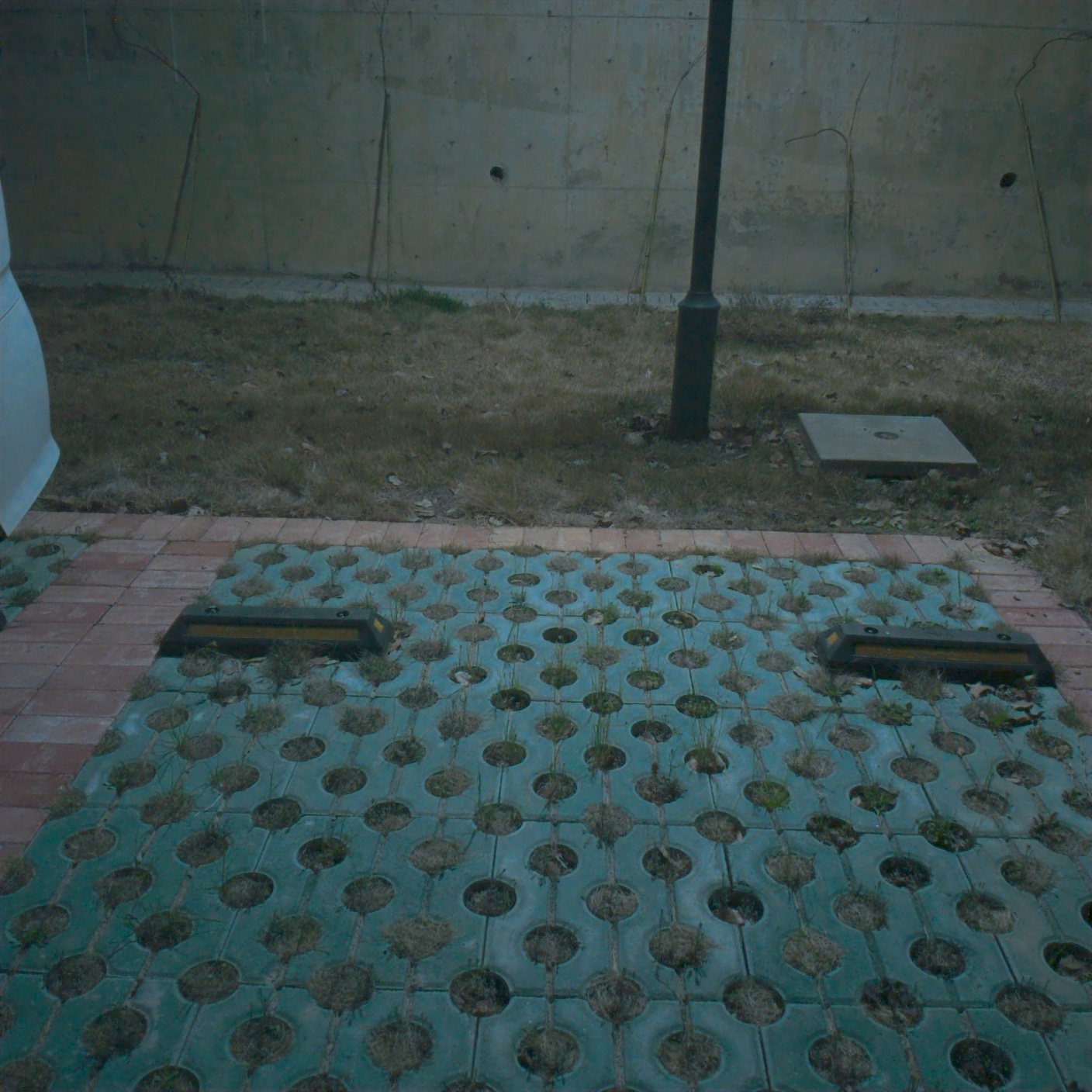} &
    \includegraphics[width=0.23\linewidth]{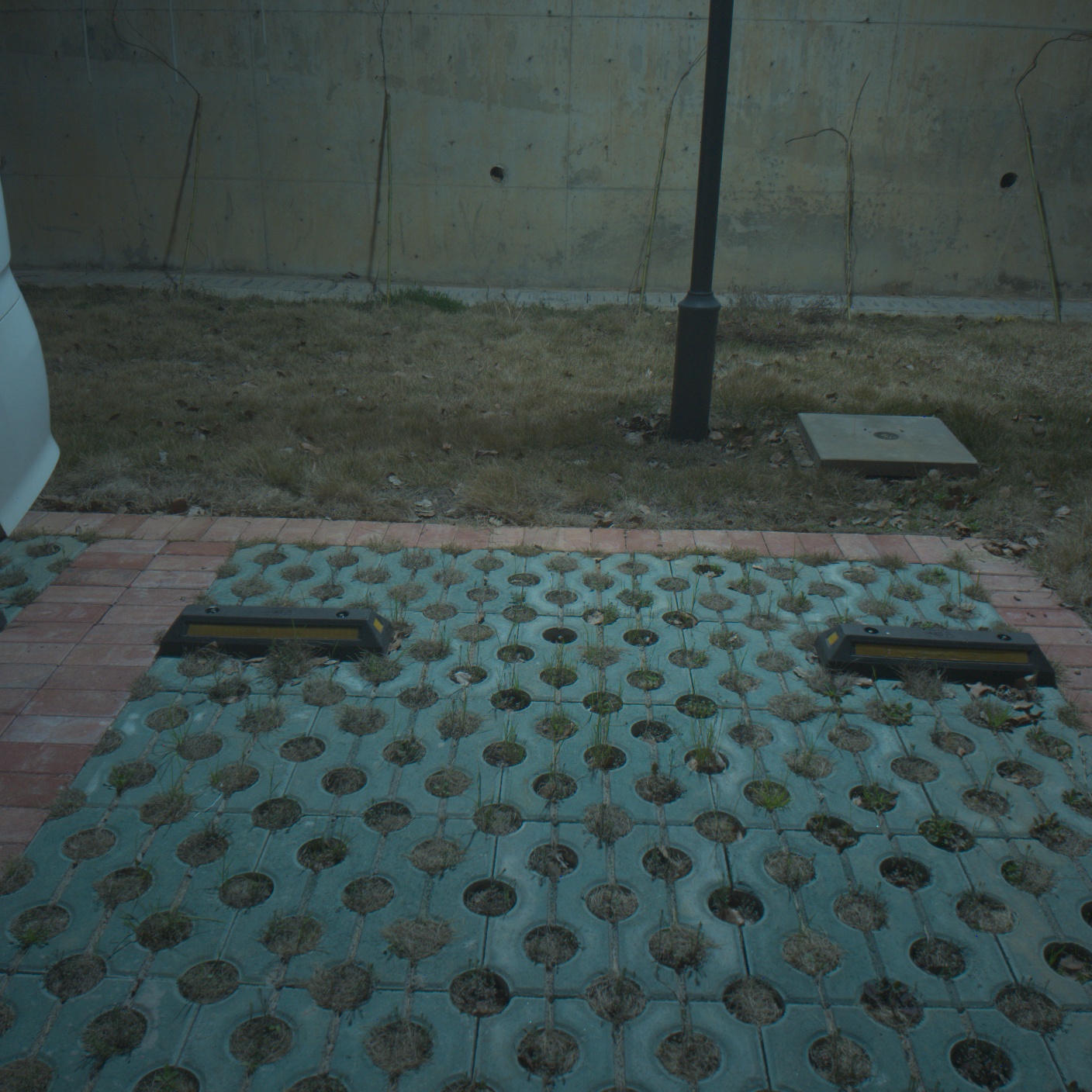} \\
    (a) Input & (b) ES~\cite{jung2025exposure} & (c) Ours & (d) Reference
    \end{tabular}
    \caption{The visualized results for the cross-camera setting. The model is trained on Canon 5D IV and evaluated on the Sony IMX586 sensor.}
    \label{fig:cross camera}
\end{figure}

\noindent{\textbf{Evaluation on Cross-Camera Setting.}} 
Trained on Canon 5D IV, RIECR transfers without retraining to the Sony IMX586 LRD dataset~\cite{zhang2023towards}. Fig.~\ref{fig:cross camera} shows corrected global tone and recovered structure despite sensor and pipeline changes, suggesting that the channel-aware decomposition is less device-specific than a learned sRGB mapping.

\begin{figure}[!t]
    \small
    \setlength\tabcolsep{2.5 pt}
    \begin{tabular}{cccc}
    \centering
    \includegraphics[width=0.21\linewidth]{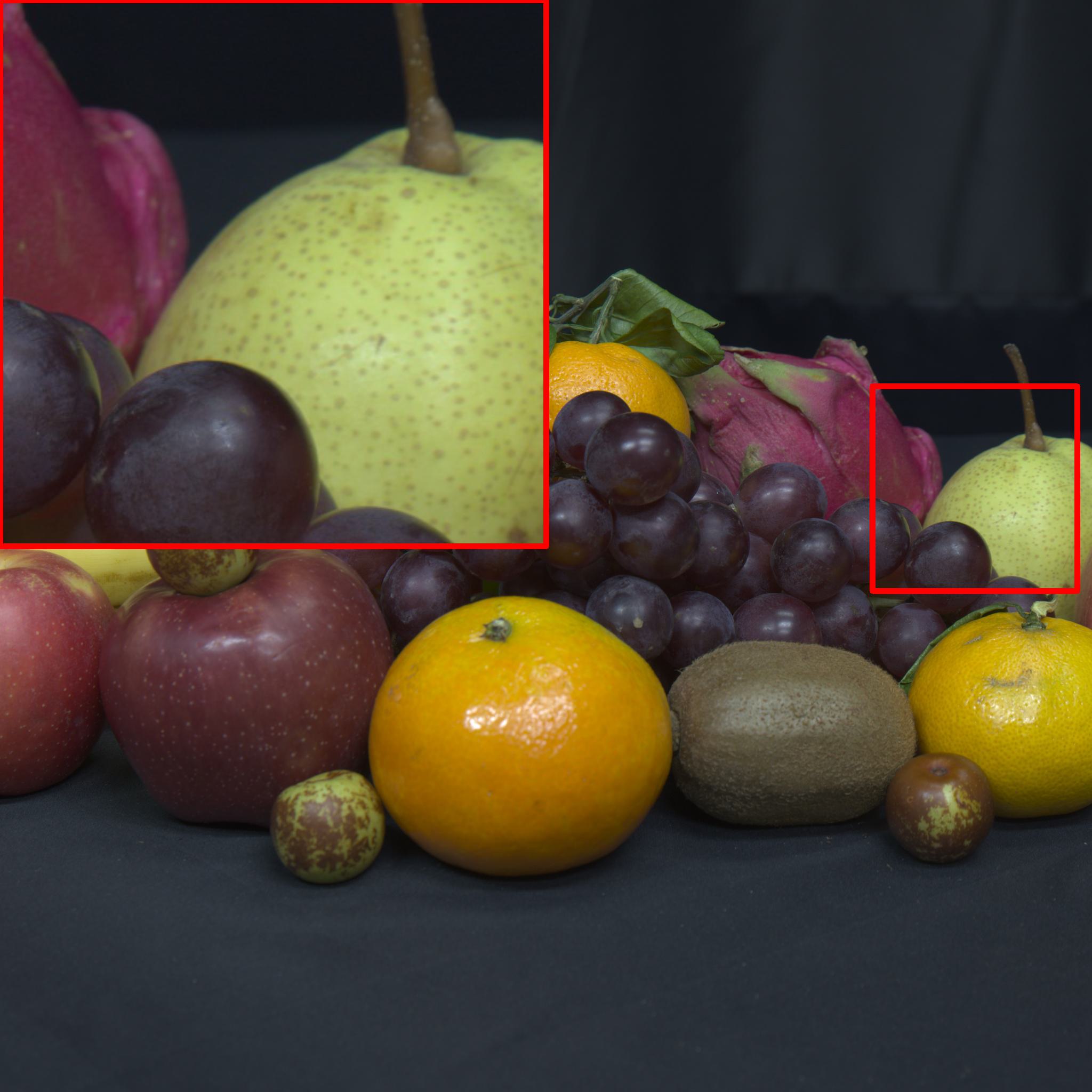} &
    \includegraphics[width=0.21\linewidth]{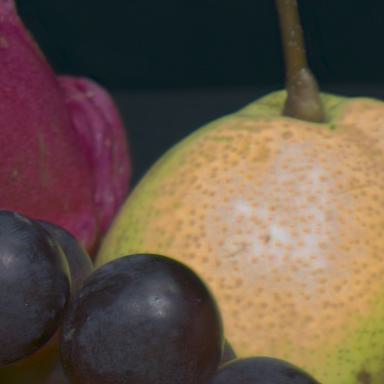} &
    \includegraphics[width=0.21\linewidth]{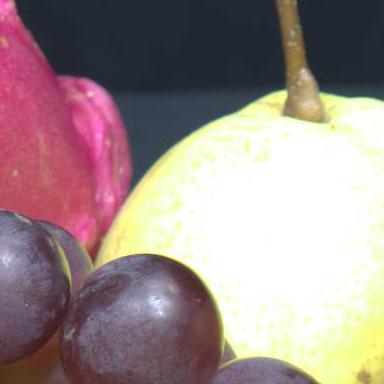} &
    \includegraphics[width=0.27\linewidth]{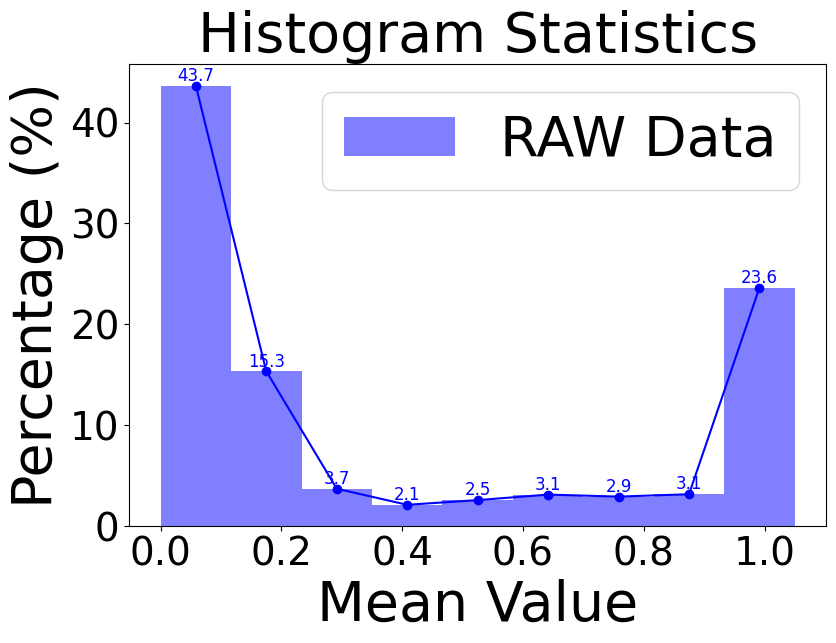} \\
    (a) Reference & (b) Ours & (c) Input & (d) Input data range
    \end{tabular}
    \caption{Failure example for over-exposure areas. The histogram of the input data range indicates that some pixels are saturated.}
    \label{fig:limitation}
\end{figure}

\noindent{\textbf{Failure Case and Limitation.}}
Table~\ref{tab: ablation domain} shows better detail recovery from RAW than from sRGB.
However, RIECR remains limited in large, fully saturated highlight regions (Fig.~\ref{fig:limitation}).
When all RAW channels are clipped, the original signal cannot be uniquely recovered from a single observation.
Contextual inference may produce plausible textures but cannot guarantee recovery of the original details or colors.
Future work could incorporate generative priors to improve visual plausibility in these severely degraded regions.

\section{Conclusion}
\label{sec:conclusion}

In this paper, we proposed RIECR, a unified RAW-domain framework that bridges the information loss of sRGB correction and the single-degradation focus of task-specific RAW restoration. Following the physical RAW formation model, RIECR decomposes joint under- and over-exposure restoration into linear illumination correction and residual enhancement. LICN estimates exposure-aware gains through channel-specific guidance, using the high SNR of Green channels for under-exposure and the less saturated Red/Blue responses for over-exposure. FEN then addresses the divergent residual degradations, while FDC separates low- and high-frequency responses to adaptively suppress amplified noise and recover weakened high-light structures. We further constructed CMER with 223 real scenes and 1,784 paired RAW images and built a synthetic dataset for pre-training. Comparisons with sixteen methods, component ablations, RAW/sRGB evaluation, efficiency analysis, and cross-camera tests validate the unified formulation and its components. RIECR achieves 33.85~dB/0.951 after pre-training and fine-tuning with 2.56~M parameters. Future work will extend CMER to dynamic scenes and broader sensors and explore metadata-aware correction and stronger priors for severely clipped regions.

\bibliographystyle{IEEEtran}
\bibliography{main}

\end{document}